\documentclass{article} 
\usepackage{iclr2025_conference,times}

\usepackage{amsmath,amsfonts,bm}

\def\eqref#1{equation~\ref{#1}}

\def\1{\bm{1}}

\def\eps{{\epsilon}}

\def\vh{{\bm{h}}}

\def\vx{{\bm{x}}}

\DeclareMathAlphabet{\mathsfit}{\encodingdefault}{\sfdefault}{m}{sl}
\SetMathAlphabet{\mathsfit}{bold}{\encodingdefault}{\sfdefault}{bx}{n}

\newcommand{\R}{\mathbb{R}}

\newcommand{\tool}[1]{#1}

\usepackage{hyperref}
\usepackage{cleveref}
\crefname{equation}{Eq.}{}
\usepackage{url}

\usepackage{ wasysym }
\usepackage{amssymb}
\usepackage{mathtools}
\usepackage{todonotes}
\usepackage{tkz-euclide}
\usepackage{booktabs}
\usepackage{subcaption}
\usepackage{spverbatim}
\usepackage{footmisc}
\usepackage{multirow}
\usepackage{orcidlink}

\newcommand\defeq{\stackrel{\mathclap{\normalfont\mbox{def}}}{=}}
\newcommand\lto[2]{\text{L2O}_{#1}(#2)}
\def\oursolver/{GANRA}
\def\ugotNL/{\textsc{ugotNL}}
\def\NRAgo/{NRAgo}

\title{Using GPUs And LLMs Can Be Satisfying \\ for Nonlinear Real Arithmetic Problems}
\author{Christopher Brix\orcidlink{0000-0002-8613-9454}, Julia Walczak\orcidlink{0009-0009-0582-4846}, Nils Lommen\orcidlink{0000-0003-3187-9217}, Thomas Noll\orcidlink{0000-0002-1865-1798} \\
RWTH Aachen University \\
\texttt{\{brix,lommen,noll\}@cs.rwth-aachen.de, julia.walczak@rwth-aachen.de}
}

\iclrfinalcopy 
\begin{document}

\maketitle
\begin{abstract}
Solving quantifier-free non-linear real arithmetic (NRA) problems is a computationally hard task.
To tackle this problem, prior work proposed a promising approach based on gradient descent.
In this work, we extend their ideas and combine \emph{LLMs} and \emph{GPU acceleration} to obtain an efficient technique.
We have implemented our findings in the novel SMT solver \textit{\oursolver/} (\textbf{G}PU \textbf{A}ccelerated solving of \textbf{N}onlinear \textbf{R}eal \textbf{A}rithmetic problems).\footnote{\url{https://github.com/ChristopherBrix/GANRA}\label{footnote:publish}}
We evaluate \oursolver/ on two different  NRA benchmarks and demonstrate significant improvements over the previous state of the art.
In particular, on the Sturm-MBO benchmark, we can prove satisfiability for more than five times as many instances in less than 1/20th of the previous state-of-the-art runtime.
\end{abstract}

\section{Introduction}
Satisfiability modulo theories (SMT) is an important framework to formalize many mathematical problems.
An SMT problem combines a formal theory with traditional satisfiability (SAT) solving.
To be precise, an SMT problem is constructed by replacing the Boolean variables of the SAT problem with formulas over the given formal theory.
Nowadays,  many problems in various areas of computer science, such as software verification, automated theorem proving, or analysis of hybrid systems, are reduced to SMT problems.
In this work, we focus on quantifier-free non-linear real arithmetic (NRA) as the underlying formal theory.
NRA considers polynomial equations and inequations over the domain $\R$ of real numbers.
For example, consider the following simple SMT problem:
\begin{equation}
    \phi(x_1,x_2,y_1,y_2) = \left( x_1^2 + x_2^2 = 1 \land y_1^2 + y_2^2 = 1 \land (x_1 - y_1)^2 + (x_2 - y_2)^2 \geq 1 \right)
    \label{eq:running_example}
\end{equation}
This SMT problem $\phi$ encodes two points $(x_1,x_2)$ and $(y_1,y_2)$ in the 2-dimensional plane on the unit sphere such that their distance is at least $1$.
This problem is also known as the ``Kissing number problem''.
Here, for example, $(x_1,x_2) = (1,0) \in \R^2$ and $(y_1,y_2) = (0,-1) \in \R^2$ is a model that satisfies $\phi$, i.e., $\phi(1,0,0,-1) = \textsc{true}$.

There already exist numerous techniques to prove or disprove satisfiability of such NRA problems, e.g., most prominently, cylindrical algebraic decomposition (CAD) \citet{collins1975QuantifierEliminationReal}, virtual substitution \cite{weispfenning1988ComplexityLinearProblems,weispfenning1997QuantifierEliminationReal}, or approaches employing Gröbner bases \citet{stengle1974NullstellensatzPositivstellensatzSemialgebraic}.
However, complete techniques such as CAD have doubly-exponential worst-case runtime.
Thus, in \citet{enricopaper,enricophd,lipparini2025satisfiability,liu_nrago}, the authors proposed an approach based on gradient descent to automatically prove satisfiability of NRA problems (and also problems with transcendental functions) efficiently.
More precisely, a function $f: \R^m \to \R$ is constructed of the NRA problem $\phi$ (where $m$ is the number of variables in $\phi$) such that if $\vx$ is a model for the original NRA problem, then $f(\vx) \leq 0$.
To find the roots of $f$, classical gradient descent is applied.
In this work, we extend their ideas and utilize GPUs to increase the performance.
To this end, we ``group'' similar operations such that GPU acceleration is applicable.
For example, all multiplications $x_1^2$, $x_2^2$, etc.\ could be performed in parallel by GPUs. 
In order to make this ``grouping'' automatable, we query LLMs.
To be precise, we use the LLM \tool{OpenAI o1-preview}.
So, we combine LLMs and GPUs, and make use of their recent progress to solve NRA problems.
Our approach is implemented as a prototype in the novel tool \oursolver/ (\textbf{G}PU \textbf{A}ccelerated solving of \textbf{N}onlinear \textbf{R}eal \textbf{A}rithmetic).\footref{footnote:publish}
We evaluated our tool on two sets of benchmarks: 
The Kissing benchmark set \citep{kissing} and the Sturm-MBO benchmark set \citep{sturm}.
Here, our prototype implementation already showed promising results.
To discuss the strengths of our approach further, we introduced a customizable benchmark set based on the Sturm-MBO benchmarks.

\paragraph{Contributions:}
In the following, we summarize our main contributions:
\begin{itemize}
    \item We establish ``grouping'' of similar operations as a key element to benefit from GPU acceleration. This is demonstrated in an empirical evaluation on two sets of benchmarks.
    \item We demonstrate that LLMs are capable of identifying patterns in benchmarks that allow for such operation grouping.
    This removes the need to manually encode similar operations, and motivates further research into optimizing this automatic detection.
    \item We provide a customizable benchmark set that allows for a detailed analysis of the performance of NRA-tools on increasingly complex polynomials.
    \item Our techniques are implemented in the novel tool \oursolver/. \oursolver/ is -- to the best of our knowledge -- the first SMT solver that uses a combination of formal methods and LLMs to solve NRA problems profiting of GPU acceleration.
\end{itemize}

\paragraph{Structure:}
First, we present related work in \Cref{sec:related_work}.
In \Cref{sec:general_approach}, we briefly recap how gradient descent can be used to prove satisfiability of NRA problems.
Afterwards, we introduce manual optimization techniques to improve GPU acceleration in \Cref{sec:manualGPU}.
In \Cref{sec:automaticGPU}, we present our automatic approach \oursolver/ in which we query the LLM \tool{OpenAI o1-preview} to perform the optimization for the GPU.
We demonstrate the strengths of our approach in an empirical evaluation on the Kissing and Sturm-MBO benchmarks in \Cref{sec:results}.
Finally, we discuss future work in \Cref{sec:future_work} and conclude in \Cref{sec:conclusion}.

\section{Related Work}
\label{sec:related_work}
Most state-of-the-art SMT solvers implemented different quantifier elimination techniques to solve NRA formulas. 
A survey on common quantifier elimination techniques such as cylindrical algebraic decomposition (CAD) \citet{collins1975QuantifierEliminationReal}, virtual substitution \citep{weispfenning1988ComplexityLinearProblems,weispfenning1997QuantifierEliminationReal}, and quantifier elimination using Gröbner bases \citep{stengle1974NullstellensatzPositivstellensatzSemialgebraic} is given by \citet{dolzmann1999RealQuantifierElimination}.
Complete quantifier elimination techniques like CAD are the centerpiece of many SMT solvers (see, e.g., \tool{CVC5} \citep{barbosa2022Cvc5VersatileIndustrialStrength}, \tool{SMT-RAT} \citep{corzilius2015SMTRATOpenSource}, \tool{Yices2} \citep{dutertre2014Yices22}, or \tool{Z3} \citep{demoura2008Z3EfficientSMT}).
However, such complete techniques have a doubly-exponential worst-case runtime.

Hence, we use a completely different (incomplete but efficient) approach in this paper.
We have implemented a technique based on gradient descent in our tool \oursolver/.
In contrast to complete techniques, gradient descent can only be used to prove (but not to disprove) NRA formulas.
The idea of identifying satisfying assignments of NRA problems by employing gradient descent was first proposed by \citet{enricopaper,enricophd} and independently by \citet{liu_nrago}.
In both cases, it was demonstrated that ``gradient descent SMT solving'' is a strong alternative to the existing techniques.
We are now continuing this line and adapt this technique such that it benefits from GPU acceleration.
Similar techniques are already showing promising results for bounded model checking \citep{osama2021GPUAccelerationBounded} or SAT solving \citep{osama2024CertifiedSATSolving}.

There exist already different approaches to parallelize SMT solvers.
However, to the best of our knowledge, there does not exist an SMT solver optimized for GPUs.
There are often two types of parallelization: Portfolio and divide-and-conquer solvers.
Portfolio solvers apply various techniques in parallel, e.g., \tool{SMT-RAT} \citep{corzilius2015SMTRATOpenSource}, \tool{SMTS} (the parallel version of \tool{OpenSMT}) \citep{marescottiSMTSDistributedVisualized}, or \tool{Z3} \citep{wintersteiger2009ConcurrentPortfolioApproach} use a portfolio of techniques.
They often improve efficiency by sharing intermediate results of different threads.
In contrast, divide-and-conquer approaches partition the SMT formula into smaller subproblems, e.g., \citep{zhao2024DistributedSMTSolving, wilson2023PartitioningStrategiesDistributed} in \tool{CVC5}, or \citep{hyvarinen2015SearchSpacePartitioningParallelizing} on top of \tool{OpenSMT}.

\section{Gradient Descent as a GPU-accelerated Heuristic}
\label{sec:general_approach}
Common NRA verification tools such as Z3 or CVC5 do not support GPUs to accelerate the computation.
\Citet{enricopaper,liu_nrago} proposed to encode a given NRA formula as a function over the reals, such that every satisfying assignment to the original NRA formula also constitutes a root of the corresponding function.
This allows to perform a gradient descent along the generated function.
For each identified root, the corresponding input can be tested to determine whether it is a valid satisfying assignment.

\subsection{Logic-to-Optimization Technique}
To utilize gradient descent to find potential satisfying inputs of the NRA problem, we must convert the original formula $\phi(\vx)$ for $\vx \in \R^m$ into a function $f: \R^m \rightarrow \R$.
This conversion is designed to guarantee that any model $\vx$ of $\phi$ (any $\vx$ that satisfies $\phi$) has $f(\vx) \leq 0$.
Notably, the inverse does not need to hold:
For some $\vx'$ with $f(\vx') \leq 0$, $\vx'$ may or may not be a model of $\phi$.
This concept of converting the formula into a corresponding function was introduced by \citet{enricopaper,liu_nrago}.
Note that our L2O (logic-to-optimization) transformations \cref{eq:l2o_eq,eq:l2o_lt,eq:l2o_leq} differ from \citep{enricopaper}, and all definitions differ from \citep{liu_nrago}.
In particular, we introduce a parameter $\eps$ that increases the set of inputs $\vx$ that yield $\lto{\eps}{\phi(\vx)} \leq 0$.
We provide a comparison with both proposals, as well as an evaluation of the impact of the choice of $\eps$, in \Cref{sec:l2o_ablation}, highlighting the importance of choosing an L2O transformation that facilitates the gradient descent.
As in \citet{enricopaper}, we assume that the formula is given as nested conjunctions and disjunctions, as well as constraints over real functions of the form $p(\vx) \thicksim 0$ for $\thicksim\, \in \{\leq, <, =\}$.
We enforce this by rewriting the formulas accordingly.
For example, we remove all negations by flipping the relevant relation comparisons.
\begin{align}
    \label{eq:l2o_eq} \lto{\eps}{p(\vx) = 0} &\defeq \max\{|p(\vx)| - \eps, 0\}  \\
    \label{eq:l2o_lt} \lto{\eps}{p(\vx) < 0} &\defeq \max\{p(\vx), -\eps\} \\
    \label{eq:l2o_leq} \lto{\eps}{p(\vx) \leq 0} &\defeq \max\{p(\vx), 0\} \\
    \label{eq:l2o_and} \lto{\eps}{\phi_1(\vx) \land \phi_2(\vx)} &\defeq \lto{\eps}{\phi_1(\vx)} + \lto{\eps}{\phi_2(\vx)}
\end{align}
Note that an encoding of a disjunction $\lto{\eps}{\phi_1(\vx) \lor \phi_2(\vx)}$ must make sure to return a non-positive value even if both $\lto{\eps}{\phi_1(\vx)} < 0$ and $\lto{\eps}{\phi_2(\vx)} < 0$ due to the choice of $\eps$, so a multiplication of both terms would not be sufficient.
\citet{liu_nrago} propose to use the minimum of both terms, which would adhere to this constraint.
As none of the two benchmarks we evaluate depend on this, we do not define it here.

\paragraph{Example}
For our running example in Equation~\ref{eq:running_example}, the L2O conversion yields the following result:
\begin{align*}
    & \lto{\eps}{\phi(x_1,x_2,y_1,y_2)} \\
    &= \lto{\eps}{x_1^2 + x_2^2 = 1} + \lto{\eps}{y_1^2 + y_2^2 = 1} + \lto{\eps}{(x_1 - y_1)^2 + (x_2 - y_2)^2 \geq 1} \\
    &= \lto{\eps}{x_1^2 + x_2^2 - 1 = 0} + \lto{\eps}{y_1^2 + y_2^2 - 1 = 0} \\
    & \phantom{XXX} + \lto{\eps}{1 - ((x_1 - y_1)^2 + (x_2 - y_2)^2) \leq 0} \\
    &= \max\{|x_1^2 + x_2^2 - 1| - \eps, 0\} + \max\{|y_1^2 + y_2^2 - 1| - \eps, 0\} \\
    & \phantom{XXX} + \max\{1 - ((x_1 - y_1)^2 + (x_2 - y_2)^2), 0\} \\
\end{align*}

\subsection{Ensuring Soundness}
As described above, the Logic-to-Optimization approach ensures that every satisfying assignment of $\phi$ guarantees $\lto{\eps}{\phi(\vx)} \leq 0$, but not every assignment $\vx'$ with $\lto{\eps}{\phi(\vx')} \leq 0$ necessarily satisfies $\phi(\vx')$.
Therefore, after identifying a potential candidate for a satisfying assignment, we verify its correctness by running Z3 on a modified version of the problem instance, where each input variable is bounded by a small region around the identified candidate assignment.
If Z3 rejects the assignment or times out, we restart the gradient descent from new random points.

If the benchmark does not contain any equality constraints, we can simplify this process:
Assume $\phi$ consists of a single constraint $p \thicksim 0$ with $\thicksim\,\in\{<,\leq\}$.
Then, we can test if $\lto{\eps}{\phi(\vx)} \thicksim 0$ holds for our found candidate solution $\vx$.
Here, this is a sufficient condition: Any such input $\vx$ is a valid model of $\phi$.
For conjunctions and disjunctions over multiple such constraints, we can perform this test individually and compute the respective conjunction and disjunction.
However, this is not a sufficient condition for constraints $p = 0$.
Thus, we use the SMT solver Z3 to prove correctness of our candidate $\vx$ for equalities.

\section{Manual Optimization for GPU Acceleration}
\label{sec:manualGPU}
GPUs can accelerate computations by performing a large number of similar operations in parallel, in particular matrix-matrix multiplications.
Therefore, to maximize the speedup gained from using a GPU, it is crucial to structure the required operations in such a way that similar operations are grouped.

We identified two possible axes for this optimization:
\begin{enumerate}
    \item Batching.
    Instead of sampling a single initial assignment $\vx$ and optimizing it via gradient descent, multiple individual initial assignments can be sampled.
    Propagating them through $\lto{\eps}{\phi(\vx)}$ can be done in parallel, effectively scaling the computation speed by the chosen batch size.
    \item Grouping.
    If the computation of the L2O function requires to perform the same operation on different terms, these can be grouped, similar to batching.
    Crucially, this does not increase the number of samples evaluated, but reduces the required processing time per batch.
    E.g., for the function $f(\vx) = (x_1 + x_2) \cdot (x_3 + x_4)$, the two sums $x_1 + x_2$ and $x_3 + x_4$ can be computed in parallel.
    This reduces the required computation to two sequential operations, instead of three.
\end{enumerate}
In the following sections, we detail the grouping optimizations we performed for the Kissing and Sturm-MBO benchmarks.
An evaluation of the importance of these optimizations is given in Section~\ref{sec:results}, where we compare the parallelized version to a sequential computation.

\subsection{Kissing}
\label{sec:manual_optimization_kissing}
The Kissing benchmark \citep{kissing} consists of 45 instances.
Each instance encodes a kissing number problem: 
One searches for the position of $N$ points on an $M$-dimensional unit sphere such that the pairwise distance between these points is greater than or equal to 1.
For $N$ unit spheres with $\vx_1, \ldots, \vx_N \in \R^M$, this is encoded by
\begin{align*}
    \phi(\vx_1, \ldots, \vx_N) = \left(\bigwedge_{n \in \{1,\dots,N\}} \vx_n^T \vx_n = 1 \land \bigwedge_{\substack{m,n \in \{1,\dots,N\}\\ m < n}} (\vx_n - \vx_m)^T (\vx_n - \vx_m) \geq 1\right)
\end{align*}
From this vectorized formulation, it is immediately obvious that the inner subtractions and all multiplications can be performed in parallel.

\subsection{Sturm-MBO}
\label{sec:manual_optimization_mbo}
The Sturm-MBO benchmark \citep{sturm} consists of 405 instances, of which we select the 105 and 15 instances respectively known or unknown to be satisfiable\footnote{As indicated in their status tag}.
Each instance defines a subset of the input variables $\{h_1, \ldots, h_6, j_2\}$ restricted to the positive reals.
A satisfying assignment is a root of a multilinear polynomial.
Here, each monomial of the polynomial is a scalar integer value, multiplied by $N$ sampled instances of the variables $\{h_1, \ldots, h_6\}$.
As $N$ is often greater than $6$, the constructed product often consists of powers of $h_i$.
Finally, each product is optionally multiplied by $j_2$ with a power of up to $N$.
E.g., for $N=10$, the polynomial function has therefore the form
\begin{align}
    \label{eq:mbo_example}
    f(\vh) = 2 \cdot h_1^4 \cdot h_2^2 \cdot h_3 \cdot h_5^3 + 13 \cdot h_1 \cdot h_2^2 \cdot h_3 \cdot h_4^3 \cdot h_5^2 \cdot h_6 \cdot j_2^8 + \ldots
\end{align}

We optimize the computation for this benchmark by first precomputing all values $\bigcup_{i \in \{1, \ldots, 6\}} \{h_i^n \mid n \in \{1, \ldots, N\}\} \cup \{j_2^n \mid n \in \{1, \ldots, N\}\}$.
As these are independent of each other, they can be computed in parallel.
Next, we initialize a vector with the scalar integers in each product.
Then, for $i \in \{1, \ldots, 6\}$, we select the required power of $h_i$ in each product and copy it into a new tensor.
We do the same for the required powers of $j_2$.
This can be done in parallel for all products.
Subsequently, these eight tensors (one for scalars, six for $h_i$, and one for $j_2$) are multiplied with each other, resulting in one tensor with the values of each product.
Finally, the tensor is summed up, to get the required sum of all products.

A satisfying assignment of the benchmark requires the sum of products to be exactly equal to zero.
Since finding those inputs via gradient descent may often be infeasible, as the optimization may constantly overshoot the actual assignment, we transform the benchmark further:
We duplicate $\vh$ to $\underline{\vh} = \{\underline{h}_1, \ldots, \underline{h}_6, \underline{j}_2\}$ and $\overline{\vh} = \{\overline{h}_1, \ldots, \overline{h}_6, \overline{j}_2\}$.
Instead of $f(\vh) = 0$, we require ${f(\underline{\vh}) < 0} \land {f(\overline{\vh}) > 0}$.
If this new benchmark version can be satisfied, the intermediate value theorem guarantees that the original benchmark could also be satisfied.
One could then use interval iteration to find precise lower and upper bounds to the original $x_i$ variables, potentially enabling other solvers to find the precise values using the limited search space.

\section{LLM-Based Optimization for GPU Acceleration}
\label{sec:automaticGPU}
The optimizations described in Sections~\ref{sec:manual_optimization_kissing}~and~\ref{sec:manual_optimization_mbo} were performed by manually inspecting the benchmark to identify the underlying patterns.
This is a time consuming approach that can only be justified if there are only a few different benchmark categories to evaluate, and if each category consists of many similarly structured problem instances.
As the number of benchmarks increases, a manual optimization may become infeasible.

Therefore, we investigated whether the manual optimization of the computation process could be offloaded to an LLM.
Ideally, the LLM would identify the underlying pattern of each benchmark, and provide the corresponding Python code to compute the L2O formula in an optimized manner.
If this succeeds, the GPU acceleration becomes feasible even for large sets of diverse benchmark categories.

The usage of LLMs to generate the Python implementation of the L2O formula has two main risks: Syntactic errors and semantic errors.
Crucially, neither of them can cause \oursolver/ to produce unsound outputs, as they would in the worst case lead to non-termination:
\begin{enumerate}
    \item Syntactic errors:
    The LLM may fail to generate Python code that can be executed.
    This can trivially be detected by trying to execute it - should the process crash, the LLM may either be queried to generate an updated code block, or grouping of computations may be disabled.
    Then, one could either still use GPU acceleration based on the non-grouped implementation of the L2O formula, that performs all operations in the order they are trivially specified.
    Alternatively, one may fall back to a solver such as Z3 that does not rely on gradient descent.
    \item Semantic errors:
    The LLM may generate code that is syntactically correct, but is not equivalent to the L2O formula.
    This kind of error may be detected by comparing the results of the LLM-generated code on a few inputs against the slow but known good ground-truth implementation without grouping.
    Crucially, even if this test fails to identify a semantic error, the overall process is still sound:
    Computing the roots of the incorrect formula will yield candidates that are not actually satisfying assignments to the original formula.
    However, as these are always checked for correctness, they would always be discarded as a spurious assignment.
    Therefore, our solver would never return an actual satisfying assignment, and eventually time out.
\end{enumerate}

The LLM is prompted by providing two example instances and asking it to generate efficient pytorch code to compute the required functions.
We provide the exact LLM prompt in Appendix~\ref{sec:llm_code_prompt}.
In the current version of \oursolver/, the reply of the LLM is manually inspected for the code block, which is then copied into the source code of \oursolver/ without further edits.
This process could be fully automated by employing an API to the LLM and automatically detecting and executing code blocks.

In our experiments, the LLM is presented with two example inputs from the benchmark category and is asked to identify patterns and extrapolate them to write efficient code.
The prompt may be improved by providing additional information such as statistics on term reuse.
For example, using Python's SymPy package \citep{sympy}, the number of occurrences of each term may be computed and passed as additional input.
Exploring such extensions is left for future work.

\subsection{Case Studies}
\label{sec:llm_optimizations} 
We provide the LLM generated code for the two benchmarks in Appendix~\ref{sec:llm_code_prompt}.
Here, we want to highlight some of the correctly identified grouping opportunities, as well as missed opportunities that cause the LLM generated code to be suboptimal.

\paragraph{Kissing}
The LLM finds some of the possible groupings, but not all of them:
It computes the multiplication of $\vx_n^T \vx_n$ for all $n \in \{1,\dots,N\}$ in parallel.
Then, it computes all pairwise squared distances between all $\vx_n, \vx_m$ vectors. 
Crucially, it does not take into account that only the distances for $m < n$ are required, doubling the necessary computations.
To account for that, only the relevant subset of the pairwise distances is then selected for the final result.

\paragraph{Sturm-MBO}
The initial generated code contained a syntax error.
We replied to the LLM by copying this error verbatim, which caused it to provide an updated version of the code.
Such a feedback loop could be done automatically.

The LLM successfully identifies the potential to speed up the computation by grouping the computation of each product, instead of performing the multiplications sequentially.
However,  if multiple products contain the same power of the same variable (e.g., $x_2^2$ in Equation~\ref{eq:mbo_example}), the LLM does not precompute this value.
Instead, it computes the power for each product from scatch, significantly increasing the computation overhead.

\section{Results}
\label{sec:results}
We compare the performance of \oursolver/ against Z3, CVC5, \ugotNL/ \citep{lipparini2025satisfiability}\footnote{Note that this is an updated version of the solver presented in \citep{enricopaper} that achieves stronger results than the initial version.} and \NRAgo/ \citep{liu_nrago} on two benchmarks from the SMT-LIB benchmark set \citep{smtlib}:
Kissing \citep{kissing} and 20161105-Sturm-MBO (Sturm-MBO) \citep{sturm}.
For each benchmark, we evaluate the following setups for \oursolver/:
\begin{enumerate}
    \item an L2O process with $\eps = 10^{-4}$ and sequential, non-optimized code,
    \item an L2O process with $\eps = 0$ and manually optimized code,
    \item an L2O process with $\eps = 10^{-4}$ and manually optimized code, and
    \item an L2O process with $\eps = 10^{-4}$ and LLM-generated code to compute real functions.
\end{enumerate}
All experiments were performed using a single CPU core with 5G RAM and an NVIDIA H100 GPU.
Unless otherwise noted, the timeout for each instance was 10 minutes.
We always randomly sample a batch of $10\,000$ initial assignments from the interval $[-1, 1]$ and optimize them in parallel, using batching.
The gradient descent was done using the Adam optimizer introduced in \citep{adam} and a learning rate of $10^{-3}$.
After each update, values were projected back to $[-1, 1]$.

\subsection{Kissing}
\begin{table}[t]
\centering
\begin{tabular}{@{}llllllllll@{}} \toprule
& & Z3 & CVC5 & \ugotNL/ & \NRAgo/ & \multicolumn{4}{c}{\oursolver/ (ours)} \\ \cmidrule(l){7-10}
& & & & & & seq. & \multicolumn{2}{c}{manual} & LLM \\ \cmidrule(lr){8-9}
& & & & & \multicolumn{1}{r}{$\eps = $} & $10^{-4}$ & $0$ & $10^{-4}$ & $10^{-4}$ \\ \midrule
\parbox[t]{2mm}{\multirow{2}{*}{\rotatebox[origin=c]{90}{Kis.}}} & SAT [$\uparrow$] & 34 & 16 & 27 & 39 & 39 & 18 & \textbf{40} & \textbf{40} \\
& avg. runtime [s] [$\downarrow$] & 38.01 & - & - & 24.72 & 30.45 & - & \textbf{9.13} & 9.61 \\ \midrule
\parbox[t]{2mm}{\multirow{2}{*}{\rotatebox[origin=c]{90}{MBO}}}
 & SAT [$\uparrow$] & 1 & 0 & 10 & 6 & 10 & \textbf{57} & \textbf{57} & 38 \\
 & avg. runtime [s] [$\downarrow$] & - & - & 139.08 & - & 333.62 & \textbf{6.39} & 14.86 & 35.92 \\ \bottomrule
\end{tabular}
\caption{Toolkit evaluation on the Kissing and Sturm-MBO benchmarks. We report the average runtime of those instances that could be verified by all tools that can prove satisfiability for at least 30 (Kissing) and 10 (Sturm-MBO) instances. $\uparrow$ ($\downarrow$): Higher (lower) values are better.}
\label{tab:results_kissing_mbo}
\end{table}

As comparing the average runtime of tools that successfully process a different set of inputs is unfair, we report the average runtime over those instances that can be solved by all tools that solve at least 30 instances.
Table~\ref{tab:results_kissing_mbo} summarizes the results.
We outperform all other tools both in terms of total number of found satisfiable instances, and the average runtime to find them.
Interestingly, the LLM-based implementation performs on par with the manually optimized one, even though it includes unnecessary computations (see Section~\ref{sec:llm_optimizations}).
This indicates that the LLM does not necessarily need to find the optimal solution, as even partially optimized code can already exhibit great performance.

Using $\eps = 0$ causes the performance of \oursolver/ to deteriorate.
We hypothesize that this is because the Kissing benchmark includes multiple equality constraints.
Without an $\eps > 0$ in Equation~\ref{eq:l2o_eq}, the L2O formula would only have pointwise roots, making the optimization problem very hard, as it would require to find the precise corresponding scalar inputs.
By setting $\eps > 0$, we increase the area in which the gradient descent may be stopped, to allow testing the potential satisfying input.

\subsection{Sturm-MBO}
Table~\ref{tab:results_kissing_mbo} summarizes the results for the Sturm-MBO benchmark.
Here, all optimized versions of \oursolver/ outperform all baselines.
The performance of the LLM-based version lacks behind the manually optimized one, but can still prove satisfiability for almost four times as many instances as \ugotNL/, the best existing tool.

Note that for \oursolver/, we use a modified version of the Sturm-MBO benchmark (see Section~\ref{sec:manual_optimization_mbo}):
Instead of finding a root of the given polynomial function, we find inputs that yield positive/negative outputs.
This simplifies the optimization process, while concrete inputs could be found using interval iteration.
The other tools perform slightly worse on this modified benchmark version, so here, we report their performance on the original instances.
We provide a comparison of the performances across both versions in Appendix~\ref{sec:ablate_preprocessing}.

\subsection{Ablation Study on Sturm-MBO Like Problems}

\begin{figure}[t]
    \centering
    \begin{subfigure}{.5\textwidth}
  \centering
  \includegraphics[width=0.9\linewidth]{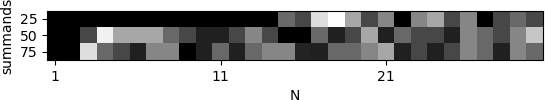}
  \caption{Z3}
  \label{fig:sub1}
\end{subfigure}%
\begin{subfigure}{.5\textwidth}
  \centering
  \includegraphics[width=0.9\linewidth]{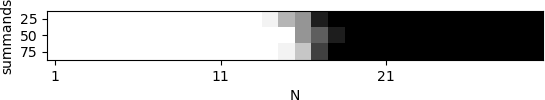}
  \caption{\oursolver/}
  \label{fig:sub2}
\end{subfigure}
    \caption{
\vspace{-0.05cm}Comparison of Z3 and \oursolver/ on Sturm-MBO-like problems with varying number of summands and $N$. White indicates 0\% success at finding a satisfying input, black 100\%.}
    \label{fig:mbo_like}
\end{figure}

In order to better demonstrate the domain in which \oursolver/ outperforms conventional approaches, we create a new Sturm-MBO like benchmark, with tunable hyperparameters.
We generate new instances (according to the pattern discussed in Section~\ref{sec:manual_optimization_mbo}) for $P$ products in the sum, and $N$ being the sum of all exponents of $h_1, \ldots, h_6$ per product.
We explore all combinations of $P \in \{25, 50, 75\}$ and $N \in \{1, \ldots 30\}$.
We provide additional information in Appendix~\ref{sec:custom_mbo_details}.
As we do not know apriori whether the generated instances will be satisfiable or unsatisfiable, we attempt to find a satisfying input using both Z3 and \oursolver/ with a shortened timeout of three minutes.
Any instance that cannot be shown to be satisfiable by at least one of the two solvers is discarded.
We repeat this process until we have 10 known satisfiable instances per hyperparameter combination.\footnote{If we cannot generate 10 instances within one hour, we terminate and use a reduced set of instances for this hyperparameter combination.}

In Figure~\ref{fig:mbo_like} we report the percentage of verified instances per tool.
It is evident that \oursolver/ shines particularly for a higher value of $N$, i.e., for complicated products.
This is intuitive, as more complex polynoms complicate the solving process for Z3, but gradient descent can be applied regardless.
Future work may investigate heuristics that could be used to predict whether a given benchmark would benefit from a gradient descent-based approach.

\subsection{Alternative L2O Transformations}
\label{sec:l2o_ablation}
\begin{table}[t]
\centering
\begin{tabular}{@{}lllllll@{}} \toprule
& \multicolumn{3}{c}{Kissing} & \multicolumn{3}{c}{Sturm-MBO} \\ \cmidrule(lr){2-4} \cmidrule(lr){5-7}
L2O version & \oursolver/ & \NRAgo/ & \ugotNL/ & \oursolver/ & \NRAgo/ & \ugotNL/ \\ \midrule
SAT [$\uparrow$] & \textbf{40} & 38 & 39 & \textbf{57} & \textbf{57} & 56 \\
avg. runtime [s] [$\downarrow$] & 9.62 & \textbf{6.79} & 12.69 & \textbf{38.42} & 42.92 & 113.01 \\ \bottomrule
\end{tabular}
\caption{Comparison of the different L2O definitions proposed in our work vs. \citep{enricopaper} and \citep{liu_nrago}. We integrated their L2O definitions into \oursolver/, so all results benefit from GPU acceleration. $\uparrow$ ($\downarrow$): Higher (lower) values are better.}
\label{tab:l2o_ablation}
\end{table}

To ablate the impact of the given L2O definition, we integrate the L2O definitions from \citep{enricopaper} and \citep{liu_nrago} in \oursolver/.
In Table~\ref{tab:l2o_ablation}, we see the respective performances of all three definitions on both benchmarks.
On the Kissing benchmark, all three versions perform with different tradeoffs between number of found satisfiable inputs and average runtime.
On the Sturm-MBO benchmark however, the \ugotNL/ L2O definition is significantly slower than both our proposed definition and the one by \citep{liu_nrago} in \NRAgo/.
We propose to investigate these differences in more detail in future work, to potentially find a new L2O definition that merges the individual benefits of each existing version.

\section{Future Work}
\label{sec:future_work}
\vspace{-0.1cm}
Even though we successfully demonstrate that LLMs are capable of writing efficient code for the computation of given formulas, they also lack behind the performance of hand-written code.
This demonstrates two orthogonal research directions:
On the one hand, a tuning of the LLM prompt, or a better feedback loop that allows the LLM to iteratively improve its code may be able to increase the performance of the LLM-based implementation.
On the other hand, a more thorough analysis of a larger set of benchmarks may unearth underlying patterns that would allow to write hand-optimized code that directly supports a wide variety of different benchmarks.
This would potentially remove the need for LLMs entirely.

We also note that our implementation does not benefit from any of the improvements implemented in \ugotNL/ and \NRAgo/, such as Basinhopping \citep{wales_basinhopping} to improve the process of finding roots, or the topological degree test \citep{cho_topological_degree_test, franek_topological_degree_test} to remove our reliance on Z3 for verification of found candidates.
Furthermore, \oursolver/ has no ability to prove unsatisfiability. 
For unsat instances, it would always result in a timeout, making them indistinguishable from hard but satisfiable instances.
Ultimately, the concept of \oursolver/ should therefore be integrated with a complete solver by calling \oursolver/ as part of a portfolio approach.

Finally, we note that our evaluation focused on two benchmarks only.
Evaluating \oursolver/ on more benchmarks, preferably the entire set of available quantifier-free NRA problems, may provide valuable insights into its strength and failure cases.

\section{Conclusion}
\label{sec:conclusion}
\vspace{-0.1cm}
We have evaluated our novel NRA solver \oursolver/ on two different benchmarks and demonstrated that using GPUs can significantly improve the speed of detecting potential satisfiable assignments for NRA problems.
To go beyond the benefit of batching, ``grouping'' of similar operations can be employed to gain further speedups.
While a manual implementation and optimization of each benchmark yields the highest performance, we have demonstrated that LLMs are capable of identifying underlying patterns and exploiting them to write efficient Python code themselves, removing the need for manual intervention.

\subsubsection*{Acknowledgments}
We want to thank the anonymous reviewers for their valuable feedback.
We also want to thank Enrico Lipparini for his help in understanding \ugotNL/, insights into the current state of the art, and positioning \oursolver/ in this research landscape.
Computations were performed with computing resources granted by RWTH Aachen University under project rwth1739 and p0023175.

\bibliography{iclr2025_conference}
\bibliographystyle{iclr2025_conference}

\newpage
\appendix
\section{Appendix}

\subsection{Custom MBO Benchmark Generation Details}
\label{sec:custom_mbo_details}
For each instance, we define a fixed $N$ (see Section~\ref{sec:manual_optimization_mbo}) and uniformly sample which of the six $h_1, \ldots, h_6$ to use with which exponents.
The variable $j_2$ is multiplied only with a subset of products: We flip a weighted coin and multiply by $j_2^k$ in $40\%$ of the cases, for a uniformly sampled $k \in \{1, \ldots, N\}$.
Finally, we uniformly sample the scalar factors of each product from the integers 1 to 20, and assign a negative sign to them in $20\%$ of the cases.

\subsection{Tool Performances on the Original vs. Modified Sturm-MBO Benchmark}
\label{sec:ablate_preprocessing}
\begin{table}[]
    \centering
\begin{tabular}{@{}lllllllll@{}} \toprule
& \multicolumn{2}{c}{Z3} & \multicolumn{2}{c}{CVC5} & \multicolumn{2}{c}{\ugotNL/} & \multicolumn{2}{c}{\NRAgo/} \\ \cmidrule(lr){2-3} \cmidrule(lr){4-5} \cmidrule(lr){6-7} \cmidrule(l){8-9}
& Orig & Interval & Orig & Interval & Orig & Interval & Orig & Interval \\ \midrule
SAT [$\uparrow$] & 1 & 0 & 0 & 0 & 10 & 9 & 6 & 2 \\ \bottomrule
\end{tabular}
    \caption{Performance of baseline tools on the original vs. a modified version of the Sturm-MBO benchmark.}
    \label{tab:ablate_preprocessing}
\end{table}
Table~\ref{tab:ablate_preprocessing} compares the performances of all baseline tools on the original version of the Sturm-MBO benchmark vs. the modified version.
In the modified version (``Interval''), we replace the search for a root of the polynomial with the search for assignments that yield a positive (negative) result.
If both assignments can be found, the existence of the root has been shown.
Interval iteration could be used to find tighter bounds on this input.

All tools perform similar or slightly worse on the interval version of the benchmark, so in the main paper, we always report their performance on the original benchmark and only run \oursolver/ on the modified version.

\subsection{LLM Code Generation Prompts}
\label{sec:llm_code_prompt}
Here, we provide the inputs and outputs to and from the LLM verbatim.
Note that the initial prompt is identical for both benchmarks, except for the two provided examples.
The samples were chosen to be small enough not to exceed the LLM character limit, but large enough to clearly demonstrate the underlying patterns.
Future work could evaluate how to determine the ideal example instances.

\subsubsection{Kissing}
\paragraph{Initial Query}
\begin{spverbatim}
You are an expert at writing highly efficient PyTorch code.

Please write efficient Python code using PyTorch to solve the following problem:
As an input, you receive a list of `K` formulas. Each formula is given as a nested list that represents a tree. The first element of inner nodes is the operation ("+", "-", "*" or "/") and which is supposed to be applied to the following elements. The leafs are strings or floating point numbers. Strings refer to variable names. You also receive an input tensor `x` of shape `[batch_size, N]`, where `N` is the number of variables across all formulas. Finally, you receive a list `inputs` that contains the names of all variables across all formulas. The order of entries maps the variable names to their indices in `x`.
As an output, you should return a tensor of shape `[batch_size, K]`, with the corresponding results for each formula.

Important: Your computation must be highly efficient! The code can utilize a GPU, so you should use this to parallelize it as much as possible. You must not use for-loops for any tensor related operation. Instead, use tensor operations to perform all required operations in parallel. Do not rely on JIT compilation for speedups, but detect patterns in the operations so you can use matrix-matrix operations and other torch features to parallelize them. If formulas depend on multiple copies of some sub-term `a`, do not recompute `a` for each instance. Instead, compute it only once and then use broadcasting or advanced indexing to duplicate the result.
Try to merge the operations in different branches of the tree: If multiple branches perform the same type of operation, compute them in parallel. E.g. instead of writing `a = x[:, 0]; b = x[:, 1]; a_squared = a ** 2; b_squared = b ** 2`, write `squared = x[:, :2] ** 2; a_squared = squared[:, 0]; b_squared = squared[:, 1]`. This merges the squaring of `a` and `b`.

Your code doesn't have to parse arbitrary input formulas. Below, you will see an example input. Extrapolate from this to detect the underlying patterns. You do not have to account for any kind of input that doesn't exhibit the same patterns. However, your code must not hard code any of these values, but be usable for any other input with similar structure.

Example:
[
['-', ['+', ['*', 'x_0_0', 'x_0_0'], ['+', ['*', 'x_0_1', 'x_0_1'], ['+', ['*', 'x_0_2', 'x_0_2'], ['*', 'x_0_3', 'x_0_3']]]], 1.0],
['-', ['+', ['*', 'x_1_0', 'x_1_0'], ['+', ['*', 'x_1_1', 'x_1_1'], ['+', ['*', 'x_1_2', 'x_1_2'], ['*', 'x_1_3', 'x_1_3']]]], 1.0],
['-', ['+', ['*', 'x_2_0', 'x_2_0'], ['+', ['*', 'x_2_1', 'x_2_1'], ['+', ['*', 'x_2_2', 'x_2_2'], ['*', 'x_2_3', 'x_2_3']]]], 1.0],
['-', ['+', ['*', 'x_3_0', 'x_3_0'], ['+', ['*', 'x_3_1', 'x_3_1'], ['+', ['*', 'x_3_2', 'x_3_2'], ['*', 'x_3_3', 'x_3_3']]]], 1.0],
['-', ['+', ['*', 'x_4_0', 'x_4_0'], ['+', ['*', 'x_4_1', 'x_4_1'], ['+', ['*', 'x_4_2', 'x_4_2'], ['*', 'x_4_3', 'x_4_3']]]], 1.0],
['-', ['+', ['*', 'x_5_0', 'x_5_0'], ['+', ['*', 'x_5_1', 'x_5_1'], ['+', ['*', 'x_5_2', 'x_5_2'], ['*', 'x_5_3', 'x_5_3']]]], 1.0],
['-', ['+', ['*', 'x_6_0', 'x_6_0'], ['+', ['*', 'x_6_1', 'x_6_1'], ['+', ['*', 'x_6_2', 'x_6_2'], ['*', 'x_6_3', 'x_6_3']]]], 1.0],
['-', ['+', ['*', 'x_7_0', 'x_7_0'], ['+', ['*', 'x_7_1', 'x_7_1'], ['+', ['*', 'x_7_2', 'x_7_2'], ['*', 'x_7_3', 'x_7_3']]]], 1.0],
['-', ['+', ['*', 'x_8_0', 'x_8_0'], ['+', ['*', 'x_8_1', 'x_8_1'], ['+', ['*', 'x_8_2', 'x_8_2'], ['*', 'x_8_3', 'x_8_3']]]], 1.0],
['-', 1.0, ['+', ['*', ['-', 'x_0_0', 'x_1_0'], ['-', 'x_0_0', 'x_1_0']], ['+', ['*', ['-', 'x_0_1', 'x_1_1'], ['-', 'x_0_1', 'x_1_1']], ['+', ['*', ['-', 'x_0_2', 'x_1_2'], ['-', 'x_0_2', 'x_1_2']], ['*', ['-', 'x_0_3', 'x_1_3'], ['-', 'x_0_3', 'x_1_3']]]]]],
['-', 1.0, ['+', ['*', ['-', 'x_0_0', 'x_2_0'], ['-', 'x_0_0', 'x_2_0']], ['+', ['*', ['-', 'x_0_1', 'x_2_1'], ['-', 'x_0_1', 'x_2_1']], ['+', ['*', ['-', 'x_0_2', 'x_2_2'], ['-', 'x_0_2', 'x_2_2']], ['*', ['-', 'x_0_3', 'x_2_3'], ['-', 'x_0_3', 'x_2_3']]]]]],
['-', 1.0, ['+', ['*', ['-', 'x_0_0', 'x_3_0'], ['-', 'x_0_0', 'x_3_0']], ['+', ['*', ['-', 'x_0_1', 'x_3_1'], ['-', 'x_0_1', 'x_3_1']], ['+', ['*', ['-', 'x_0_2', 'x_3_2'], ['-', 'x_0_2', 'x_3_2']], ['*', ['-', 'x_0_3', 'x_3_3'], ['-', 'x_0_3', 'x_3_3']]]]]],
['-', 1.0, ['+', ['*', ['-', 'x_0_0', 'x_4_0'], ['-', 'x_0_0', 'x_4_0']], ['+', ['*', ['-', 'x_0_1', 'x_4_1'], ['-', 'x_0_1', 'x_4_1']], ['+', ['*', ['-', 'x_0_2', 'x_4_2'], ['-', 'x_0_2', 'x_4_2']], ['*', ['-', 'x_0_3', 'x_4_3'], ['-', 'x_0_3', 'x_4_3']]]]]],
['-', 1.0, ['+', ['*', ['-', 'x_0_0', 'x_5_0'], ['-', 'x_0_0', 'x_5_0']], ['+', ['*', ['-', 'x_0_1', 'x_5_1'], ['-', 'x_0_1', 'x_5_1']], ['+', ['*', ['-', 'x_0_2', 'x_5_2'], ['-', 'x_0_2', 'x_5_2']], ['*', ['-', 'x_0_3', 'x_5_3'], ['-', 'x_0_3', 'x_5_3']]]]]],
['-', 1.0, ['+', ['*', ['-', 'x_0_0', 'x_6_0'], ['-', 'x_0_0', 'x_6_0']], ['+', ['*', ['-', 'x_0_1', 'x_6_1'], ['-', 'x_0_1', 'x_6_1']], ['+', ['*', ['-', 'x_0_2', 'x_6_2'], ['-', 'x_0_2', 'x_6_2']], ['*', ['-', 'x_0_3', 'x_6_3'], ['-', 'x_0_3', 'x_6_3']]]]]],
['-', 1.0, ['+', ['*', ['-', 'x_0_0', 'x_7_0'], ['-', 'x_0_0', 'x_7_0']], ['+', ['*', ['-', 'x_0_1', 'x_7_1'], ['-', 'x_0_1', 'x_7_1']], ['+', ['*', ['-', 'x_0_2', 'x_7_2'], ['-', 'x_0_2', 'x_7_2']], ['*', ['-', 'x_0_3', 'x_7_3'], ['-', 'x_0_3', 'x_7_3']]]]]],
['-', 1.0, ['+', ['*', ['-', 'x_0_0', 'x_8_0'], ['-', 'x_0_0', 'x_8_0']], ['+', ['*', ['-', 'x_0_1', 'x_8_1'], ['-', 'x_0_1', 'x_8_1']], ['+', ['*', ['-', 'x_0_2', 'x_8_2'], ['-', 'x_0_2', 'x_8_2']], ['*', ['-', 'x_0_3', 'x_8_3'], ['-', 'x_0_3', 'x_8_3']]]]]],
['-', 1.0, ['+', ['*', ['-', 'x_1_0', 'x_2_0'], ['-', 'x_1_0', 'x_2_0']], ['+', ['*', ['-', 'x_1_1', 'x_2_1'], ['-', 'x_1_1', 'x_2_1']], ['+', ['*', ['-', 'x_1_2', 'x_2_2'], ['-', 'x_1_2', 'x_2_2']], ['*', ['-', 'x_1_3', 'x_2_3'], ['-', 'x_1_3', 'x_2_3']]]]]],
['-', 1.0, ['+', ['*', ['-', 'x_1_0', 'x_3_0'], ['-', 'x_1_0', 'x_3_0']], ['+', ['*', ['-', 'x_1_1', 'x_3_1'], ['-', 'x_1_1', 'x_3_1']], ['+', ['*', ['-', 'x_1_2', 'x_3_2'], ['-', 'x_1_2', 'x_3_2']], ['*', ['-', 'x_1_3', 'x_3_3'], ['-', 'x_1_3', 'x_3_3']]]]]],
['-', 1.0, ['+', ['*', ['-', 'x_1_0', 'x_4_0'], ['-', 'x_1_0', 'x_4_0']], ['+', ['*', ['-', 'x_1_1', 'x_4_1'], ['-', 'x_1_1', 'x_4_1']], ['+', ['*', ['-', 'x_1_2', 'x_4_2'], ['-', 'x_1_2', 'x_4_2']], ['*', ['-', 'x_1_3', 'x_4_3'], ['-', 'x_1_3', 'x_4_3']]]]]],
['-', 1.0, ['+', ['*', ['-', 'x_1_0', 'x_5_0'], ['-', 'x_1_0', 'x_5_0']], ['+', ['*', ['-', 'x_1_1', 'x_5_1'], ['-', 'x_1_1', 'x_5_1']], ['+', ['*', ['-', 'x_1_2', 'x_5_2'], ['-', 'x_1_2', 'x_5_2']], ['*', ['-', 'x_1_3', 'x_5_3'], ['-', 'x_1_3', 'x_5_3']]]]]],
['-', 1.0, ['+', ['*', ['-', 'x_1_0', 'x_6_0'], ['-', 'x_1_0', 'x_6_0']], ['+', ['*', ['-', 'x_1_1', 'x_6_1'], ['-', 'x_1_1', 'x_6_1']], ['+', ['*', ['-', 'x_1_2', 'x_6_2'], ['-', 'x_1_2', 'x_6_2']], ['*', ['-', 'x_1_3', 'x_6_3'], ['-', 'x_1_3', 'x_6_3']]]]]],
['-', 1.0, ['+', ['*', ['-', 'x_1_0', 'x_7_0'], ['-', 'x_1_0', 'x_7_0']], ['+', ['*', ['-', 'x_1_1', 'x_7_1'], ['-', 'x_1_1', 'x_7_1']], ['+', ['*', ['-', 'x_1_2', 'x_7_2'], ['-', 'x_1_2', 'x_7_2']], ['*', ['-', 'x_1_3', 'x_7_3'], ['-', 'x_1_3', 'x_7_3']]]]]],
['-', 1.0, ['+', ['*', ['-', 'x_1_0', 'x_8_0'], ['-', 'x_1_0', 'x_8_0']], ['+', ['*', ['-', 'x_1_1', 'x_8_1'], ['-', 'x_1_1', 'x_8_1']], ['+', ['*', ['-', 'x_1_2', 'x_8_2'], ['-', 'x_1_2', 'x_8_2']], ['*', ['-', 'x_1_3', 'x_8_3'], ['-', 'x_1_3', 'x_8_3']]]]]],
['-', 1.0, ['+', ['*', ['-', 'x_2_0', 'x_3_0'], ['-', 'x_2_0', 'x_3_0']], ['+', ['*', ['-', 'x_2_1', 'x_3_1'], ['-', 'x_2_1', 'x_3_1']], ['+', ['*', ['-', 'x_2_2', 'x_3_2'], ['-', 'x_2_2', 'x_3_2']], ['*', ['-', 'x_2_3', 'x_3_3'], ['-', 'x_2_3', 'x_3_3']]]]]],
['-', 1.0, ['+', ['*', ['-', 'x_2_0', 'x_4_0'], ['-', 'x_2_0', 'x_4_0']], ['+', ['*', ['-', 'x_2_1', 'x_4_1'], ['-', 'x_2_1', 'x_4_1']], ['+', ['*', ['-', 'x_2_2', 'x_4_2'], ['-', 'x_2_2', 'x_4_2']], ['*', ['-', 'x_2_3', 'x_4_3'], ['-', 'x_2_3', 'x_4_3']]]]]],
['-', 1.0, ['+', ['*', ['-', 'x_2_0', 'x_5_0'], ['-', 'x_2_0', 'x_5_0']], ['+', ['*', ['-', 'x_2_1', 'x_5_1'], ['-', 'x_2_1', 'x_5_1']], ['+', ['*', ['-', 'x_2_2', 'x_5_2'], ['-', 'x_2_2', 'x_5_2']], ['*', ['-', 'x_2_3', 'x_5_3'], ['-', 'x_2_3', 'x_5_3']]]]]],
['-', 1.0, ['+', ['*', ['-', 'x_2_0', 'x_6_0'], ['-', 'x_2_0', 'x_6_0']], ['+', ['*', ['-', 'x_2_1', 'x_6_1'], ['-', 'x_2_1', 'x_6_1']], ['+', ['*', ['-', 'x_2_2', 'x_6_2'], ['-', 'x_2_2', 'x_6_2']], ['*', ['-', 'x_2_3', 'x_6_3'], ['-', 'x_2_3', 'x_6_3']]]]]],
['-', 1.0, ['+', ['*', ['-', 'x_2_0', 'x_7_0'], ['-', 'x_2_0', 'x_7_0']], ['+', ['*', ['-', 'x_2_1', 'x_7_1'], ['-', 'x_2_1', 'x_7_1']], ['+', ['*', ['-', 'x_2_2', 'x_7_2'], ['-', 'x_2_2', 'x_7_2']], ['*', ['-', 'x_2_3', 'x_7_3'], ['-', 'x_2_3', 'x_7_3']]]]]],
['-', 1.0, ['+', ['*', ['-', 'x_2_0', 'x_8_0'], ['-', 'x_2_0', 'x_8_0']], ['+', ['*', ['-', 'x_2_1', 'x_8_1'], ['-', 'x_2_1', 'x_8_1']], ['+', ['*', ['-', 'x_2_2', 'x_8_2'], ['-', 'x_2_2', 'x_8_2']], ['*', ['-', 'x_2_3', 'x_8_3'], ['-', 'x_2_3', 'x_8_3']]]]]],
['-', 1.0, ['+', ['*', ['-', 'x_3_0', 'x_4_0'], ['-', 'x_3_0', 'x_4_0']], ['+', ['*', ['-', 'x_3_1', 'x_4_1'], ['-', 'x_3_1', 'x_4_1']], ['+', ['*', ['-', 'x_3_2', 'x_4_2'], ['-', 'x_3_2', 'x_4_2']], ['*', ['-', 'x_3_3', 'x_4_3'], ['-', 'x_3_3', 'x_4_3']]]]]],
['-', 1.0, ['+', ['*', ['-', 'x_3_0', 'x_5_0'], ['-', 'x_3_0', 'x_5_0']], ['+', ['*', ['-', 'x_3_1', 'x_5_1'], ['-', 'x_3_1', 'x_5_1']], ['+', ['*', ['-', 'x_3_2', 'x_5_2'], ['-', 'x_3_2', 'x_5_2']], ['*', ['-', 'x_3_3', 'x_5_3'], ['-', 'x_3_3', 'x_5_3']]]]]],
['-', 1.0, ['+', ['*', ['-', 'x_3_0', 'x_6_0'], ['-', 'x_3_0', 'x_6_0']], ['+', ['*', ['-', 'x_3_1', 'x_6_1'], ['-', 'x_3_1', 'x_6_1']], ['+', ['*', ['-', 'x_3_2', 'x_6_2'], ['-', 'x_3_2', 'x_6_2']], ['*', ['-', 'x_3_3', 'x_6_3'], ['-', 'x_3_3', 'x_6_3']]]]]],
['-', 1.0, ['+', ['*', ['-', 'x_3_0', 'x_7_0'], ['-', 'x_3_0', 'x_7_0']], ['+', ['*', ['-', 'x_3_1', 'x_7_1'], ['-', 'x_3_1', 'x_7_1']], ['+', ['*', ['-', 'x_3_2', 'x_7_2'], ['-', 'x_3_2', 'x_7_2']], ['*', ['-', 'x_3_3', 'x_7_3'], ['-', 'x_3_3', 'x_7_3']]]]]],
['-', 1.0, ['+', ['*', ['-', 'x_3_0', 'x_8_0'], ['-', 'x_3_0', 'x_8_0']], ['+', ['*', ['-', 'x_3_1', 'x_8_1'], ['-', 'x_3_1', 'x_8_1']], ['+', ['*', ['-', 'x_3_2', 'x_8_2'], ['-', 'x_3_2', 'x_8_2']], ['*', ['-', 'x_3_3', 'x_8_3'], ['-', 'x_3_3', 'x_8_3']]]]]],
['-', 1.0, ['+', ['*', ['-', 'x_4_0', 'x_5_0'], ['-', 'x_4_0', 'x_5_0']], ['+', ['*', ['-', 'x_4_1', 'x_5_1'], ['-', 'x_4_1', 'x_5_1']], ['+', ['*', ['-', 'x_4_2', 'x_5_2'], ['-', 'x_4_2', 'x_5_2']], ['*', ['-', 'x_4_3', 'x_5_3'], ['-', 'x_4_3', 'x_5_3']]]]]],
['-', 1.0, ['+', ['*', ['-', 'x_4_0', 'x_6_0'], ['-', 'x_4_0', 'x_6_0']], ['+', ['*', ['-', 'x_4_1', 'x_6_1'], ['-', 'x_4_1', 'x_6_1']], ['+', ['*', ['-', 'x_4_2', 'x_6_2'], ['-', 'x_4_2', 'x_6_2']], ['*', ['-', 'x_4_3', 'x_6_3'], ['-', 'x_4_3', 'x_6_3']]]]]],
['-', 1.0, ['+', ['*', ['-', 'x_4_0', 'x_7_0'], ['-', 'x_4_0', 'x_7_0']], ['+', ['*', ['-', 'x_4_1', 'x_7_1'], ['-', 'x_4_1', 'x_7_1']], ['+', ['*', ['-', 'x_4_2', 'x_7_2'], ['-', 'x_4_2', 'x_7_2']], ['*', ['-', 'x_4_3', 'x_7_3'], ['-', 'x_4_3', 'x_7_3']]]]]],
['-', 1.0, ['+', ['*', ['-', 'x_4_0', 'x_8_0'], ['-', 'x_4_0', 'x_8_0']], ['+', ['*', ['-', 'x_4_1', 'x_8_1'], ['-', 'x_4_1', 'x_8_1']], ['+', ['*', ['-', 'x_4_2', 'x_8_2'], ['-', 'x_4_2', 'x_8_2']], ['*', ['-', 'x_4_3', 'x_8_3'], ['-', 'x_4_3', 'x_8_3']]]]]],
['-', 1.0, ['+', ['*', ['-', 'x_5_0', 'x_6_0'], ['-', 'x_5_0', 'x_6_0']], ['+', ['*', ['-', 'x_5_1', 'x_6_1'], ['-', 'x_5_1', 'x_6_1']], ['+', ['*', ['-', 'x_5_2', 'x_6_2'], ['-', 'x_5_2', 'x_6_2']], ['*', ['-', 'x_5_3', 'x_6_3'], ['-', 'x_5_3', 'x_6_3']]]]]],
['-', 1.0, ['+', ['*', ['-', 'x_5_0', 'x_7_0'], ['-', 'x_5_0', 'x_7_0']], ['+', ['*', ['-', 'x_5_1', 'x_7_1'], ['-', 'x_5_1', 'x_7_1']], ['+', ['*', ['-', 'x_5_2', 'x_7_2'], ['-', 'x_5_2', 'x_7_2']], ['*', ['-', 'x_5_3', 'x_7_3'], ['-', 'x_5_3', 'x_7_3']]]]]],
['-', 1.0, ['+', ['*', ['-', 'x_5_0', 'x_8_0'], ['-', 'x_5_0', 'x_8_0']], ['+', ['*', ['-', 'x_5_1', 'x_8_1'], ['-', 'x_5_1', 'x_8_1']], ['+', ['*', ['-', 'x_5_2', 'x_8_2'], ['-', 'x_5_2', 'x_8_2']], ['*', ['-', 'x_5_3', 'x_8_3'], ['-', 'x_5_3', 'x_8_3']]]]]],
['-', 1.0, ['+', ['*', ['-', 'x_6_0', 'x_7_0'], ['-', 'x_6_0', 'x_7_0']], ['+', ['*', ['-', 'x_6_1', 'x_7_1'], ['-', 'x_6_1', 'x_7_1']], ['+', ['*', ['-', 'x_6_2', 'x_7_2'], ['-', 'x_6_2', 'x_7_2']], ['*', ['-', 'x_6_3', 'x_7_3'], ['-', 'x_6_3', 'x_7_3']]]]]],
['-', 1.0, ['+', ['*', ['-', 'x_6_0', 'x_8_0'], ['-', 'x_6_0', 'x_8_0']], ['+', ['*', ['-', 'x_6_1', 'x_8_1'], ['-', 'x_6_1', 'x_8_1']], ['+', ['*', ['-', 'x_6_2', 'x_8_2'], ['-', 'x_6_2', 'x_8_2']], ['*', ['-', 'x_6_3', 'x_8_3'], ['-', 'x_6_3', 'x_8_3']]]]]],
['-', 1.0, ['+', ['*', ['-', 'x_7_0', 'x_8_0'], ['-', 'x_7_0', 'x_8_0']], ['+', ['*', ['-', 'x_7_1', 'x_8_1'], ['-', 'x_7_1', 'x_8_1']], ['+', ['*', ['-', 'x_7_2', 'x_8_2'], ['-', 'x_7_2', 'x_8_2']], ['*', ['-', 'x_7_3', 'x_8_3'], ['-', 'x_7_3', 'x_8_3']]]]]]
]

Another example:
[
['-', ['+', ['*', 'x_0_0', 'x_0_0'], ['+', ['*', 'x_0_1', 'x_0_1'], ['*', 'x_0_2', 'x_0_2']]], 1.0],
['-', ['+', ['*', 'x_1_0', 'x_1_0'], ['+', ['*', 'x_1_1', 'x_1_1'], ['*', 'x_1_2', 'x_1_2']]], 1.0],
['-', ['+', ['*', 'x_2_0', 'x_2_0'], ['+', ['*', 'x_2_1', 'x_2_1'], ['*', 'x_2_2', 'x_2_2']]], 1.0],
['-', ['+', ['*', 'x_3_0', 'x_3_0'], ['+', ['*', 'x_3_1', 'x_3_1'], ['*', 'x_3_2', 'x_3_2']]], 1.0],
['-', ['+', ['*', 'x_4_0', 'x_4_0'], ['+', ['*', 'x_4_1', 'x_4_1'], ['*', 'x_4_2', 'x_4_2']]], 1.0],
['-', ['+', ['*', 'x_5_0', 'x_5_0'], ['+', ['*', 'x_5_1', 'x_5_1'], ['*', 'x_5_2', 'x_5_2']]], 1.0],
['-', ['+', ['*', 'x_6_0', 'x_6_0'], ['+', ['*', 'x_6_1', 'x_6_1'], ['*', 'x_6_2', 'x_6_2']]], 1.0],
['-', ['+', ['*', 'x_7_0', 'x_7_0'], ['+', ['*', 'x_7_1', 'x_7_1'], ['*', 'x_7_2', 'x_7_2']]], 1.0],
['-', ['+', ['*', 'x_8_0', 'x_8_0'], ['+', ['*', 'x_8_1', 'x_8_1'], ['*', 'x_8_2', 'x_8_2']]], 1.0],
['-', ['+', ['*', 'x_9_0', 'x_9_0'], ['+', ['*', 'x_9_1', 'x_9_1'], ['*', 'x_9_2', 'x_9_2']]], 1.0],
['-', 1.0, ['+', ['*', ['-', 'x_0_0', 'x_1_0'], ['-', 'x_0_0', 'x_1_0']], ['+', ['*', ['-', 'x_0_1', 'x_1_1'], ['-', 'x_0_1', 'x_1_1']], ['*', ['-', 'x_0_2', 'x_1_2'], ['-', 'x_0_2', 'x_1_2']]]]],
['-', 1.0, ['+', ['*', ['-', 'x_0_0', 'x_2_0'], ['-', 'x_0_0', 'x_2_0']], ['+', ['*', ['-', 'x_0_1', 'x_2_1'], ['-', 'x_0_1', 'x_2_1']], ['*', ['-', 'x_0_2', 'x_2_2'], ['-', 'x_0_2', 'x_2_2']]]]],
['-', 1.0, ['+', ['*', ['-', 'x_0_0', 'x_3_0'], ['-', 'x_0_0', 'x_3_0']], ['+', ['*', ['-', 'x_0_1', 'x_3_1'], ['-', 'x_0_1', 'x_3_1']], ['*', ['-', 'x_0_2', 'x_3_2'], ['-', 'x_0_2', 'x_3_2']]]]],
['-', 1.0, ['+', ['*', ['-', 'x_0_0', 'x_4_0'], ['-', 'x_0_0', 'x_4_0']], ['+', ['*', ['-', 'x_0_1', 'x_4_1'], ['-', 'x_0_1', 'x_4_1']], ['*', ['-', 'x_0_2', 'x_4_2'], ['-', 'x_0_2', 'x_4_2']]]]],
['-', 1.0, ['+', ['*', ['-', 'x_0_0', 'x_5_0'], ['-', 'x_0_0', 'x_5_0']], ['+', ['*', ['-', 'x_0_1', 'x_5_1'], ['-', 'x_0_1', 'x_5_1']], ['*', ['-', 'x_0_2', 'x_5_2'], ['-', 'x_0_2', 'x_5_2']]]]],
['-', 1.0, ['+', ['*', ['-', 'x_0_0', 'x_6_0'], ['-', 'x_0_0', 'x_6_0']], ['+', ['*', ['-', 'x_0_1', 'x_6_1'], ['-', 'x_0_1', 'x_6_1']], ['*', ['-', 'x_0_2', 'x_6_2'], ['-', 'x_0_2', 'x_6_2']]]]],
['-', 1.0, ['+', ['*', ['-', 'x_0_0', 'x_7_0'], ['-', 'x_0_0', 'x_7_0']], ['+', ['*', ['-', 'x_0_1', 'x_7_1'], ['-', 'x_0_1', 'x_7_1']], ['*', ['-', 'x_0_2', 'x_7_2'], ['-', 'x_0_2', 'x_7_2']]]]],
['-', 1.0, ['+', ['*', ['-', 'x_0_0', 'x_8_0'], ['-', 'x_0_0', 'x_8_0']], ['+', ['*', ['-', 'x_0_1', 'x_8_1'], ['-', 'x_0_1', 'x_8_1']], ['*', ['-', 'x_0_2', 'x_8_2'], ['-', 'x_0_2', 'x_8_2']]]]],
['-', 1.0, ['+', ['*', ['-', 'x_0_0', 'x_9_0'], ['-', 'x_0_0', 'x_9_0']], ['+', ['*', ['-', 'x_0_1', 'x_9_1'], ['-', 'x_0_1', 'x_9_1']], ['*', ['-', 'x_0_2', 'x_9_2'], ['-', 'x_0_2', 'x_9_2']]]]],
['-', 1.0, ['+', ['*', ['-', 'x_1_0', 'x_2_0'], ['-', 'x_1_0', 'x_2_0']], ['+', ['*', ['-', 'x_1_1', 'x_2_1'], ['-', 'x_1_1', 'x_2_1']], ['*', ['-', 'x_1_2', 'x_2_2'], ['-', 'x_1_2', 'x_2_2']]]]],
['-', 1.0, ['+', ['*', ['-', 'x_1_0', 'x_3_0'], ['-', 'x_1_0', 'x_3_0']], ['+', ['*', ['-', 'x_1_1', 'x_3_1'], ['-', 'x_1_1', 'x_3_1']], ['*', ['-', 'x_1_2', 'x_3_2'], ['-', 'x_1_2', 'x_3_2']]]]],
['-', 1.0, ['+', ['*', ['-', 'x_1_0', 'x_4_0'], ['-', 'x_1_0', 'x_4_0']], ['+', ['*', ['-', 'x_1_1', 'x_4_1'], ['-', 'x_1_1', 'x_4_1']], ['*', ['-', 'x_1_2', 'x_4_2'], ['-', 'x_1_2', 'x_4_2']]]]],
['-', 1.0, ['+', ['*', ['-', 'x_1_0', 'x_5_0'], ['-', 'x_1_0', 'x_5_0']], ['+', ['*', ['-', 'x_1_1', 'x_5_1'], ['-', 'x_1_1', 'x_5_1']], ['*', ['-', 'x_1_2', 'x_5_2'], ['-', 'x_1_2', 'x_5_2']]]]],
['-', 1.0, ['+', ['*', ['-', 'x_1_0', 'x_6_0'], ['-', 'x_1_0', 'x_6_0']], ['+', ['*', ['-', 'x_1_1', 'x_6_1'], ['-', 'x_1_1', 'x_6_1']], ['*', ['-', 'x_1_2', 'x_6_2'], ['-', 'x_1_2', 'x_6_2']]]]],
['-', 1.0, ['+', ['*', ['-', 'x_1_0', 'x_7_0'], ['-', 'x_1_0', 'x_7_0']], ['+', ['*', ['-', 'x_1_1', 'x_7_1'], ['-', 'x_1_1', 'x_7_1']], ['*', ['-', 'x_1_2', 'x_7_2'], ['-', 'x_1_2', 'x_7_2']]]]],
['-', 1.0, ['+', ['*', ['-', 'x_1_0', 'x_8_0'], ['-', 'x_1_0', 'x_8_0']], ['+', ['*', ['-', 'x_1_1', 'x_8_1'], ['-', 'x_1_1', 'x_8_1']], ['*', ['-', 'x_1_2', 'x_8_2'], ['-', 'x_1_2', 'x_8_2']]]]],
['-', 1.0, ['+', ['*', ['-', 'x_1_0', 'x_9_0'], ['-', 'x_1_0', 'x_9_0']], ['+', ['*', ['-', 'x_1_1', 'x_9_1'], ['-', 'x_1_1', 'x_9_1']], ['*', ['-', 'x_1_2', 'x_9_2'], ['-', 'x_1_2', 'x_9_2']]]]],
['-', 1.0, ['+', ['*', ['-', 'x_2_0', 'x_3_0'], ['-', 'x_2_0', 'x_3_0']], ['+', ['*', ['-', 'x_2_1', 'x_3_1'], ['-', 'x_2_1', 'x_3_1']], ['*', ['-', 'x_2_2', 'x_3_2'], ['-', 'x_2_2', 'x_3_2']]]]],
['-', 1.0, ['+', ['*', ['-', 'x_2_0', 'x_4_0'], ['-', 'x_2_0', 'x_4_0']], ['+', ['*', ['-', 'x_2_1', 'x_4_1'], ['-', 'x_2_1', 'x_4_1']], ['*', ['-', 'x_2_2', 'x_4_2'], ['-', 'x_2_2', 'x_4_2']]]]],
['-', 1.0, ['+', ['*', ['-', 'x_2_0', 'x_5_0'], ['-', 'x_2_0', 'x_5_0']], ['+', ['*', ['-', 'x_2_1', 'x_5_1'], ['-', 'x_2_1', 'x_5_1']], ['*', ['-', 'x_2_2', 'x_5_2'], ['-', 'x_2_2', 'x_5_2']]]]],
['-', 1.0, ['+', ['*', ['-', 'x_2_0', 'x_6_0'], ['-', 'x_2_0', 'x_6_0']], ['+', ['*', ['-', 'x_2_1', 'x_6_1'], ['-', 'x_2_1', 'x_6_1']], ['*', ['-', 'x_2_2', 'x_6_2'], ['-', 'x_2_2', 'x_6_2']]]]],
['-', 1.0, ['+', ['*', ['-', 'x_2_0', 'x_7_0'], ['-', 'x_2_0', 'x_7_0']], ['+', ['*', ['-', 'x_2_1', 'x_7_1'], ['-', 'x_2_1', 'x_7_1']], ['*', ['-', 'x_2_2', 'x_7_2'], ['-', 'x_2_2', 'x_7_2']]]]],
['-', 1.0, ['+', ['*', ['-', 'x_2_0', 'x_8_0'], ['-', 'x_2_0', 'x_8_0']], ['+', ['*', ['-', 'x_2_1', 'x_8_1'], ['-', 'x_2_1', 'x_8_1']], ['*', ['-', 'x_2_2', 'x_8_2'], ['-', 'x_2_2', 'x_8_2']]]]],
['-', 1.0, ['+', ['*', ['-', 'x_2_0', 'x_9_0'], ['-', 'x_2_0', 'x_9_0']], ['+', ['*', ['-', 'x_2_1', 'x_9_1'], ['-', 'x_2_1', 'x_9_1']], ['*', ['-', 'x_2_2', 'x_9_2'], ['-', 'x_2_2', 'x_9_2']]]]],
['-', 1.0, ['+', ['*', ['-', 'x_3_0', 'x_4_0'], ['-', 'x_3_0', 'x_4_0']], ['+', ['*', ['-', 'x_3_1', 'x_4_1'], ['-', 'x_3_1', 'x_4_1']], ['*', ['-', 'x_3_2', 'x_4_2'], ['-', 'x_3_2', 'x_4_2']]]]],
['-', 1.0, ['+', ['*', ['-', 'x_3_0', 'x_5_0'], ['-', 'x_3_0', 'x_5_0']], ['+', ['*', ['-', 'x_3_1', 'x_5_1'], ['-', 'x_3_1', 'x_5_1']], ['*', ['-', 'x_3_2', 'x_5_2'], ['-', 'x_3_2', 'x_5_2']]]]],
['-', 1.0, ['+', ['*', ['-', 'x_3_0', 'x_6_0'], ['-', 'x_3_0', 'x_6_0']], ['+', ['*', ['-', 'x_3_1', 'x_6_1'], ['-', 'x_3_1', 'x_6_1']], ['*', ['-', 'x_3_2', 'x_6_2'], ['-', 'x_3_2', 'x_6_2']]]]],
['-', 1.0, ['+', ['*', ['-', 'x_3_0', 'x_7_0'], ['-', 'x_3_0', 'x_7_0']], ['+', ['*', ['-', 'x_3_1', 'x_7_1'], ['-', 'x_3_1', 'x_7_1']], ['*', ['-', 'x_3_2', 'x_7_2'], ['-', 'x_3_2', 'x_7_2']]]]],
['-', 1.0, ['+', ['*', ['-', 'x_3_0', 'x_8_0'], ['-', 'x_3_0', 'x_8_0']], ['+', ['*', ['-', 'x_3_1', 'x_8_1'], ['-', 'x_3_1', 'x_8_1']], ['*', ['-', 'x_3_2', 'x_8_2'], ['-', 'x_3_2', 'x_8_2']]]]],
['-', 1.0, ['+', ['*', ['-', 'x_3_0', 'x_9_0'], ['-', 'x_3_0', 'x_9_0']], ['+', ['*', ['-', 'x_3_1', 'x_9_1'], ['-', 'x_3_1', 'x_9_1']], ['*', ['-', 'x_3_2', 'x_9_2'], ['-', 'x_3_2', 'x_9_2']]]]],
['-', 1.0, ['+', ['*', ['-', 'x_4_0', 'x_5_0'], ['-', 'x_4_0', 'x_5_0']], ['+', ['*', ['-', 'x_4_1', 'x_5_1'], ['-', 'x_4_1', 'x_5_1']], ['*', ['-', 'x_4_2', 'x_5_2'], ['-', 'x_4_2', 'x_5_2']]]]],
['-', 1.0, ['+', ['*', ['-', 'x_4_0', 'x_6_0'], ['-', 'x_4_0', 'x_6_0']], ['+', ['*', ['-', 'x_4_1', 'x_6_1'], ['-', 'x_4_1', 'x_6_1']], ['*', ['-', 'x_4_2', 'x_6_2'], ['-', 'x_4_2', 'x_6_2']]]]],
['-', 1.0, ['+', ['*', ['-', 'x_4_0', 'x_7_0'], ['-', 'x_4_0', 'x_7_0']], ['+', ['*', ['-', 'x_4_1', 'x_7_1'], ['-', 'x_4_1', 'x_7_1']], ['*', ['-', 'x_4_2', 'x_7_2'], ['-', 'x_4_2', 'x_7_2']]]]],
['-', 1.0, ['+', ['*', ['-', 'x_4_0', 'x_8_0'], ['-', 'x_4_0', 'x_8_0']], ['+', ['*', ['-', 'x_4_1', 'x_8_1'], ['-', 'x_4_1', 'x_8_1']], ['*', ['-', 'x_4_2', 'x_8_2'], ['-', 'x_4_2', 'x_8_2']]]]],
['-', 1.0, ['+', ['*', ['-', 'x_4_0', 'x_9_0'], ['-', 'x_4_0', 'x_9_0']], ['+', ['*', ['-', 'x_4_1', 'x_9_1'], ['-', 'x_4_1', 'x_9_1']], ['*', ['-', 'x_4_2', 'x_9_2'], ['-', 'x_4_2', 'x_9_2']]]]],
['-', 1.0, ['+', ['*', ['-', 'x_5_0', 'x_6_0'], ['-', 'x_5_0', 'x_6_0']], ['+', ['*', ['-', 'x_5_1', 'x_6_1'], ['-', 'x_5_1', 'x_6_1']], ['*', ['-', 'x_5_2', 'x_6_2'], ['-', 'x_5_2', 'x_6_2']]]]],
['-', 1.0, ['+', ['*', ['-', 'x_5_0', 'x_7_0'], ['-', 'x_5_0', 'x_7_0']], ['+', ['*', ['-', 'x_5_1', 'x_7_1'], ['-', 'x_5_1', 'x_7_1']], ['*', ['-', 'x_5_2', 'x_7_2'], ['-', 'x_5_2', 'x_7_2']]]]],
['-', 1.0, ['+', ['*', ['-', 'x_5_0', 'x_8_0'], ['-', 'x_5_0', 'x_8_0']], ['+', ['*', ['-', 'x_5_1', 'x_8_1'], ['-', 'x_5_1', 'x_8_1']], ['*', ['-', 'x_5_2', 'x_8_2'], ['-', 'x_5_2', 'x_8_2']]]]],
['-', 1.0, ['+', ['*', ['-', 'x_5_0', 'x_9_0'], ['-', 'x_5_0', 'x_9_0']], ['+', ['*', ['-', 'x_5_1', 'x_9_1'], ['-', 'x_5_1', 'x_9_1']], ['*', ['-', 'x_5_2', 'x_9_2'], ['-', 'x_5_2', 'x_9_2']]]]],
['-', 1.0, ['+', ['*', ['-', 'x_6_0', 'x_7_0'], ['-', 'x_6_0', 'x_7_0']], ['+', ['*', ['-', 'x_6_1', 'x_7_1'], ['-', 'x_6_1', 'x_7_1']], ['*', ['-', 'x_6_2', 'x_7_2'], ['-', 'x_6_2', 'x_7_2']]]]],
['-', 1.0, ['+', ['*', ['-', 'x_6_0', 'x_8_0'], ['-', 'x_6_0', 'x_8_0']], ['+', ['*', ['-', 'x_6_1', 'x_8_1'], ['-', 'x_6_1', 'x_8_1']], ['*', ['-', 'x_6_2', 'x_8_2'], ['-', 'x_6_2', 'x_8_2']]]]],
['-', 1.0, ['+', ['*', ['-', 'x_6_0', 'x_9_0'], ['-', 'x_6_0', 'x_9_0']], ['+', ['*', ['-', 'x_6_1', 'x_9_1'], ['-', 'x_6_1', 'x_9_1']], ['*', ['-', 'x_6_2', 'x_9_2'], ['-', 'x_6_2', 'x_9_2']]]]],
['-', 1.0, ['+', ['*', ['-', 'x_7_0', 'x_8_0'], ['-', 'x_7_0', 'x_8_0']], ['+', ['*', ['-', 'x_7_1', 'x_8_1'], ['-', 'x_7_1', 'x_8_1']], ['*', ['-', 'x_7_2', 'x_8_2'], ['-', 'x_7_2', 'x_8_2']]]]],
['-', 1.0, ['+', ['*', ['-', 'x_7_0', 'x_9_0'], ['-', 'x_7_0', 'x_9_0']], ['+', ['*', ['-', 'x_7_1', 'x_9_1'], ['-', 'x_7_1', 'x_9_1']], ['*', ['-', 'x_7_2', 'x_9_2'], ['-', 'x_7_2', 'x_9_2']]]]],
['-', 1.0, ['+', ['*', ['-', 'x_8_0', 'x_9_0'], ['-', 'x_8_0', 'x_9_0']], ['+', ['*', ['-', 'x_8_1', 'x_9_1'], ['-', 'x_8_1', 'x_9_1']], ['*', ['-', 'x_8_2', 'x_9_2'], ['-', 'x_8_2', 'x_9_2']]]]]
]

Perform the following steps: First, analyze the example and detect the underlying patterns. Finally, implement the logic and provide the python code. Your code must be able to deal with negative variables, too. So do not use logarithms. Make sure to place all tensors on the same device that the x tensor is on.

Write a function `compute_result(formulas, x, inputs)` that receives the formulas, the variables x and the mapping to variable names as an input, and returns a tensor with the corresponding results.
\end{spverbatim}

\paragraph{LLM Response}
\begin{spverbatim}
Based on the patterns observed in the provided examples, the formulas calculate two types of expressions:

1. For each unique index `i`, compute \( s_i = \sum_{k} x_{i,k}^2 - 1.0 \).
2. For each pair of unique indices `(i, j)`, compute \( s_{ij} = 1.0 - \sum_{k} (x_{i,k} - x_{j,k})^2 \).

Here's the efficient PyTorch implementation:

```python
import torch

def compute_result(formulas, x, inputs):
    device = x.device
    batch_size = x.shape[0]
    
    # Build mapping from variable names to indices in x
    variable_mapping = {}
    for idx, var_name in enumerate(inputs):
        tokens = var_name.split('_')
        if len(tokens) != 3 or not tokens[1].isdigit() or not tokens[2].isdigit():
            continue  # Skip invalid variable names
        i = int(tokens[1])
        k = int(tokens[2])
        variable_mapping[var_name] = {'i': i, 'k': k, 'index_in_x': idx}
    
    # Map i to list of (k, idx_in_x)
    i_to_k_idx = {}
    for var_name, data in variable_mapping.items():
        i = data['i']
        k = data['k']
        idx_in_x = data['index_in_x']
        if i not in i_to_k_idx:
            i_to_k_idx[i] = []
        i_to_k_idx[i].append((k, idx_in_x))
    
    # Get sorted list of unique i's
    unique_is = sorted(i_to_k_idx.keys())
    num_i = len(unique_is)
    
    # Build x_i_list
    x_i_list = []
    for i in unique_is:
        k_idx_pairs = sorted(i_to_k_idx[i], key=lambda x: x[0])  # Sort by k
        idxs_for_i = [idx_in_x for k, idx_in_x in k_idx_pairs]
        x_i = x[:, idxs_for_i]  # Shape [batch_size, D]
        x_i_list.append(x_i)
    
    # Stack x_i tensors
    X = torch.stack(x_i_list, dim=0).to(device)  # Shape [num_i, batch_size, D]
    
    # Transpose to shape [batch_size, num_i, D]
    X = X.permute(1, 0, 2)  # Shape [batch_size, num_i, D]
    
    # Compute x_i squared norms
    x_i_squared_norms = (X ** 2).sum(dim=2)  # Shape [batch_size, num_i]
    
    s_i = x_i_squared_norms - 1.0  # Shape [batch_size, num_i]
    
    # Compute pairwise squared distances
    x_i_xj_dot = torch.bmm(X, X.transpose(1,2))  # Shape [batch_size, num_i, num_i]
    
    # Compute s_ij = 1.0 - ||x_i - x_j||^2
    s_ij = 1.0 - (
        x_i_squared_norms.unsqueeze(2) + x_i_squared_norms.unsqueeze(1) - 2 * x_i_xj_dot
    )  # Shape [batch_size, num_i, num_i]
    
    # Extract upper triangle indices
    indices_i, indices_j = torch.triu_indices(num_i, num_i, offset=1)
    
    # s_ij_upper has shape [batch_size, num_pairs]
    s_ij_upper = s_ij[:, indices_i, indices_j]
    
    # Concatenate s_i and s_ij_upper
    output = torch.cat([s_i, s_ij_upper], dim=1)  # Shape [batch_size, K]
    
    return output
```

This function efficiently computes the required results by:

- Parsing the input variable names to map each unique `i` to its corresponding columns in `x`.
- Stacking the vectors \( x_i \) for each unique `i` to form a tensor.
- Computing \( s_i \) and \( s_{ij} \) using vectorized operations without any for-loops.
- Ensuring that all tensors are on the same device as `x`, and handling negative values correctly by avoiding square roots or logarithms.

This code should work efficiently for inputs with similar structures to the examples provided.
\end{spverbatim}

\subsubsection{Sturm-MBO}
\paragraph{Initial Query}
\begin{spverbatim}
You are an expert at writing highly efficient PyTorch code.

Please write efficient Python code using PyTorch to solve the following problem:
As an input, you receive a list of `K` formulas. Each formula is given as a nested list that represents a tree. The first element of inner nodes is the operation ("+", "-", "*" or "/") and which is supposed to be applied to the following elements. The leafs are strings or floating point numbers. Strings refer to variable names. You also receive an input tensor `x` of shape `[batch_size, N]`, where `N` is the number of variables across all formulas. Finally, you receive a list `inputs` that contains the names of all variables across all formulas. The order of entries maps the variable names to their indices in `x`.
As an output, you should return a tensor of shape `[batch_size, K]`, with the corresponding results for each formula.

Important: Your computation must be highly efficient! The code can utilize a GPU, so you should use this to parallelize it as much as possible. You must not use for-loops for any tensor related operation. Instead, use tensor operations to perform all required operations in parallel. Do not rely on JIT compilation for speedups, but detect patterns in the operations so you can use matrix-matrix operations and other torch features to parallelize them. If formulas depend on multiple copies of some sub-term `a`, do not recompute `a` for each instance. Instead, compute it only once and then use broadcasting or advanced indexing to duplicate the result.
Try to merge the operations in different branches of the tree: If multiple branches perform the same type of operation, compute them in parallel. E.g. instead of writing `a = x[:, 0]; b = x[:, 1]; a_squared = a ** 2; b_squared = b ** 2`, write `squared = x[:, :2] ** 2; a_squared = squared[:, 0]; b_squared = squared[:, 1]`. This merges the squaring of `a` and `b`.

Your code doesn't have to parse arbitrary input formulas. Below, you will see an example input. Extrapolate from this to detect the underlying patterns. You do not have to account for any kind of input that doesn't exhibit the same patterns. However, your code must not hard code any of these values, but be usable for any other input with similar structure.

Example:
[
['-', 0.0, ['+', ['*', -1.0, ['*', 2.0, ['*', 'h1p', 'h1p', 'h1p'], ['*', 'h2p', 'h2p'], 'h3p']], ['*', 2.0, ['*', 'h1p', 'h1p', 'h1p'], ['*', 'h2p', 'h2p'], 'h5p'], ['*', -1.0, ['*', 8.0, ['*', 'h1p', 'h1p', 'h1p'], 'h2p', ['*', 'h3p', 'h3p']]], ['*', 4.0, ['*', 'h1p', 'h1p', 'h1p'], 'h2p', 'h3p', 'h4p'], ['*', 10.0, ['*', 'h1p', 'h1p', 'h1p'], 'h2p', 'h3p', 'h5p'], ['*', 10.0, ['*', 'h1p', 'h1p', 'h1p'], 'h2p', 'h3p', 'h6p'], ['*', 2.0, ['*', 'h1p', 'h1p', 'h1p'], 'h2p', 'h4p', 'h5p'], ['*', 2.0, ['*', 'h1p', 'h1p', 'h1p'], 'h2p', ['*', 'h5p', 'h5p']], ['*', 24.0, ['*', 'h1p', 'h1p', 'h1p'], ['*', 'h3p', 'h3p'], 'h4p'], ['*', 16.0, ['*', 'h1p', 'h1p', 'h1p'], ['*', 'h3p', 'h3p'], 'h5p'], ['*', 16.0, ['*', 'h1p', 'h1p', 'h1p'], ['*', 'h3p', 'h3p'], 'h6p'], ['*', 6.0, ['*', 'h1p', 'h1p', 'h1p'], 'h3p', ['*', 'h4p', 'h4p']], ['*', 6.0, ['*', 'h1p', 'h1p', 'h1p'], 'h3p', 'h4p', 'h5p'], ['*', 10.0, ['*', 'h1p', 'h1p', 'h1p'], 'h3p', 'h4p', 'h6p'], ['*', 4.0, ['*', 'h1p', 'h1p', 'h1p'], 'h3p', ['*', 'h5p', 'h5p']], ['*', 8.0, ['*', 'h1p', 'h1p', 'h1p'], 'h3p', 'h5p', 'h6p'], ['*', 4.0, ['*', 'h1p', 'h1p', 'h1p'], 'h3p', ['*', 'h6p', 'h6p']], ['*', 2.0, ['*', 'h1p', 'h1p', 'h1p'], 'h4p', 'h5p', 'h6p'], ['*', 2.0, ['*', 'h1p', 'h1p', 'h1p'], ['*', 'h5p', 'h5p'], 'h6p'], ['*', 2.0, ['*', 'h1p', 'h1p', 'h1p'], 'h5p', ['*', 'h6p', 'h6p']], ['*', -1.0, ['*', ['*', 'h1p', 'h1p'], ['*', 'h2p', 'h2p', 'h2p'], 'h3p']], ['*', ['*', 'h1p', 'h1p'], ['*', 'h2p', 'h2p', 'h2p'], 'h5p'], ['*', -1.0, ['*', 9.0, ['*', 'h1p', 'h1p'], ['*', 'h2p', 'h2p'], ['*', 'h3p', 'h3p']]], ['*', ['*', 'h1p', 'h1p'], ['*', 'h2p', 'h2p'], 'h3p', 'h4p'], ['*', 10.0, ['*', 'h1p', 'h1p'], ['*', 'h2p', 'h2p'], 'h3p', 'h5p'], ['*', 8.0, ['*', 'h1p', 'h1p'], ['*', 'h2p', 'h2p'], 'h3p', 'h6p'], ['*', 6.0, ['*', 'h1p', 'h1p'], ['*', 'h2p', 'h2p'], 'h4p', 'h5p'], ['*', 3.0, ['*', 'h1p', 'h1p'], ['*', 'h2p', 'h2p'], ['*', 'h5p', 'h5p']], ['*', 3.0, ['*', 'h1p', 'h1p'], ['*', 'h2p', 'h2p'], 'h5p', 'h6p'], ['*', -1.0, ['*', 8.0, ['*', 'h1p', 'h1p'], 'h2p', ['*', 'h3p', 'h3p', 'h3p']]], ['*', 28.0, ['*', 'h1p', 'h1p'], 'h2p', ['*', 'h3p', 'h3p'], 'h4p'], ['*', 26.0, ['*', 'h1p', 'h1p'], 'h2p', ['*', 'h3p', 'h3p'], 'h5p'], ['*', 34.0, ['*', 'h1p', 'h1p'], 'h2p', ['*', 'h3p', 'h3p'], 'h6p'], ['*', 13.0, ['*', 'h1p', 'h1p'], 'h2p', 'h3p', ['*', 'h4p', 'h4p']], ['*', 28.0, ['*', 'h1p', 'h1p'], 'h2p', 'h3p', 'h4p', 'h5p'], ['*', 30.0, ['*', 'h1p', 'h1p'], 'h2p', 'h3p', 'h4p', 'h6p'], ['*', 15.0, ['*', 'h1p', 'h1p'], 'h2p', 'h3p', ['*', 'h5p', 'h5p']], ['*', 32.0, ['*', 'h1p', 'h1p'], 'h2p', 'h3p', 'h5p', 'h6p'], ['*', 17.0, ['*', 'h1p', 'h1p'], 'h2p', 'h3p', ['*', 'h6p', 'h6p']], ['*', 3.0, ['*', 'h1p', 'h1p'], 'h2p', ['*', 'h4p', 'h4p'], 'h5p'], ['*', 6.0, ['*', 'h1p', 'h1p'], 'h2p', 'h4p', ['*', 'h5p', 'h5p']], ['*', 6.0, ['*', 'h1p', 'h1p'], 'h2p', 'h4p', 'h5p', 'h6p'], ['*', ['*', 'h1p', 'h1p'], 'h2p', ['*', 'h5p', 'h5p', 'h5p']], ['*', 4.0, ['*', 'h1p', 'h1p'], 'h2p', ['*', 'h5p', 'h5p'], 'h6p'], ['*', 3.0, ['*', 'h1p', 'h1p'], 'h2p', 'h5p', ['*', 'h6p', 'h6p']], ['*', 24.0, ['*', 'h1p', 'h1p'], ['*', 'h3p', 'h3p', 'h3p'], 'h4p'], ['*', 16.0, ['*', 'h1p', 'h1p'], ['*', 'h3p', 'h3p', 'h3p'], 'h5p'], ['*', 16.0, ['*', 'h1p', 'h1p'], ['*', 'h3p', 'h3p', 'h3p'], 'h6p'], ['*', 21.0, ['*', 'h1p', 'h1p'], ['*', 'h3p', 'h3p'], ['*', 'h4p', 'h4p']], ['*', 46.0, ['*', 'h1p', 'h1p'], ['*', 'h3p', 'h3p'], 'h4p', 'h5p'], ['*', 38.0, ['*', 'h1p', 'h1p'], ['*', 'h3p', 'h3p'], 'h4p', 'h6p'], ['*', 16.0, ['*', 'h1p', 'h1p'], ['*', 'h3p', 'h3p'], ['*', 'h5p', 'h5p']], ['*', 40.0, ['*', 'h1p', 'h1p'], ['*', 'h3p', 'h3p'], 'h5p', 'h6p'], ['*', 16.0, ['*', 'h1p', 'h1p'], ['*', 'h3p', 'h3p'], ['*', 'h6p', 'h6p']], ['*', 3.0, ['*', 'h1p', 'h1p'], 'h3p', ['*', 'h4p', 'h4p', 'h4p']], ['*', 14.0, ['*', 'h1p', 'h1p'], 'h3p', ['*', 'h4p', 'h4p'], 'h5p'], ['*', 8.0, ['*', 'h1p', 'h1p'], 'h3p', ['*', 'h4p', 'h4p'], 'h6p'], ['*', 11.0, ['*', 'h1p', 'h1p'], 'h3p', 'h4p', ['*', 'h5p', 'h5p']], ['*', 24.0, ['*', 'h1p', 'h1p'], 'h3p', 'h4p', 'h5p', 'h6p'], ['*', 7.0, ['*', 'h1p', 'h1p'], 'h3p', 'h4p', ['*', 'h6p', 'h6p']], ['*', 2.0, ['*', 'h1p', 'h1p'], 'h3p', ['*', 'h5p', 'h5p', 'h5p']], ['*', 12.0, ['*', 'h1p', 'h1p'], 'h3p', ['*', 'h5p', 'h5p'], 'h6p'], ['*', 12.0, ['*', 'h1p', 'h1p'], 'h3p', 'h5p', ['*', 'h6p', 'h6p']], ['*', 2.0, ['*', 'h1p', 'h1p'], 'h3p', ['*', 'h6p', 'h6p', 'h6p']], ['*', ['*', 'h1p', 'h1p'], ['*', 'h4p', 'h4p'], 'h5p', 'h6p'], ['*', 4.0, ['*', 'h1p', 'h1p'], 'h4p', ['*', 'h5p', 'h5p'], 'h6p'], ['*', 2.0, ['*', 'h1p', 'h1p'], 'h4p', 'h5p', ['*', 'h6p', 'h6p']], ['*', ['*', 'h1p', 'h1p'], ['*', 'h5p', 'h5p', 'h5p'], 'h6p'], ['*', 3.0, ['*', 'h1p', 'h1p'], ['*', 'h5p', 'h5p'], ['*', 'h6p', 'h6p']], ['*', ['*', 'h1p', 'h1p'], 'h5p', ['*', 'h6p', 'h6p', 'h6p']], ['*', -1.0, ['*', 'h1p', ['*', 'h2p', 'h2p', 'h2p'], ['*', 'h3p', 'h3p']]], ['*', 'h1p', ['*', 'h2p', 'h2p', 'h2p'], 'h3p', 'h5p'], ['*', 2.0, 'h1p', ['*', 'h2p', 'h2p', 'h2p'], 'h3p', 'h6p'], ['*', 2.0, 'h1p', ['*', 'h2p', 'h2p', 'h2p'], 'h4p', 'h5p'], ['*', 'h1p', ['*', 'h2p', 'h2p', 'h2p'], ['*', 'h5p', 'h5p']], ['*', 'h1p', ['*', 'h2p', 'h2p', 'h2p'], 'h5p', 'h6p'], ['*', -1.0, ['*', 2.0, 'h1p', ['*', 'h2p', 'h2p'], ['*', 'h3p', 'h3p', 'h3p']]], ['*', 8.0, 'h1p', ['*', 'h2p', 'h2p'], ['*', 'h3p', 'h3p'], 'h4p'], ['*', 9.0, 'h1p', ['*', 'h2p', 'h2p'], ['*', 'h3p', 'h3p'], 'h5p'], ['*', 15.0, 'h1p', ['*', 'h2p', 'h2p'], ['*', 'h3p', 'h3p'], 'h6p'], ['*', 6.0, 'h1p', ['*', 'h2p', 'h2p'], 'h3p', ['*', 'h4p', 'h4p']], ['*', 20.0, 'h1p', ['*', 'h2p', 'h2p'], 'h3p', 'h4p', 'h5p'], ['*', 15.0, 'h1p', ['*', 'h2p', 'h2p'], 'h3p', 'h4p', 'h6p'], ['*', 8.0, 'h1p', ['*', 'h2p', 'h2p'], 'h3p', ['*', 'h5p', 'h5p']], ['*', 19.0, 'h1p', ['*', 'h2p', 'h2p'], 'h3p', 'h5p', 'h6p'], ['*', 9.0, 'h1p', ['*', 'h2p', 'h2p'], 'h3p', ['*', 'h6p', 'h6p']], ['*', 5.0, 'h1p', ['*', 'h2p', 'h2p'], ['*', 'h4p', 'h4p'], 'h5p'], ['*', 5.0, 'h1p', ['*', 'h2p', 'h2p'], 'h4p', ['*', 'h5p', 'h5p']], ['*', 9.0, 'h1p', ['*', 'h2p', 'h2p'], 'h4p', 'h5p', 'h6p'], ['*', 'h1p', ['*', 'h2p', 'h2p'], ['*', 'h5p', 'h5p', 'h5p']], ['*', 4.0, 'h1p', ['*', 'h2p', 'h2p'], ['*', 'h5p', 'h5p'], 'h6p'], ['*', 4.0, 'h1p', ['*', 'h2p', 'h2p'], 'h5p', ['*', 'h6p', 'h6p']], ['*', 10.0, 'h1p', 'h2p', ['*', 'h3p', 'h3p', 'h3p'], 'h4p'], ['*', 8.0, 'h1p', 'h2p', ['*', 'h3p', 'h3p', 'h3p'], 'h5p'], ['*', 16.0, 'h1p', 'h2p', ['*', 'h3p', 'h3p', 'h3p'], 'h6p'], ['*', 17.0, 'h1p', 'h2p', ['*', 'h3p', 'h3p'], ['*', 'h4p', 'h4p']], ['*', 49.0, 'h1p', 'h2p', ['*', 'h3p', 'h3p'], 'h4p', 'h5p'], ['*', 39.0, 'h1p', 'h2p', ['*', 'h3p', 'h3p'], 'h4p', 'h6p'], ['*', 18.0, 'h1p', 'h2p', ['*', 'h3p', 'h3p'], ['*', 'h5p', 'h5p']], ['*', 48.0, 'h1p', 'h2p', ['*', 'h3p', 'h3p'], 'h5p', 'h6p'], ['*', 22.0, 'h1p', 'h2p', ['*', 'h3p', 'h3p'], ['*', 'h6p', 'h6p']], ['*', 4.0, 'h1p', 'h2p', 'h3p', ['*', 'h4p', 'h4p', 'h4p']], ['*', 23.0, 'h1p', 'h2p', 'h3p', ['*', 'h4p', 'h4p'], 'h5p'], ['*', 13.0, 'h1p', 'h2p', 'h3p', ['*', 'h4p', 'h4p'], 'h6p'], ['*', 23.0, 'h1p', 'h2p', 'h3p', 'h4p', ['*', 'h5p', 'h5p']], ['*', 46.0, 'h1p', 'h2p', 'h3p', 'h4p', 'h5p', 'h6p'], ['*', 14.0, 'h1p', 'h2p', 'h3p', 'h4p', ['*', 'h6p', 'h6p']], ['*', 5.0, 'h1p', 'h2p', 'h3p', ['*', 'h5p', 'h5p', 'h5p']], ['*', 21.0, 'h1p', 'h2p', 'h3p', ['*', 'h5p', 'h5p'], 'h6p'], ['*', 23.0, 'h1p', 'h2p', 'h3p', 'h5p', ['*', 'h6p', 'h6p']], ['*', 5.0, 'h1p', 'h2p', 'h3p', ['*', 'h6p', 'h6p', 'h6p']], ['*', 'h1p', 'h2p', ['*', 'h4p', 'h4p', 'h4p'], 'h5p'], ['*', 5.0, 'h1p', 'h2p', ['*', 'h4p', 'h4p'], ['*', 'h5p', 'h5p']], ['*', 3.0, 'h1p', 'h2p', ['*', 'h4p', 'h4p'], 'h5p', 'h6p'], ['*', 2.0, 'h1p', 'h2p', 'h4p', ['*', 'h5p', 'h5p', 'h5p']], ['*', 9.0, 'h1p', 'h2p', 'h4p', ['*', 'h5p', 'h5p'], 'h6p'], ['*', 3.0, 'h1p', 'h2p', 'h4p', 'h5p', ['*', 'h6p', 'h6p']], ['*', 2.0, 'h1p', 'h2p', ['*', 'h5p', 'h5p', 'h5p'], 'h6p'], ['*', 4.0, 'h1p', 'h2p', ['*', 'h5p', 'h5p'], ['*', 'h6p', 'h6p']], ['*', 'h1p', 'h2p', 'h5p', ['*', 'h6p', 'h6p', 'h6p']], ['*', 12.0, 'h1p', ['*', 'h3p', 'h3p', 'h3p'], ['*', 'h4p', 'h4p']], ['*', 32.0, 'h1p', ['*', 'h3p', 'h3p', 'h3p'], 'h4p', 'h5p'], ['*', 20.0, 'h1p', ['*', 'h3p', 'h3p', 'h3p'], 'h4p', 'h6p'], ['*', 8.0, 'h1p', ['*', 'h3p', 'h3p', 'h3p'], ['*', 'h5p', 'h5p']], ['*', 24.0, 'h1p', ['*', 'h3p', 'h3p', 'h3p'], 'h5p', 'h6p'], ['*', 8.0, 'h1p', ['*', 'h3p', 'h3p', 'h3p'], ['*', 'h6p', 'h6p']], ['*', 6.0, 'h1p', ['*', 'h3p', 'h3p'], ['*', 'h4p', 'h4p', 'h4p']], ['*', 26.0, 'h1p', ['*', 'h3p', 'h3p'], ['*', 'h4p', 'h4p'], 'h5p'], ['*', 16.0, 'h1p', ['*', 'h3p', 'h3p'], ['*', 'h4p', 'h4p'], 'h6p'], ['*', 26.0, 'h1p', ['*', 'h3p', 'h3p'], 'h4p', ['*', 'h5p', 'h5p']], ['*', 50.0, 'h1p', ['*', 'h3p', 'h3p'], 'h4p', 'h5p', 'h6p'], ['*', 14.0, 'h1p', ['*', 'h3p', 'h3p'], 'h4p', ['*', 'h6p', 'h6p']], ['*', 4.0, 'h1p', ['*', 'h3p', 'h3p'], ['*', 'h5p', 'h5p', 'h5p']], ['*', 24.0, 'h1p', ['*', 'h3p', 'h3p'], ['*', 'h5p', 'h5p'], 'h6p'], ['*', 24.0, 'h1p', ['*', 'h3p', 'h3p'], 'h5p', ['*', 'h6p', 'h6p']], ['*', 4.0, 'h1p', ['*', 'h3p', 'h3p'], ['*', 'h6p', 'h6p', 'h6p']], ['*', 4.0, 'h1p', 'h3p', ['*', 'h4p', 'h4p', 'h4p'], 'h5p'], ['*', 11.0, 'h1p', 'h3p', ['*', 'h4p', 'h4p'], ['*', 'h5p', 'h5p']], ['*', 12.0, 'h1p', 'h3p', ['*', 'h4p', 'h4p'], 'h5p', 'h6p'], ['*', 3.0, 'h1p', 'h3p', 'h4p', ['*', 'h5p', 'h5p', 'h5p']], ['*', 21.0, 'h1p', 'h3p', 'h4p', ['*', 'h5p', 'h5p'], 'h6p'], ['*', 12.0, 'h1p', 'h3p', 'h4p', 'h5p', ['*', 'h6p', 'h6p']], ['*', 4.0, 'h1p', 'h3p', ['*', 'h5p', 'h5p', 'h5p'], 'h6p'], ['*', 10.0, 'h1p', 'h3p', ['*', 'h5p', 'h5p'], ['*', 'h6p', 'h6p']], ['*', 4.0, 'h1p', 'h3p', 'h5p', ['*', 'h6p', 'h6p', 'h6p']], ['*', 'h1p', ['*', 'h4p', 'h4p'], ['*', 'h5p', 'h5p'], 'h6p'], ['*', 'h1p', 'h4p', ['*', 'h5p', 'h5p', 'h5p'], 'h6p'], ['*', 2.0, 'h1p', 'h4p', ['*', 'h5p', 'h5p'], ['*', 'h6p', 'h6p']], ['*', 'h1p', ['*', 'h5p', 'h5p', 'h5p'], ['*', 'h6p', 'h6p']], ['*', 'h1p', ['*', 'h5p', 'h5p'], ['*', 'h6p', 'h6p', 'h6p']], ['*', ['*', 'h2p', 'h2p', 'h2p'], ['*', 'h3p', 'h3p'], 'h4p'], ['*', ['*', 'h2p', 'h2p', 'h2p'], ['*', 'h3p', 'h3p'], 'h5p'], ['*', ['*', 'h2p', 'h2p', 'h2p'], ['*', 'h3p', 'h3p'], 'h6p'], ['*', ['*', 'h2p', 'h2p', 'h2p'], 'h3p', ['*', 'h4p', 'h4p']], ['*', 2.0, ['*', 'h2p', 'h2p', 'h2p'], 'h3p', 'h4p', 'h5p'], ['*', 2.0, ['*', 'h2p', 'h2p', 'h2p'], 'h3p', 'h4p', 'h6p'], ['*', ['*', 'h2p', 'h2p', 'h2p'], 'h3p', ['*', 'h5p', 'h5p']], ['*', 2.0, ['*', 'h2p', 'h2p', 'h2p'], 'h3p', 'h5p', 'h6p'], ['*', ['*', 'h2p', 'h2p', 'h2p'], 'h3p', ['*', 'h6p', 'h6p']], ['*', ['*', 'h2p', 'h2p', 'h2p'], ['*', 'h4p', 'h4p'], 'h5p'], ['*', ['*', 'h2p', 'h2p', 'h2p'], 'h4p', ['*', 'h5p', 'h5p']], ['*', 2.0, ['*', 'h2p', 'h2p', 'h2p'], 'h4p', 'h5p', 'h6p'], ['*', ['*', 'h2p', 'h2p', 'h2p'], ['*', 'h5p', 'h5p'], 'h6p'], ['*', ['*', 'h2p', 'h2p', 'h2p'], 'h5p', ['*', 'h6p', 'h6p']], ['*', 2.0, ['*', 'h2p', 'h2p'], ['*', 'h3p', 'h3p', 'h3p'], 'h4p'], ['*', 2.0, ['*', 'h2p', 'h2p'], ['*', 'h3p', 'h3p', 'h3p'], 'h5p'], ['*', 2.0, ['*', 'h2p', 'h2p'], ['*', 'h3p', 'h3p', 'h3p'], 'h6p'], ['*', 4.0, ['*', 'h2p', 'h2p'], ['*', 'h3p', 'h3p'], ['*', 'h4p', 'h4p']], ['*', 8.0, ['*', 'h2p', 'h2p'], ['*', 'h3p', 'h3p'], 'h4p', 'h5p'], ['*', 8.0, ['*', 'h2p', 'h2p'], ['*', 'h3p', 'h3p'], 'h4p', 'h6p'], ['*', 4.0, ['*', 'h2p', 'h2p'], ['*', 'h3p', 'h3p'], ['*', 'h5p', 'h5p']], ['*', 8.0, ['*', 'h2p', 'h2p'], ['*', 'h3p', 'h3p'], 'h5p', 'h6p'], ['*', 4.0, ['*', 'h2p', 'h2p'], ['*', 'h3p', 'h3p'], ['*', 'h6p', 'h6p']], ['*', ['*', 'h2p', 'h2p'], 'h3p', ['*', 'h4p', 'h4p', 'h4p']], ['*', 7.0, ['*', 'h2p', 'h2p'], 'h3p', ['*', 'h4p', 'h4p'], 'h5p'], ['*', 3.0, ['*', 'h2p', 'h2p'], 'h3p', ['*', 'h4p', 'h4p'], 'h6p'], ['*', 7.0, ['*', 'h2p', 'h2p'], 'h3p', 'h4p', ['*', 'h5p', 'h5p']], ['*', 14.0, ['*', 'h2p', 'h2p'], 'h3p', 'h4p', 'h5p', 'h6p'], ['*', 3.0, ['*', 'h2p', 'h2p'], 'h3p', 'h4p', ['*', 'h6p', 'h6p']], ['*', ['*', 'h2p', 'h2p'], 'h3p', ['*', 'h5p', 'h5p', 'h5p']], ['*', 7.0, ['*', 'h2p', 'h2p'], 'h3p', ['*', 'h5p', 'h5p'], 'h6p'], ['*', 7.0, ['*', 'h2p', 'h2p'], 'h3p', 'h5p', ['*', 'h6p', 'h6p']], ['*', ['*', 'h2p', 'h2p'], 'h3p', ['*', 'h6p', 'h6p', 'h6p']], ['*', ['*', 'h2p', 'h2p'], ['*', 'h4p', 'h4p', 'h4p'], 'h5p'], ['*', 2.0, ['*', 'h2p', 'h2p'], ['*', 'h4p', 'h4p'], ['*', 'h5p', 'h5p']], ['*', 3.0, ['*', 'h2p', 'h2p'], ['*', 'h4p', 'h4p'], 'h5p', 'h6p'], ['*', ['*', 'h2p', 'h2p'], 'h4p', ['*', 'h5p', 'h5p', 'h5p']], ['*', 4.0, ['*', 'h2p', 'h2p'], 'h4p', ['*', 'h5p', 'h5p'], 'h6p'], ['*', 3.0, ['*', 'h2p', 'h2p'], 'h4p', 'h5p', ['*', 'h6p', 'h6p']], ['*', ['*', 'h2p', 'h2p'], ['*', 'h5p', 'h5p', 'h5p'], 'h6p'], ['*', 2.0, ['*', 'h2p', 'h2p'], ['*', 'h5p', 'h5p'], ['*', 'h6p', 'h6p']], ['*', ['*', 'h2p', 'h2p'], 'h5p', ['*', 'h6p', 'h6p', 'h6p']], ['*', 4.0, 'h2p', ['*', 'h3p', 'h3p', 'h3p'], ['*', 'h4p', 'h4p']], ['*', 8.0, 'h2p', ['*', 'h3p', 'h3p', 'h3p'], 'h4p', 'h5p'], ['*', 8.0, 'h2p', ['*', 'h3p', 'h3p', 'h3p'], 'h4p', 'h6p'], ['*', 4.0, 'h2p', ['*', 'h3p', 'h3p', 'h3p'], ['*', 'h5p', 'h5p']], ['*', 8.0, 'h2p', ['*', 'h3p', 'h3p', 'h3p'], 'h5p', 'h6p'], ['*', 4.0, 'h2p', ['*', 'h3p', 'h3p', 'h3p'], ['*', 'h6p', 'h6p']], ['*', 2.0, 'h2p', ['*', 'h3p', 'h3p'], ['*', 'h4p', 'h4p', 'h4p']], ['*', 14.0, 'h2p', ['*', 'h3p', 'h3p'], ['*', 'h4p', 'h4p'], 'h5p'], ['*', 6.0, 'h2p', ['*', 'h3p', 'h3p'], ['*', 'h4p', 'h4p'], 'h6p'], ['*', 14.0, 'h2p', ['*', 'h3p', 'h3p'], 'h4p', ['*', 'h5p', 'h5p']], ['*', 28.0, 'h2p', ['*', 'h3p', 'h3p'], 'h4p', 'h5p', 'h6p'], ['*', 6.0, 'h2p', ['*', 'h3p', 'h3p'], 'h4p', ['*', 'h6p', 'h6p']], ['*', 2.0, 'h2p', ['*', 'h3p', 'h3p'], ['*', 'h5p', 'h5p', 'h5p']], ['*', 14.0, 'h2p', ['*', 'h3p', 'h3p'], ['*', 'h5p', 'h5p'], 'h6p'], ['*', 14.0, 'h2p', ['*', 'h3p', 'h3p'], 'h5p', ['*', 'h6p', 'h6p']], ['*', 2.0, 'h2p', ['*', 'h3p', 'h3p'], ['*', 'h6p', 'h6p', 'h6p']], ['*', 4.0, 'h2p', 'h3p', ['*', 'h4p', 'h4p', 'h4p'], 'h5p'], ['*', 8.0, 'h2p', 'h3p', ['*', 'h4p', 'h4p'], ['*', 'h5p', 'h5p']], ['*', 12.0, 'h2p', 'h3p', ['*', 'h4p', 'h4p'], 'h5p', 'h6p'], ['*', 4.0, 'h2p', 'h3p', 'h4p', ['*', 'h5p', 'h5p', 'h5p']], ['*', 16.0, 'h2p', 'h3p', 'h4p', ['*', 'h5p', 'h5p'], 'h6p'], ['*', 12.0, 'h2p', 'h3p', 'h4p', 'h5p', ['*', 'h6p', 'h6p']], ['*', 4.0, 'h2p', 'h3p', ['*', 'h5p', 'h5p', 'h5p'], 'h6p'], ['*', 8.0, 'h2p', 'h3p', ['*', 'h5p', 'h5p'], ['*', 'h6p', 'h6p']], ['*', 4.0, 'h2p', 'h3p', 'h5p', ['*', 'h6p', 'h6p', 'h6p']], ['*', 'h2p', ['*', 'h4p', 'h4p', 'h4p'], ['*', 'h5p', 'h5p']], ['*', 'h2p', ['*', 'h4p', 'h4p'], ['*', 'h5p', 'h5p', 'h5p']], ['*', 3.0, 'h2p', ['*', 'h4p', 'h4p'], ['*', 'h5p', 'h5p'], 'h6p'], ['*', 2.0, 'h2p', 'h4p', ['*', 'h5p', 'h5p', 'h5p'], 'h6p'], ['*', 3.0, 'h2p', 'h4p', ['*', 'h5p', 'h5p'], ['*', 'h6p', 'h6p']], ['*', 'h2p', ['*', 'h5p', 'h5p', 'h5p'], ['*', 'h6p', 'h6p']], ['*', 'h2p', ['*', 'h5p', 'h5p'], ['*', 'h6p', 'h6p', 'h6p']], ['*', 8.0, ['*', 'h3p', 'h3p', 'h3p'], ['*', 'h4p', 'h4p'], 'h5p'], ['*', 8.0, ['*', 'h3p', 'h3p', 'h3p'], 'h4p', ['*', 'h5p', 'h5p']], ['*', 16.0, ['*', 'h3p', 'h3p', 'h3p'], 'h4p', 'h5p', 'h6p'], ['*', 8.0, ['*', 'h3p', 'h3p', 'h3p'], ['*', 'h5p', 'h5p'], 'h6p'], ['*', 8.0, ['*', 'h3p', 'h3p', 'h3p'], 'h5p', ['*', 'h6p', 'h6p']], ['*', 4.0, ['*', 'h3p', 'h3p'], ['*', 'h4p', 'h4p', 'h4p'], 'h5p'], ['*', 8.0, ['*', 'h3p', 'h3p'], ['*', 'h4p', 'h4p'], ['*', 'h5p', 'h5p']], ['*', 12.0, ['*', 'h3p', 'h3p'], ['*', 'h4p', 'h4p'], 'h5p', 'h6p'], ['*', 4.0, ['*', 'h3p', 'h3p'], 'h4p', ['*', 'h5p', 'h5p', 'h5p']], ['*', 16.0, ['*', 'h3p', 'h3p'], 'h4p', ['*', 'h5p', 'h5p'], 'h6p'], ['*', 12.0, ['*', 'h3p', 'h3p'], 'h4p', 'h5p', ['*', 'h6p', 'h6p']], ['*', 4.0, ['*', 'h3p', 'h3p'], ['*', 'h5p', 'h5p', 'h5p'], 'h6p'], ['*', 8.0, ['*', 'h3p', 'h3p'], ['*', 'h5p', 'h5p'], ['*', 'h6p', 'h6p']], ['*', 4.0, ['*', 'h3p', 'h3p'], 'h5p', ['*', 'h6p', 'h6p', 'h6p']], ['*', 2.0, 'h3p', ['*', 'h4p', 'h4p', 'h4p'], ['*', 'h5p', 'h5p']], ['*', 2.0, 'h3p', ['*', 'h4p', 'h4p'], ['*', 'h5p', 'h5p', 'h5p']], ['*', 6.0, 'h3p', ['*', 'h4p', 'h4p'], ['*', 'h5p', 'h5p'], 'h6p'], ['*', 4.0, 'h3p', 'h4p', ['*', 'h5p', 'h5p', 'h5p'], 'h6p'], ['*', 6.0, 'h3p', 'h4p', ['*', 'h5p', 'h5p'], ['*', 'h6p', 'h6p']], ['*', 2.0, 'h3p', ['*', 'h5p', 'h5p', 'h5p'], ['*', 'h6p', 'h6p']], ['*', 2.0, 'h3p', ['*', 'h5p', 'h5p'], ['*', 'h6p', 'h6p', 'h6p']]]],
['+', ['*', -1.0, ['*', 2.0, ['*', 'h1n', 'h1n', 'h1n'], ['*', 'h2n', 'h2n'], 'h3n']], ['*', 2.0, ['*', 'h1n', 'h1n', 'h1n'], ['*', 'h2n', 'h2n'], 'h5n'], ['*', -1.0, ['*', 8.0, ['*', 'h1n', 'h1n', 'h1n'], 'h2n', ['*', 'h3n', 'h3n']]], ['*', 4.0, ['*', 'h1n', 'h1n', 'h1n'], 'h2n', 'h3n', 'h4n'], ['*', 10.0, ['*', 'h1n', 'h1n', 'h1n'], 'h2n', 'h3n', 'h5n'], ['*', 10.0, ['*', 'h1n', 'h1n', 'h1n'], 'h2n', 'h3n', 'h6n'], ['*', 2.0, ['*', 'h1n', 'h1n', 'h1n'], 'h2n', 'h4n', 'h5n'], ['*', 2.0, ['*', 'h1n', 'h1n', 'h1n'], 'h2n', ['*', 'h5n', 'h5n']], ['*', 24.0, ['*', 'h1n', 'h1n', 'h1n'], ['*', 'h3n', 'h3n'], 'h4n'], ['*', 16.0, ['*', 'h1n', 'h1n', 'h1n'], ['*', 'h3n', 'h3n'], 'h5n'], ['*', 16.0, ['*', 'h1n', 'h1n', 'h1n'], ['*', 'h3n', 'h3n'], 'h6n'], ['*', 6.0, ['*', 'h1n', 'h1n', 'h1n'], 'h3n', ['*', 'h4n', 'h4n']], ['*', 6.0, ['*', 'h1n', 'h1n', 'h1n'], 'h3n', 'h4n', 'h5n'], ['*', 10.0, ['*', 'h1n', 'h1n', 'h1n'], 'h3n', 'h4n', 'h6n'], ['*', 4.0, ['*', 'h1n', 'h1n', 'h1n'], 'h3n', ['*', 'h5n', 'h5n']], ['*', 8.0, ['*', 'h1n', 'h1n', 'h1n'], 'h3n', 'h5n', 'h6n'], ['*', 4.0, ['*', 'h1n', 'h1n', 'h1n'], 'h3n', ['*', 'h6n', 'h6n']], ['*', 2.0, ['*', 'h1n', 'h1n', 'h1n'], 'h4n', 'h5n', 'h6n'], ['*', 2.0, ['*', 'h1n', 'h1n', 'h1n'], ['*', 'h5n', 'h5n'], 'h6n'], ['*', 2.0, ['*', 'h1n', 'h1n', 'h1n'], 'h5n', ['*', 'h6n', 'h6n']], ['*', -1.0, ['*', ['*', 'h1n', 'h1n'], ['*', 'h2n', 'h2n', 'h2n'], 'h3n']], ['*', ['*', 'h1n', 'h1n'], ['*', 'h2n', 'h2n', 'h2n'], 'h5n'], ['*', -1.0, ['*', 9.0, ['*', 'h1n', 'h1n'], ['*', 'h2n', 'h2n'], ['*', 'h3n', 'h3n']]], ['*', ['*', 'h1n', 'h1n'], ['*', 'h2n', 'h2n'], 'h3n', 'h4n'], ['*', 10.0, ['*', 'h1n', 'h1n'], ['*', 'h2n', 'h2n'], 'h3n', 'h5n'], ['*', 8.0, ['*', 'h1n', 'h1n'], ['*', 'h2n', 'h2n'], 'h3n', 'h6n'], ['*', 6.0, ['*', 'h1n', 'h1n'], ['*', 'h2n', 'h2n'], 'h4n', 'h5n'], ['*', 3.0, ['*', 'h1n', 'h1n'], ['*', 'h2n', 'h2n'], ['*', 'h5n', 'h5n']], ['*', 3.0, ['*', 'h1n', 'h1n'], ['*', 'h2n', 'h2n'], 'h5n', 'h6n'], ['*', -1.0, ['*', 8.0, ['*', 'h1n', 'h1n'], 'h2n', ['*', 'h3n', 'h3n', 'h3n']]], ['*', 28.0, ['*', 'h1n', 'h1n'], 'h2n', ['*', 'h3n', 'h3n'], 'h4n'], ['*', 26.0, ['*', 'h1n', 'h1n'], 'h2n', ['*', 'h3n', 'h3n'], 'h5n'], ['*', 34.0, ['*', 'h1n', 'h1n'], 'h2n', ['*', 'h3n', 'h3n'], 'h6n'], ['*', 13.0, ['*', 'h1n', 'h1n'], 'h2n', 'h3n', ['*', 'h4n', 'h4n']], ['*', 28.0, ['*', 'h1n', 'h1n'], 'h2n', 'h3n', 'h4n', 'h5n'], ['*', 30.0, ['*', 'h1n', 'h1n'], 'h2n', 'h3n', 'h4n', 'h6n'], ['*', 15.0, ['*', 'h1n', 'h1n'], 'h2n', 'h3n', ['*', 'h5n', 'h5n']], ['*', 32.0, ['*', 'h1n', 'h1n'], 'h2n', 'h3n', 'h5n', 'h6n'], ['*', 17.0, ['*', 'h1n', 'h1n'], 'h2n', 'h3n', ['*', 'h6n', 'h6n']], ['*', 3.0, ['*', 'h1n', 'h1n'], 'h2n', ['*', 'h4n', 'h4n'], 'h5n'], ['*', 6.0, ['*', 'h1n', 'h1n'], 'h2n', 'h4n', ['*', 'h5n', 'h5n']], ['*', 6.0, ['*', 'h1n', 'h1n'], 'h2n', 'h4n', 'h5n', 'h6n'], ['*', ['*', 'h1n', 'h1n'], 'h2n', ['*', 'h5n', 'h5n', 'h5n']], ['*', 4.0, ['*', 'h1n', 'h1n'], 'h2n', ['*', 'h5n', 'h5n'], 'h6n'], ['*', 3.0, ['*', 'h1n', 'h1n'], 'h2n', 'h5n', ['*', 'h6n', 'h6n']], ['*', 24.0, ['*', 'h1n', 'h1n'], ['*', 'h3n', 'h3n', 'h3n'], 'h4n'], ['*', 16.0, ['*', 'h1n', 'h1n'], ['*', 'h3n', 'h3n', 'h3n'], 'h5n'], ['*', 16.0, ['*', 'h1n', 'h1n'], ['*', 'h3n', 'h3n', 'h3n'], 'h6n'], ['*', 21.0, ['*', 'h1n', 'h1n'], ['*', 'h3n', 'h3n'], ['*', 'h4n', 'h4n']], ['*', 46.0, ['*', 'h1n', 'h1n'], ['*', 'h3n', 'h3n'], 'h4n', 'h5n'], ['*', 38.0, ['*', 'h1n', 'h1n'], ['*', 'h3n', 'h3n'], 'h4n', 'h6n'], ['*', 16.0, ['*', 'h1n', 'h1n'], ['*', 'h3n', 'h3n'], ['*', 'h5n', 'h5n']], ['*', 40.0, ['*', 'h1n', 'h1n'], ['*', 'h3n', 'h3n'], 'h5n', 'h6n'], ['*', 16.0, ['*', 'h1n', 'h1n'], ['*', 'h3n', 'h3n'], ['*', 'h6n', 'h6n']], ['*', 3.0, ['*', 'h1n', 'h1n'], 'h3n', ['*', 'h4n', 'h4n', 'h4n']], ['*', 14.0, ['*', 'h1n', 'h1n'], 'h3n', ['*', 'h4n', 'h4n'], 'h5n'], ['*', 8.0, ['*', 'h1n', 'h1n'], 'h3n', ['*', 'h4n', 'h4n'], 'h6n'], ['*', 11.0, ['*', 'h1n', 'h1n'], 'h3n', 'h4n', ['*', 'h5n', 'h5n']], ['*', 24.0, ['*', 'h1n', 'h1n'], 'h3n', 'h4n', 'h5n', 'h6n'], ['*', 7.0, ['*', 'h1n', 'h1n'], 'h3n', 'h4n', ['*', 'h6n', 'h6n']], ['*', 2.0, ['*', 'h1n', 'h1n'], 'h3n', ['*', 'h5n', 'h5n', 'h5n']], ['*', 12.0, ['*', 'h1n', 'h1n'], 'h3n', ['*', 'h5n', 'h5n'], 'h6n'], ['*', 12.0, ['*', 'h1n', 'h1n'], 'h3n', 'h5n', ['*', 'h6n', 'h6n']], ['*', 2.0, ['*', 'h1n', 'h1n'], 'h3n', ['*', 'h6n', 'h6n', 'h6n']], ['*', ['*', 'h1n', 'h1n'], ['*', 'h4n', 'h4n'], 'h5n', 'h6n'], ['*', 4.0, ['*', 'h1n', 'h1n'], 'h4n', ['*', 'h5n', 'h5n'], 'h6n'], ['*', 2.0, ['*', 'h1n', 'h1n'], 'h4n', 'h5n', ['*', 'h6n', 'h6n']], ['*', ['*', 'h1n', 'h1n'], ['*', 'h5n', 'h5n', 'h5n'], 'h6n'], ['*', 3.0, ['*', 'h1n', 'h1n'], ['*', 'h5n', 'h5n'], ['*', 'h6n', 'h6n']], ['*', ['*', 'h1n', 'h1n'], 'h5n', ['*', 'h6n', 'h6n', 'h6n']], ['*', -1.0, ['*', 'h1n', ['*', 'h2n', 'h2n', 'h2n'], ['*', 'h3n', 'h3n']]], ['*', 'h1n', ['*', 'h2n', 'h2n', 'h2n'], 'h3n', 'h5n'], ['*', 2.0, 'h1n', ['*', 'h2n', 'h2n', 'h2n'], 'h3n', 'h6n'], ['*', 2.0, 'h1n', ['*', 'h2n', 'h2n', 'h2n'], 'h4n', 'h5n'], ['*', 'h1n', ['*', 'h2n', 'h2n', 'h2n'], ['*', 'h5n', 'h5n']], ['*', 'h1n', ['*', 'h2n', 'h2n', 'h2n'], 'h5n', 'h6n'], ['*', -1.0, ['*', 2.0, 'h1n', ['*', 'h2n', 'h2n'], ['*', 'h3n', 'h3n', 'h3n']]], ['*', 8.0, 'h1n', ['*', 'h2n', 'h2n'], ['*', 'h3n', 'h3n'], 'h4n'], ['*', 9.0, 'h1n', ['*', 'h2n', 'h2n'], ['*', 'h3n', 'h3n'], 'h5n'], ['*', 15.0, 'h1n', ['*', 'h2n', 'h2n'], ['*', 'h3n', 'h3n'], 'h6n'], ['*', 6.0, 'h1n', ['*', 'h2n', 'h2n'], 'h3n', ['*', 'h4n', 'h4n']], ['*', 20.0, 'h1n', ['*', 'h2n', 'h2n'], 'h3n', 'h4n', 'h5n'], ['*', 15.0, 'h1n', ['*', 'h2n', 'h2n'], 'h3n', 'h4n', 'h6n'], ['*', 8.0, 'h1n', ['*', 'h2n', 'h2n'], 'h3n', ['*', 'h5n', 'h5n']], ['*', 19.0, 'h1n', ['*', 'h2n', 'h2n'], 'h3n', 'h5n', 'h6n'], ['*', 9.0, 'h1n', ['*', 'h2n', 'h2n'], 'h3n', ['*', 'h6n', 'h6n']], ['*', 5.0, 'h1n', ['*', 'h2n', 'h2n'], ['*', 'h4n', 'h4n'], 'h5n'], ['*', 5.0, 'h1n', ['*', 'h2n', 'h2n'], 'h4n', ['*', 'h5n', 'h5n']], ['*', 9.0, 'h1n', ['*', 'h2n', 'h2n'], 'h4n', 'h5n', 'h6n'], ['*', 'h1n', ['*', 'h2n', 'h2n'], ['*', 'h5n', 'h5n', 'h5n']], ['*', 4.0, 'h1n', ['*', 'h2n', 'h2n'], ['*', 'h5n', 'h5n'], 'h6n'], ['*', 4.0, 'h1n', ['*', 'h2n', 'h2n'], 'h5n', ['*', 'h6n', 'h6n']], ['*', 10.0, 'h1n', 'h2n', ['*', 'h3n', 'h3n', 'h3n'], 'h4n'], ['*', 8.0, 'h1n', 'h2n', ['*', 'h3n', 'h3n', 'h3n'], 'h5n'], ['*', 16.0, 'h1n', 'h2n', ['*', 'h3n', 'h3n', 'h3n'], 'h6n'], ['*', 17.0, 'h1n', 'h2n', ['*', 'h3n', 'h3n'], ['*', 'h4n', 'h4n']], ['*', 49.0, 'h1n', 'h2n', ['*', 'h3n', 'h3n'], 'h4n', 'h5n'], ['*', 39.0, 'h1n', 'h2n', ['*', 'h3n', 'h3n'], 'h4n', 'h6n'], ['*', 18.0, 'h1n', 'h2n', ['*', 'h3n', 'h3n'], ['*', 'h5n', 'h5n']], ['*', 48.0, 'h1n', 'h2n', ['*', 'h3n', 'h3n'], 'h5n', 'h6n'], ['*', 22.0, 'h1n', 'h2n', ['*', 'h3n', 'h3n'], ['*', 'h6n', 'h6n']], ['*', 4.0, 'h1n', 'h2n', 'h3n', ['*', 'h4n', 'h4n', 'h4n']], ['*', 23.0, 'h1n', 'h2n', 'h3n', ['*', 'h4n', 'h4n'], 'h5n'], ['*', 13.0, 'h1n', 'h2n', 'h3n', ['*', 'h4n', 'h4n'], 'h6n'], ['*', 23.0, 'h1n', 'h2n', 'h3n', 'h4n', ['*', 'h5n', 'h5n']], ['*', 46.0, 'h1n', 'h2n', 'h3n', 'h4n', 'h5n', 'h6n'], ['*', 14.0, 'h1n', 'h2n', 'h3n', 'h4n', ['*', 'h6n', 'h6n']], ['*', 5.0, 'h1n', 'h2n', 'h3n', ['*', 'h5n', 'h5n', 'h5n']], ['*', 21.0, 'h1n', 'h2n', 'h3n', ['*', 'h5n', 'h5n'], 'h6n'], ['*', 23.0, 'h1n', 'h2n', 'h3n', 'h5n', ['*', 'h6n', 'h6n']], ['*', 5.0, 'h1n', 'h2n', 'h3n', ['*', 'h6n', 'h6n', 'h6n']], ['*', 'h1n', 'h2n', ['*', 'h4n', 'h4n', 'h4n'], 'h5n'], ['*', 5.0, 'h1n', 'h2n', ['*', 'h4n', 'h4n'], ['*', 'h5n', 'h5n']], ['*', 3.0, 'h1n', 'h2n', ['*', 'h4n', 'h4n'], 'h5n', 'h6n'], ['*', 2.0, 'h1n', 'h2n', 'h4n', ['*', 'h5n', 'h5n', 'h5n']], ['*', 9.0, 'h1n', 'h2n', 'h4n', ['*', 'h5n', 'h5n'], 'h6n'], ['*', 3.0, 'h1n', 'h2n', 'h4n', 'h5n', ['*', 'h6n', 'h6n']], ['*', 2.0, 'h1n', 'h2n', ['*', 'h5n', 'h5n', 'h5n'], 'h6n'], ['*', 4.0, 'h1n', 'h2n', ['*', 'h5n', 'h5n'], ['*', 'h6n', 'h6n']], ['*', 'h1n', 'h2n', 'h5n', ['*', 'h6n', 'h6n', 'h6n']], ['*', 12.0, 'h1n', ['*', 'h3n', 'h3n', 'h3n'], ['*', 'h4n', 'h4n']], ['*', 32.0, 'h1n', ['*', 'h3n', 'h3n', 'h3n'], 'h4n', 'h5n'], ['*', 20.0, 'h1n', ['*', 'h3n', 'h3n', 'h3n'], 'h4n', 'h6n'], ['*', 8.0, 'h1n', ['*', 'h3n', 'h3n', 'h3n'], ['*', 'h5n', 'h5n']], ['*', 24.0, 'h1n', ['*', 'h3n', 'h3n', 'h3n'], 'h5n', 'h6n'], ['*', 8.0, 'h1n', ['*', 'h3n', 'h3n', 'h3n'], ['*', 'h6n', 'h6n']], ['*', 6.0, 'h1n', ['*', 'h3n', 'h3n'], ['*', 'h4n', 'h4n', 'h4n']], ['*', 26.0, 'h1n', ['*', 'h3n', 'h3n'], ['*', 'h4n', 'h4n'], 'h5n'], ['*', 16.0, 'h1n', ['*', 'h3n', 'h3n'], ['*', 'h4n', 'h4n'], 'h6n'], ['*', 26.0, 'h1n', ['*', 'h3n', 'h3n'], 'h4n', ['*', 'h5n', 'h5n']], ['*', 50.0, 'h1n', ['*', 'h3n', 'h3n'], 'h4n', 'h5n', 'h6n'], ['*', 14.0, 'h1n', ['*', 'h3n', 'h3n'], 'h4n', ['*', 'h6n', 'h6n']], ['*', 4.0, 'h1n', ['*', 'h3n', 'h3n'], ['*', 'h5n', 'h5n', 'h5n']], ['*', 24.0, 'h1n', ['*', 'h3n', 'h3n'], ['*', 'h5n', 'h5n'], 'h6n'], ['*', 24.0, 'h1n', ['*', 'h3n', 'h3n'], 'h5n', ['*', 'h6n', 'h6n']], ['*', 4.0, 'h1n', ['*', 'h3n', 'h3n'], ['*', 'h6n', 'h6n', 'h6n']], ['*', 4.0, 'h1n', 'h3n', ['*', 'h4n', 'h4n', 'h4n'], 'h5n'], ['*', 11.0, 'h1n', 'h3n', ['*', 'h4n', 'h4n'], ['*', 'h5n', 'h5n']], ['*', 12.0, 'h1n', 'h3n', ['*', 'h4n', 'h4n'], 'h5n', 'h6n'], ['*', 3.0, 'h1n', 'h3n', 'h4n', ['*', 'h5n', 'h5n', 'h5n']], ['*', 21.0, 'h1n', 'h3n', 'h4n', ['*', 'h5n', 'h5n'], 'h6n'], ['*', 12.0, 'h1n', 'h3n', 'h4n', 'h5n', ['*', 'h6n', 'h6n']], ['*', 4.0, 'h1n', 'h3n', ['*', 'h5n', 'h5n', 'h5n'], 'h6n'], ['*', 10.0, 'h1n', 'h3n', ['*', 'h5n', 'h5n'], ['*', 'h6n', 'h6n']], ['*', 4.0, 'h1n', 'h3n', 'h5n', ['*', 'h6n', 'h6n', 'h6n']], ['*', 'h1n', ['*', 'h4n', 'h4n'], ['*', 'h5n', 'h5n'], 'h6n'], ['*', 'h1n', 'h4n', ['*', 'h5n', 'h5n', 'h5n'], 'h6n'], ['*', 2.0, 'h1n', 'h4n', ['*', 'h5n', 'h5n'], ['*', 'h6n', 'h6n']], ['*', 'h1n', ['*', 'h5n', 'h5n', 'h5n'], ['*', 'h6n', 'h6n']], ['*', 'h1n', ['*', 'h5n', 'h5n'], ['*', 'h6n', 'h6n', 'h6n']], ['*', ['*', 'h2n', 'h2n', 'h2n'], ['*', 'h3n', 'h3n'], 'h4n'], ['*', ['*', 'h2n', 'h2n', 'h2n'], ['*', 'h3n', 'h3n'], 'h5n'], ['*', ['*', 'h2n', 'h2n', 'h2n'], ['*', 'h3n', 'h3n'], 'h6n'], ['*', ['*', 'h2n', 'h2n', 'h2n'], 'h3n', ['*', 'h4n', 'h4n']], ['*', 2.0, ['*', 'h2n', 'h2n', 'h2n'], 'h3n', 'h4n', 'h5n'], ['*', 2.0, ['*', 'h2n', 'h2n', 'h2n'], 'h3n', 'h4n', 'h6n'], ['*', ['*', 'h2n', 'h2n', 'h2n'], 'h3n', ['*', 'h5n', 'h5n']], ['*', 2.0, ['*', 'h2n', 'h2n', 'h2n'], 'h3n', 'h5n', 'h6n'], ['*', ['*', 'h2n', 'h2n', 'h2n'], 'h3n', ['*', 'h6n', 'h6n']], ['*', ['*', 'h2n', 'h2n', 'h2n'], ['*', 'h4n', 'h4n'], 'h5n'], ['*', ['*', 'h2n', 'h2n', 'h2n'], 'h4n', ['*', 'h5n', 'h5n']], ['*', 2.0, ['*', 'h2n', 'h2n', 'h2n'], 'h4n', 'h5n', 'h6n'], ['*', ['*', 'h2n', 'h2n', 'h2n'], ['*', 'h5n', 'h5n'], 'h6n'], ['*', ['*', 'h2n', 'h2n', 'h2n'], 'h5n', ['*', 'h6n', 'h6n']], ['*', 2.0, ['*', 'h2n', 'h2n'], ['*', 'h3n', 'h3n', 'h3n'], 'h4n'], ['*', 2.0, ['*', 'h2n', 'h2n'], ['*', 'h3n', 'h3n', 'h3n'], 'h5n'], ['*', 2.0, ['*', 'h2n', 'h2n'], ['*', 'h3n', 'h3n', 'h3n'], 'h6n'], ['*', 4.0, ['*', 'h2n', 'h2n'], ['*', 'h3n', 'h3n'], ['*', 'h4n', 'h4n']], ['*', 8.0, ['*', 'h2n', 'h2n'], ['*', 'h3n', 'h3n'], 'h4n', 'h5n'], ['*', 8.0, ['*', 'h2n', 'h2n'], ['*', 'h3n', 'h3n'], 'h4n', 'h6n'], ['*', 4.0, ['*', 'h2n', 'h2n'], ['*', 'h3n', 'h3n'], ['*', 'h5n', 'h5n']], ['*', 8.0, ['*', 'h2n', 'h2n'], ['*', 'h3n', 'h3n'], 'h5n', 'h6n'], ['*', 4.0, ['*', 'h2n', 'h2n'], ['*', 'h3n', 'h3n'], ['*', 'h6n', 'h6n']], ['*', ['*', 'h2n', 'h2n'], 'h3n', ['*', 'h4n', 'h4n', 'h4n']], ['*', 7.0, ['*', 'h2n', 'h2n'], 'h3n', ['*', 'h4n', 'h4n'], 'h5n'], ['*', 3.0, ['*', 'h2n', 'h2n'], 'h3n', ['*', 'h4n', 'h4n'], 'h6n'], ['*', 7.0, ['*', 'h2n', 'h2n'], 'h3n', 'h4n', ['*', 'h5n', 'h5n']], ['*', 14.0, ['*', 'h2n', 'h2n'], 'h3n', 'h4n', 'h5n', 'h6n'], ['*', 3.0, ['*', 'h2n', 'h2n'], 'h3n', 'h4n', ['*', 'h6n', 'h6n']], ['*', ['*', 'h2n', 'h2n'], 'h3n', ['*', 'h5n', 'h5n', 'h5n']], ['*', 7.0, ['*', 'h2n', 'h2n'], 'h3n', ['*', 'h5n', 'h5n'], 'h6n'], ['*', 7.0, ['*', 'h2n', 'h2n'], 'h3n', 'h5n', ['*', 'h6n', 'h6n']], ['*', ['*', 'h2n', 'h2n'], 'h3n', ['*', 'h6n', 'h6n', 'h6n']], ['*', ['*', 'h2n', 'h2n'], ['*', 'h4n', 'h4n', 'h4n'], 'h5n'], ['*', 2.0, ['*', 'h2n', 'h2n'], ['*', 'h4n', 'h4n'], ['*', 'h5n', 'h5n']], ['*', 3.0, ['*', 'h2n', 'h2n'], ['*', 'h4n', 'h4n'], 'h5n', 'h6n'], ['*', ['*', 'h2n', 'h2n'], 'h4n', ['*', 'h5n', 'h5n', 'h5n']], ['*', 4.0, ['*', 'h2n', 'h2n'], 'h4n', ['*', 'h5n', 'h5n'], 'h6n'], ['*', 3.0, ['*', 'h2n', 'h2n'], 'h4n', 'h5n', ['*', 'h6n', 'h6n']], ['*', ['*', 'h2n', 'h2n'], ['*', 'h5n', 'h5n', 'h5n'], 'h6n'], ['*', 2.0, ['*', 'h2n', 'h2n'], ['*', 'h5n', 'h5n'], ['*', 'h6n', 'h6n']], ['*', ['*', 'h2n', 'h2n'], 'h5n', ['*', 'h6n', 'h6n', 'h6n']], ['*', 4.0, 'h2n', ['*', 'h3n', 'h3n', 'h3n'], ['*', 'h4n', 'h4n']], ['*', 8.0, 'h2n', ['*', 'h3n', 'h3n', 'h3n'], 'h4n', 'h5n'], ['*', 8.0, 'h2n', ['*', 'h3n', 'h3n', 'h3n'], 'h4n', 'h6n'], ['*', 4.0, 'h2n', ['*', 'h3n', 'h3n', 'h3n'], ['*', 'h5n', 'h5n']], ['*', 8.0, 'h2n', ['*', 'h3n', 'h3n', 'h3n'], 'h5n', 'h6n'], ['*', 4.0, 'h2n', ['*', 'h3n', 'h3n', 'h3n'], ['*', 'h6n', 'h6n']], ['*', 2.0, 'h2n', ['*', 'h3n', 'h3n'], ['*', 'h4n', 'h4n', 'h4n']], ['*', 14.0, 'h2n', ['*', 'h3n', 'h3n'], ['*', 'h4n', 'h4n'], 'h5n'], ['*', 6.0, 'h2n', ['*', 'h3n', 'h3n'], ['*', 'h4n', 'h4n'], 'h6n'], ['*', 14.0, 'h2n', ['*', 'h3n', 'h3n'], 'h4n', ['*', 'h5n', 'h5n']], ['*', 28.0, 'h2n', ['*', 'h3n', 'h3n'], 'h4n', 'h5n', 'h6n'], ['*', 6.0, 'h2n', ['*', 'h3n', 'h3n'], 'h4n', ['*', 'h6n', 'h6n']], ['*', 2.0, 'h2n', ['*', 'h3n', 'h3n'], ['*', 'h5n', 'h5n', 'h5n']], ['*', 14.0, 'h2n', ['*', 'h3n', 'h3n'], ['*', 'h5n', 'h5n'], 'h6n'], ['*', 14.0, 'h2n', ['*', 'h3n', 'h3n'], 'h5n', ['*', 'h6n', 'h6n']], ['*', 2.0, 'h2n', ['*', 'h3n', 'h3n'], ['*', 'h6n', 'h6n', 'h6n']], ['*', 4.0, 'h2n', 'h3n', ['*', 'h4n', 'h4n', 'h4n'], 'h5n'], ['*', 8.0, 'h2n', 'h3n', ['*', 'h4n', 'h4n'], ['*', 'h5n', 'h5n']], ['*', 12.0, 'h2n', 'h3n', ['*', 'h4n', 'h4n'], 'h5n', 'h6n'], ['*', 4.0, 'h2n', 'h3n', 'h4n', ['*', 'h5n', 'h5n', 'h5n']], ['*', 16.0, 'h2n', 'h3n', 'h4n', ['*', 'h5n', 'h5n'], 'h6n'], ['*', 12.0, 'h2n', 'h3n', 'h4n', 'h5n', ['*', 'h6n', 'h6n']], ['*', 4.0, 'h2n', 'h3n', ['*', 'h5n', 'h5n', 'h5n'], 'h6n'], ['*', 8.0, 'h2n', 'h3n', ['*', 'h5n', 'h5n'], ['*', 'h6n', 'h6n']], ['*', 4.0, 'h2n', 'h3n', 'h5n', ['*', 'h6n', 'h6n', 'h6n']], ['*', 'h2n', ['*', 'h4n', 'h4n', 'h4n'], ['*', 'h5n', 'h5n']], ['*', 'h2n', ['*', 'h4n', 'h4n'], ['*', 'h5n', 'h5n', 'h5n']], ['*', 3.0, 'h2n', ['*', 'h4n', 'h4n'], ['*', 'h5n', 'h5n'], 'h6n'], ['*', 2.0, 'h2n', 'h4n', ['*', 'h5n', 'h5n', 'h5n'], 'h6n'], ['*', 3.0, 'h2n', 'h4n', ['*', 'h5n', 'h5n'], ['*', 'h6n', 'h6n']], ['*', 'h2n', ['*', 'h5n', 'h5n', 'h5n'], ['*', 'h6n', 'h6n']], ['*', 'h2n', ['*', 'h5n', 'h5n'], ['*', 'h6n', 'h6n', 'h6n']], ['*', 8.0, ['*', 'h3n', 'h3n', 'h3n'], ['*', 'h4n', 'h4n'], 'h5n'], ['*', 8.0, ['*', 'h3n', 'h3n', 'h3n'], 'h4n', ['*', 'h5n', 'h5n']], ['*', 16.0, ['*', 'h3n', 'h3n', 'h3n'], 'h4n', 'h5n', 'h6n'], ['*', 8.0, ['*', 'h3n', 'h3n', 'h3n'], ['*', 'h5n', 'h5n'], 'h6n'], ['*', 8.0, ['*', 'h3n', 'h3n', 'h3n'], 'h5n', ['*', 'h6n', 'h6n']], ['*', 4.0, ['*', 'h3n', 'h3n'], ['*', 'h4n', 'h4n', 'h4n'], 'h5n'], ['*', 8.0, ['*', 'h3n', 'h3n'], ['*', 'h4n', 'h4n'], ['*', 'h5n', 'h5n']], ['*', 12.0, ['*', 'h3n', 'h3n'], ['*', 'h4n', 'h4n'], 'h5n', 'h6n'], ['*', 4.0, ['*', 'h3n', 'h3n'], 'h4n', ['*', 'h5n', 'h5n', 'h5n']], ['*', 16.0, ['*', 'h3n', 'h3n'], 'h4n', ['*', 'h5n', 'h5n'], 'h6n'], ['*', 12.0, ['*', 'h3n', 'h3n'], 'h4n', 'h5n', ['*', 'h6n', 'h6n']], ['*', 4.0, ['*', 'h3n', 'h3n'], ['*', 'h5n', 'h5n', 'h5n'], 'h6n'], ['*', 8.0, ['*', 'h3n', 'h3n'], ['*', 'h5n', 'h5n'], ['*', 'h6n', 'h6n']], ['*', 4.0, ['*', 'h3n', 'h3n'], 'h5n', ['*', 'h6n', 'h6n', 'h6n']], ['*', 2.0, 'h3n', ['*', 'h4n', 'h4n', 'h4n'], ['*', 'h5n', 'h5n']], ['*', 2.0, 'h3n', ['*', 'h4n', 'h4n'], ['*', 'h5n', 'h5n', 'h5n']], ['*', 6.0, 'h3n', ['*', 'h4n', 'h4n'], ['*', 'h5n', 'h5n'], 'h6n'], ['*', 4.0, 'h3n', 'h4n', ['*', 'h5n', 'h5n', 'h5n'], 'h6n'], ['*', 6.0, 'h3n', 'h4n', ['*', 'h5n', 'h5n'], ['*', 'h6n', 'h6n']], ['*', 2.0, 'h3n', ['*', 'h5n', 'h5n', 'h5n'], ['*', 'h6n', 'h6n']], ['*', 2.0, 'h3n', ['*', 'h5n', 'h5n'], ['*', 'h6n', 'h6n', 'h6n']]]
]

Another example:
[
['-', 0.0, ['+', ['*', ['*', 'h1p', 'h1p', 'h1p'], ['*', 'h3p', 'h3p'], 'h5p', ['*', 'j2p', 'j2p']], ['*', ['*', 'h1p', 'h1p', 'h1p'], ['*', 'h3p', 'h3p'], 'h5p', 'j2p'], ['*', 4.0, ['*', 'h1p', 'h1p', 'h1p'], ['*', 'h3p', 'h3p'], 'h6p', ['*', 'j2p', 'j2p']], ['*', 4.0, ['*', 'h1p', 'h1p', 'h1p'], ['*', 'h3p', 'h3p'], 'h6p', 'j2p'], ['*', ['*', 'h1p', 'h1p', 'h1p'], 'h3p', ['*', 'h5p', 'h5p'], ['*', 'j2p', 'j2p']], ['*', ['*', 'h1p', 'h1p', 'h1p'], 'h3p', 'h5p', 'h6p', ['*', 'j2p', 'j2p', 'j2p']], ['*', 5.0, ['*', 'h1p', 'h1p', 'h1p'], 'h3p', 'h5p', 'h6p', ['*', 'j2p', 'j2p']], ['*', 2.0, ['*', 'h1p', 'h1p', 'h1p'], 'h3p', 'h5p', 'h6p', 'j2p'], ['*', 4.0, ['*', 'h1p', 'h1p', 'h1p'], 'h3p', ['*', 'h6p', 'h6p'], ['*', 'j2p', 'j2p']], ['*', 4.0, ['*', 'h1p', 'h1p', 'h1p'], 'h3p', ['*', 'h6p', 'h6p'], 'j2p'], ['*', ['*', 'h1p', 'h1p', 'h1p'], ['*', 'h5p', 'h5p'], 'h6p', ['*', 'j2p', 'j2p', 'j2p']], ['*', ['*', 'h1p', 'h1p', 'h1p'], ['*', 'h5p', 'h5p'], 'h6p', ['*', 'j2p', 'j2p']], ['*', ['*', 'h1p', 'h1p', 'h1p'], 'h5p', ['*', 'h6p', 'h6p'], ['*', 'j2p', 'j2p', 'j2p']], ['*', 2.0, ['*', 'h1p', 'h1p', 'h1p'], 'h5p', ['*', 'h6p', 'h6p'], ['*', 'j2p', 'j2p']], ['*', ['*', 'h1p', 'h1p', 'h1p'], 'h5p', ['*', 'h6p', 'h6p'], 'j2p'], ['*', 5.0, ['*', 'h1p', 'h1p'], 'h2p', ['*', 'h3p', 'h3p'], 'h5p', ['*', 'j2p', 'j2p']], ['*', 6.0, ['*', 'h1p', 'h1p'], 'h2p', ['*', 'h3p', 'h3p'], 'h5p', 'j2p'], ['*', ['*', 'h1p', 'h1p'], 'h2p', ['*', 'h3p', 'h3p'], 'h5p'], ['*', 12.0, ['*', 'h1p', 'h1p'], 'h2p', ['*', 'h3p', 'h3p'], 'h6p', ['*', 'j2p', 'j2p']], ['*', 12.0, ['*', 'h1p', 'h1p'], 'h2p', ['*', 'h3p', 'h3p'], 'h6p', 'j2p'], ['*', 5.0, ['*', 'h1p', 'h1p'], 'h2p', 'h3p', ['*', 'h5p', 'h5p'], ['*', 'j2p', 'j2p']], ['*', 2.0, ['*', 'h1p', 'h1p'], 'h2p', 'h3p', ['*', 'h5p', 'h5p'], 'j2p'], ['*', 3.0, ['*', 'h1p', 'h1p'], 'h2p', 'h3p', 'h5p', 'h6p', ['*', 'j2p', 'j2p', 'j2p']], ['*', 16.0, ['*', 'h1p', 'h1p'], 'h2p', 'h3p', 'h5p', 'h6p', ['*', 'j2p', 'j2p']], ['*', 11.0, ['*', 'h1p', 'h1p'], 'h2p', 'h3p', 'h5p', 'h6p', 'j2p'], ['*', 2.0, ['*', 'h1p', 'h1p'], 'h2p', 'h3p', 'h5p', 'h6p'], ['*', 12.0, ['*', 'h1p', 'h1p'], 'h2p', 'h3p', ['*', 'h6p', 'h6p'], ['*', 'j2p', 'j2p']], ['*', 12.0, ['*', 'h1p', 'h1p'], 'h2p', 'h3p', ['*', 'h6p', 'h6p'], 'j2p'], ['*', 3.0, ['*', 'h1p', 'h1p'], 'h2p', ['*', 'h5p', 'h5p'], 'h6p', ['*', 'j2p', 'j2p', 'j2p']], ['*', 5.0, ['*', 'h1p', 'h1p'], 'h2p', ['*', 'h5p', 'h5p'], 'h6p', ['*', 'j2p', 'j2p']], ['*', 2.0, ['*', 'h1p', 'h1p'], 'h2p', ['*', 'h5p', 'h5p'], 'h6p', 'j2p'], ['*', 3.0, ['*', 'h1p', 'h1p'], 'h2p', 'h5p', ['*', 'h6p', 'h6p'], ['*', 'j2p', 'j2p', 'j2p']], ['*', 7.0, ['*', 'h1p', 'h1p'], 'h2p', 'h5p', ['*', 'h6p', 'h6p'], ['*', 'j2p', 'j2p']], ['*', 5.0, ['*', 'h1p', 'h1p'], 'h2p', 'h5p', ['*', 'h6p', 'h6p'], 'j2p'], ['*', ['*', 'h1p', 'h1p'], 'h2p', 'h5p', ['*', 'h6p', 'h6p']], ['*', ['*', 'h1p', 'h1p'], ['*', 'h3p', 'h3p', 'h3p'], 'h5p', ['*', 'j2p', 'j2p', 'j2p']], ['*', 2.0, ['*', 'h1p', 'h1p'], ['*', 'h3p', 'h3p', 'h3p'], 'h5p', ['*', 'j2p', 'j2p']], ['*', ['*', 'h1p', 'h1p'], ['*', 'h3p', 'h3p', 'h3p'], 'h5p', 'j2p'], ['*', 4.0, ['*', 'h1p', 'h1p'], ['*', 'h3p', 'h3p', 'h3p'], 'h6p', ['*', 'j2p', 'j2p', 'j2p']], ['*', 8.0, ['*', 'h1p', 'h1p'], ['*', 'h3p', 'h3p', 'h3p'], 'h6p', ['*', 'j2p', 'j2p']], ['*', 4.0, ['*', 'h1p', 'h1p'], ['*', 'h3p', 'h3p', 'h3p'], 'h6p', 'j2p'], ['*', 2.0, ['*', 'h1p', 'h1p'], ['*', 'h3p', 'h3p'], ['*', 'h5p', 'h5p'], ['*', 'j2p', 'j2p', 'j2p']], ['*', 4.0, ['*', 'h1p', 'h1p'], ['*', 'h3p', 'h3p'], ['*', 'h5p', 'h5p'], ['*', 'j2p', 'j2p']], ['*', 2.0, ['*', 'h1p', 'h1p'], ['*', 'h3p', 'h3p'], ['*', 'h5p', 'h5p'], 'j2p'], ['*', ['*', 'h1p', 'h1p'], ['*', 'h3p', 'h3p'], 'h5p', 'h6p', ['*', 'j2p', 'j2p', 'j2p', 'j2p']], ['*', 12.0, ['*', 'h1p', 'h1p'], ['*', 'h3p', 'h3p'], 'h5p', 'h6p', ['*', 'j2p', 'j2p', 'j2p']], ['*', 25.0, ['*', 'h1p', 'h1p'], ['*', 'h3p', 'h3p'], 'h5p', 'h6p', ['*', 'j2p', 'j2p']], ['*', 14.0, ['*', 'h1p', 'h1p'], ['*', 'h3p', 'h3p'], 'h5p', 'h6p', 'j2p'], ['*', 8.0, ['*', 'h1p', 'h1p'], ['*', 'h3p', 'h3p'], ['*', 'h6p', 'h6p'], ['*', 'j2p', 'j2p', 'j2p']], ['*', 16.0, ['*', 'h1p', 'h1p'], ['*', 'h3p', 'h3p'], ['*', 'h6p', 'h6p'], ['*', 'j2p', 'j2p']], ['*', 8.0, ['*', 'h1p', 'h1p'], ['*', 'h3p', 'h3p'], ['*', 'h6p', 'h6p'], 'j2p'], ['*', ['*', 'h1p', 'h1p'], 'h3p', ['*', 'h5p', 'h5p', 'h5p'], ['*', 'j2p', 'j2p', 'j2p']], ['*', ['*', 'h1p', 'h1p'], 'h3p', ['*', 'h5p', 'h5p', 'h5p'], ['*', 'j2p', 'j2p']], ['*', 2.0, ['*', 'h1p', 'h1p'], 'h3p', ['*', 'h5p', 'h5p'], 'h6p', ['*', 'j2p', 'j2p', 'j2p', 'j2p']], ['*', 11.0, ['*', 'h1p', 'h1p'], 'h3p', ['*', 'h5p', 'h5p'], 'h6p', ['*', 'j2p', 'j2p', 'j2p']], ['*', 13.0, ['*', 'h1p', 'h1p'], 'h3p', ['*', 'h5p', 'h5p'], 'h6p', ['*', 'j2p', 'j2p']], ['*', 4.0, ['*', 'h1p', 'h1p'], 'h3p', ['*', 'h5p', 'h5p'], 'h6p', 'j2p'], ['*', 2.0, ['*', 'h1p', 'h1p'], 'h3p', 'h5p', ['*', 'h6p', 'h6p'], ['*', 'j2p', 'j2p', 'j2p', 'j2p']], ['*', 14.0, ['*', 'h1p', 'h1p'], 'h3p', 'h5p', ['*', 'h6p', 'h6p'], ['*', 'j2p', 'j2p', 'j2p']], ['*', 26.0, ['*', 'h1p', 'h1p'], 'h3p', 'h5p', ['*', 'h6p', 'h6p'], ['*', 'j2p', 'j2p']], ['*', 14.0, ['*', 'h1p', 'h1p'], 'h3p', 'h5p', ['*', 'h6p', 'h6p'], 'j2p'], ['*', 4.0, ['*', 'h1p', 'h1p'], 'h3p', ['*', 'h6p', 'h6p', 'h6p'], ['*', 'j2p', 'j2p', 'j2p']], ['*', 8.0, ['*', 'h1p', 'h1p'], 'h3p', ['*', 'h6p', 'h6p', 'h6p'], ['*', 'j2p', 'j2p']], ['*', 4.0, ['*', 'h1p', 'h1p'], 'h3p', ['*', 'h6p', 'h6p', 'h6p'], 'j2p'], ['*', ['*', 'h1p', 'h1p'], ['*', 'h5p', 'h5p', 'h5p'], 'h6p', ['*', 'j2p', 'j2p', 'j2p', 'j2p']], ['*', 2.0, ['*', 'h1p', 'h1p'], ['*', 'h5p', 'h5p', 'h5p'], 'h6p', ['*', 'j2p', 'j2p', 'j2p']], ['*', ['*', 'h1p', 'h1p'], ['*', 'h5p', 'h5p', 'h5p'], 'h6p', ['*', 'j2p', 'j2p']], ['*', 2.0, ['*', 'h1p', 'h1p'], ['*', 'h5p', 'h5p'], ['*', 'h6p', 'h6p'], ['*', 'j2p', 'j2p', 'j2p', 'j2p']], ['*', 6.0, ['*', 'h1p', 'h1p'], ['*', 'h5p', 'h5p'], ['*', 'h6p', 'h6p'], ['*', 'j2p', 'j2p', 'j2p']], ['*', 6.0, ['*', 'h1p', 'h1p'], ['*', 'h5p', 'h5p'], ['*', 'h6p', 'h6p'], ['*', 'j2p', 'j2p']], ['*', 2.0, ['*', 'h1p', 'h1p'], ['*', 'h5p', 'h5p'], ['*', 'h6p', 'h6p'], 'j2p'], ['*', ['*', 'h1p', 'h1p'], 'h5p', ['*', 'h6p', 'h6p', 'h6p'], ['*', 'j2p', 'j2p', 'j2p', 'j2p']], ['*', 3.0, ['*', 'h1p', 'h1p'], 'h5p', ['*', 'h6p', 'h6p', 'h6p'], ['*', 'j2p', 'j2p', 'j2p']], ['*', 3.0, ['*', 'h1p', 'h1p'], 'h5p', ['*', 'h6p', 'h6p', 'h6p'], ['*', 'j2p', 'j2p']], ['*', ['*', 'h1p', 'h1p'], 'h5p', ['*', 'h6p', 'h6p', 'h6p'], 'j2p'], ['*', 7.0, 'h1p', ['*', 'h2p', 'h2p'], ['*', 'h3p', 'h3p'], 'h5p', ['*', 'j2p', 'j2p']], ['*', 9.0, 'h1p', ['*', 'h2p', 'h2p'], ['*', 'h3p', 'h3p'], 'h5p', 'j2p'], ['*', 2.0, 'h1p', ['*', 'h2p', 'h2p'], ['*', 'h3p', 'h3p'], 'h5p'], ['*', 12.0, 'h1p', ['*', 'h2p', 'h2p'], ['*', 'h3p', 'h3p'], 'h6p', ['*', 'j2p', 'j2p']], ['*', 12.0, 'h1p', ['*', 'h2p', 'h2p'], ['*', 'h3p', 'h3p'], 'h6p', 'j2p'], ['*', 7.0, 'h1p', ['*', 'h2p', 'h2p'], 'h3p', ['*', 'h5p', 'h5p'], ['*', 'j2p', 'j2p']], ['*', 6.0, 'h1p', ['*', 'h2p', 'h2p'], 'h3p', ['*', 'h5p', 'h5p'], 'j2p'], ['*', 'h1p', ['*', 'h2p', 'h2p'], 'h3p', ['*', 'h5p', 'h5p']], ['*', 3.0, 'h1p', ['*', 'h2p', 'h2p'], 'h3p', 'h5p', 'h6p', ['*', 'j2p', 'j2p', 'j2p']], ['*', 17.0, 'h1p', ['*', 'h2p', 'h2p'], 'h3p', 'h5p', 'h6p', ['*', 'j2p', 'j2p']], ['*', 16.0, 'h1p', ['*', 'h2p', 'h2p'], 'h3p', 'h5p', 'h6p', 'j2p'], ['*', 4.0, 'h1p', ['*', 'h2p', 'h2p'], 'h3p', 'h5p', 'h6p'], ['*', 12.0, 'h1p', ['*', 'h2p', 'h2p'], 'h3p', ['*', 'h6p', 'h6p'], ['*', 'j2p', 'j2p']], ['*', 12.0, 'h1p', ['*', 'h2p', 'h2p'], 'h3p', ['*', 'h6p', 'h6p'], 'j2p'], ['*', 3.0, 'h1p', ['*', 'h2p', 'h2p'], ['*', 'h5p', 'h5p'], 'h6p', ['*', 'j2p', 'j2p', 'j2p']], ['*', 7.0, 'h1p', ['*', 'h2p', 'h2p'], ['*', 'h5p', 'h5p'], 'h6p', ['*', 'j2p', 'j2p']], ['*', 5.0, 'h1p', ['*', 'h2p', 'h2p'], ['*', 'h5p', 'h5p'], 'h6p', 'j2p'], ['*', 'h1p', ['*', 'h2p', 'h2p'], ['*', 'h5p', 'h5p'], 'h6p'], ['*', 3.0, 'h1p', ['*', 'h2p', 'h2p'], 'h5p', ['*', 'h6p', 'h6p'], ['*', 'j2p', 'j2p', 'j2p']], ['*', 8.0, 'h1p', ['*', 'h2p', 'h2p'], 'h5p', ['*', 'h6p', 'h6p'], ['*', 'j2p', 'j2p']], ['*', 7.0, 'h1p', ['*', 'h2p', 'h2p'], 'h5p', ['*', 'h6p', 'h6p'], 'j2p'], ['*', 2.0, 'h1p', ['*', 'h2p', 'h2p'], 'h5p', ['*', 'h6p', 'h6p']], ['*', 4.0, 'h1p', 'h2p', ['*', 'h3p', 'h3p', 'h3p'], 'h5p', ['*', 'j2p', 'j2p', 'j2p']], ['*', 9.0, 'h1p', 'h2p', ['*', 'h3p', 'h3p', 'h3p'], 'h5p', ['*', 'j2p', 'j2p']], ['*', 6.0, 'h1p', 'h2p', ['*', 'h3p', 'h3p', 'h3p'], 'h5p', 'j2p'], ['*', 'h1p', 'h2p', ['*', 'h3p', 'h3p', 'h3p'], 'h5p'], ['*', 8.0, 'h1p', 'h2p', ['*', 'h3p', 'h3p', 'h3p'], 'h6p', ['*', 'j2p', 'j2p', 'j2p']], ['*', 16.0, 'h1p', 'h2p', ['*', 'h3p', 'h3p', 'h3p'], 'h6p', ['*', 'j2p', 'j2p']], ['*', 8.0, 'h1p', 'h2p', ['*', 'h3p', 'h3p', 'h3p'], 'h6p', 'j2p'], ['*', 8.0, 'h1p', 'h2p', ['*', 'h3p', 'h3p'], ['*', 'h5p', 'h5p'], ['*', 'j2p', 'j2p', 'j2p']], ['*', 17.0, 'h1p', 'h2p', ['*', 'h3p', 'h3p'], ['*', 'h5p', 'h5p'], ['*', 'j2p', 'j2p']], ['*', 12.0, 'h1p', 'h2p', ['*', 'h3p', 'h3p'], ['*', 'h5p', 'h5p'], 'j2p'], ['*', 2.0, 'h1p', 'h2p', ['*', 'h3p', 'h3p'], ['*', 'h5p', 'h5p']], ['*', 2.0, 'h1p', 'h2p', ['*', 'h3p', 'h3p'], 'h5p', 'h6p', ['*', 'j2p', 'j2p', 'j2p', 'j2p']], ['*', 25.0, 'h1p', 'h2p', ['*', 'h3p', 'h3p'], 'h5p', 'h6p', ['*', 'j2p', 'j2p', 'j2p']], ['*', 47.0, 'h1p', 'h2p', ['*', 'h3p', 'h3p'], 'h5p', 'h6p', ['*', 'j2p', 'j2p']], ['*', 31.0, 'h1p', 'h2p', ['*', 'h3p', 'h3p'], 'h5p', 'h6p', 'j2p'], ['*', 3.0, 'h1p', 'h2p', ['*', 'h3p', 'h3p'], 'h5p', 'h6p'], ['*', 16.0, 'h1p', 'h2p', ['*', 'h3p', 'h3p'], ['*', 'h6p', 'h6p'], ['*', 'j2p', 'j2p', 'j2p']], ['*', 32.0, 'h1p', 'h2p', ['*', 'h3p', 'h3p'], ['*', 'h6p', 'h6p'], ['*', 'j2p', 'j2p']], ['*', 16.0, 'h1p', 'h2p', ['*', 'h3p', 'h3p'], ['*', 'h6p', 'h6p'], 'j2p'], ['*', 4.0, 'h1p', 'h2p', 'h3p', ['*', 'h5p', 'h5p', 'h5p'], ['*', 'j2p', 'j2p', 'j2p']], ['*', 6.0, 'h1p', 'h2p', 'h3p', ['*', 'h5p', 'h5p', 'h5p'], ['*', 'j2p', 'j2p']], ['*', 2.0, 'h1p', 'h2p', 'h3p', ['*', 'h5p', 'h5p', 'h5p'], 'j2p'], ['*', 4.0, 'h1p', 'h2p', 'h3p', ['*', 'h5p', 'h5p'], 'h6p', ['*', 'j2p', 'j2p', 'j2p', 'j2p']], ['*', 23.0, 'h1p', 'h2p', 'h3p', ['*', 'h5p', 'h5p'], 'h6p', ['*', 'j2p', 'j2p', 'j2p']], ['*', 36.0, 'h1p', 'h2p', 'h3p', ['*', 'h5p', 'h5p'], 'h6p', ['*', 'j2p', 'j2p']], ['*', 21.0, 'h1p', 'h2p', 'h3p', ['*', 'h5p', 'h5p'], 'h6p', 'j2p'], ['*', 4.0, 'h1p', 'h2p', 'h3p', ['*', 'h5p', 'h5p'], 'h6p'], ['*', 4.0, 'h1p', 'h2p', 'h3p', 'h5p', ['*', 'h6p', 'h6p'], ['*', 'j2p', 'j2p', 'j2p', 'j2p']], ['*', 28.0, 'h1p', 'h2p', 'h3p', 'h5p', ['*', 'h6p', 'h6p'], ['*', 'j2p', 'j2p', 'j2p']], ['*', 51.0, 'h1p', 'h2p', 'h3p', 'h5p', ['*', 'h6p', 'h6p'], ['*', 'j2p', 'j2p']], ['*', 30.0, 'h1p', 'h2p', 'h3p', 'h5p', ['*', 'h6p', 'h6p'], 'j2p'], ['*', 3.0, 'h1p', 'h2p', 'h3p', 'h5p', ['*', 'h6p', 'h6p']], ['*', 8.0, 'h1p', 'h2p', 'h3p', ['*', 'h6p', 'h6p', 'h6p'], ['*', 'j2p', 'j2p', 'j2p']], ['*', 16.0, 'h1p', 'h2p', 'h3p', ['*', 'h6p', 'h6p', 'h6p'], ['*', 'j2p', 'j2p']], ['*', 8.0, 'h1p', 'h2p', 'h3p', ['*', 'h6p', 'h6p', 'h6p'], 'j2p'], ['*', 2.0, 'h1p', 'h2p', ['*', 'h5p', 'h5p', 'h5p'], 'h6p', ['*', 'j2p', 'j2p', 'j2p', 'j2p']], ['*', 6.0, 'h1p', 'h2p', ['*', 'h5p', 'h5p', 'h5p'], 'h6p', ['*', 'j2p', 'j2p', 'j2p']], ['*', 6.0, 'h1p', 'h2p', ['*', 'h5p', 'h5p', 'h5p'], 'h6p', ['*', 'j2p', 'j2p']], ['*', 2.0, 'h1p', 'h2p', ['*', 'h5p', 'h5p', 'h5p'], 'h6p', 'j2p'], ['*', 4.0, 'h1p', 'h2p', ['*', 'h5p', 'h5p'], ['*', 'h6p', 'h6p'], ['*', 'j2p', 'j2p', 'j2p', 'j2p']], ['*', 14.0, 'h1p', 'h2p', ['*', 'h5p', 'h5p'], ['*', 'h6p', 'h6p'], ['*', 'j2p', 'j2p', 'j2p']], ['*', 18.0, 'h1p', 'h2p', ['*', 'h5p', 'h5p'], ['*', 'h6p', 'h6p'], ['*', 'j2p', 'j2p']], ['*', 10.0, 'h1p', 'h2p', ['*', 'h5p', 'h5p'], ['*', 'h6p', 'h6p'], 'j2p'], ['*', 2.0, 'h1p', 'h2p', ['*', 'h5p', 'h5p'], ['*', 'h6p', 'h6p']], ['*', 2.0, 'h1p', 'h2p', 'h5p', ['*', 'h6p', 'h6p', 'h6p'], ['*', 'j2p', 'j2p', 'j2p', 'j2p']], ['*', 7.0, 'h1p', 'h2p', 'h5p', ['*', 'h6p', 'h6p', 'h6p'], ['*', 'j2p', 'j2p', 'j2p']], ['*', 9.0, 'h1p', 'h2p', 'h5p', ['*', 'h6p', 'h6p', 'h6p'], ['*', 'j2p', 'j2p']], ['*', 5.0, 'h1p', 'h2p', 'h5p', ['*', 'h6p', 'h6p', 'h6p'], 'j2p'], ['*', 'h1p', 'h2p', 'h5p', ['*', 'h6p', 'h6p', 'h6p']], ['*', 2.0, 'h1p', ['*', 'h3p', 'h3p', 'h3p'], ['*', 'h5p', 'h5p'], ['*', 'j2p', 'j2p', 'j2p']], ['*', 3.0, 'h1p', ['*', 'h3p', 'h3p', 'h3p'], ['*', 'h5p', 'h5p'], ['*', 'j2p', 'j2p']], ['*', 'h1p', ['*', 'h3p', 'h3p', 'h3p'], ['*', 'h5p', 'h5p'], 'j2p'], ['*', 'h1p', ['*', 'h3p', 'h3p', 'h3p'], 'h5p', 'h6p', ['*', 'j2p', 'j2p', 'j2p', 'j2p']], ['*', 17.0, 'h1p', ['*', 'h3p', 'h3p', 'h3p'], 'h5p', 'h6p', ['*', 'j2p', 'j2p', 'j2p']], ['*', 23.0, 'h1p', ['*', 'h3p', 'h3p', 'h3p'], 'h5p', 'h6p', ['*', 'j2p', 'j2p']], ['*', 7.0, 'h1p', ['*', 'h3p', 'h3p', 'h3p'], 'h5p', 'h6p', 'j2p'], ['*', 16.0, 'h1p', ['*', 'h3p', 'h3p', 'h3p'], ['*', 'h6p', 'h6p'], ['*', 'j2p', 'j2p', 'j2p']], ['*', 16.0, 'h1p', ['*', 'h3p', 'h3p', 'h3p'], ['*', 'h6p', 'h6p'], ['*', 'j2p', 'j2p']], ['*', 2.0, 'h1p', ['*', 'h3p', 'h3p'], ['*', 'h5p', 'h5p', 'h5p'], ['*', 'j2p', 'j2p', 'j2p']], ['*', 3.0, 'h1p', ['*', 'h3p', 'h3p'], ['*', 'h5p', 'h5p', 'h5p'], ['*', 'j2p', 'j2p']], ['*', 'h1p', ['*', 'h3p', 'h3p'], ['*', 'h5p', 'h5p', 'h5p'], 'j2p'], ['*', 5.0, 'h1p', ['*', 'h3p', 'h3p'], ['*', 'h5p', 'h5p'], 'h6p', ['*', 'j2p', 'j2p', 'j2p', 'j2p']], ['*', 29.0, 'h1p', ['*', 'h3p', 'h3p'], ['*', 'h5p', 'h5p'], 'h6p', ['*', 'j2p', 'j2p', 'j2p']], ['*', 37.0, 'h1p', ['*', 'h3p', 'h3p'], ['*', 'h5p', 'h5p'], 'h6p', ['*', 'j2p', 'j2p']], ['*', 13.0, 'h1p', ['*', 'h3p', 'h3p'], ['*', 'h5p', 'h5p'], 'h6p', 'j2p'], ['*', 10.0, 'h1p', ['*', 'h3p', 'h3p'], 'h5p', ['*', 'h6p', 'h6p'], ['*', 'j2p', 'j2p', 'j2p', 'j2p']], ['*', 46.0, 'h1p', ['*', 'h3p', 'h3p'], 'h5p', ['*', 'h6p', 'h6p'], ['*', 'j2p', 'j2p', 'j2p']], ['*', 50.0, 'h1p', ['*', 'h3p', 'h3p'], 'h5p', ['*', 'h6p', 'h6p'], ['*', 'j2p', 'j2p']], ['*', 14.0, 'h1p', ['*', 'h3p', 'h3p'], 'h5p', ['*', 'h6p', 'h6p'], 'j2p'], ['*', 16.0, 'h1p', ['*', 'h3p', 'h3p'], ['*', 'h6p', 'h6p', 'h6p'], ['*', 'j2p', 'j2p', 'j2p']], ['*', 16.0, 'h1p', ['*', 'h3p', 'h3p'], ['*', 'h6p', 'h6p', 'h6p'], ['*', 'j2p', 'j2p']], ['*', 4.0, 'h1p', 'h3p', ['*', 'h5p', 'h5p', 'h5p'], 'h6p', ['*', 'j2p', 'j2p', 'j2p', 'j2p']], ['*', 10.0, 'h1p', 'h3p', ['*', 'h5p', 'h5p', 'h5p'], 'h6p', ['*', 'j2p', 'j2p', 'j2p']], ['*', 8.0, 'h1p', 'h3p', ['*', 'h5p', 'h5p', 'h5p'], 'h6p', ['*', 'j2p', 'j2p']], ['*', 2.0, 'h1p', 'h3p', ['*', 'h5p', 'h5p', 'h5p'], 'h6p', 'j2p'], ['*', 'h1p', 'h3p', ['*', 'h5p', 'h5p'], ['*', 'h6p', 'h6p'], ['*', 'j2p', 'j2p', 'j2p', 'j2p', 'j2p']], ['*', 14.0, 'h1p', 'h3p', ['*', 'h5p', 'h5p'], ['*', 'h6p', 'h6p'], ['*', 'j2p', 'j2p', 'j2p', 'j2p']], ['*', 38.0, 'h1p', 'h3p', ['*', 'h5p', 'h5p'], ['*', 'h6p', 'h6p'], ['*', 'j2p', 'j2p', 'j2p']], ['*', 38.0, 'h1p', 'h3p', ['*', 'h5p', 'h5p'], ['*', 'h6p', 'h6p'], ['*', 'j2p', 'j2p']], ['*', 13.0, 'h1p', 'h3p', ['*', 'h5p', 'h5p'], ['*', 'h6p', 'h6p'], 'j2p'], ['*', 9.0, 'h1p', 'h3p', 'h5p', ['*', 'h6p', 'h6p', 'h6p'], ['*', 'j2p', 'j2p', 'j2p', 'j2p']], ['*', 25.0, 'h1p', 'h3p', 'h5p', ['*', 'h6p', 'h6p', 'h6p'], ['*', 'j2p', 'j2p', 'j2p']], ['*', 23.0, 'h1p', 'h3p', 'h5p', ['*', 'h6p', 'h6p', 'h6p'], ['*', 'j2p', 'j2p']], ['*', 7.0, 'h1p', 'h3p', 'h5p', ['*', 'h6p', 'h6p', 'h6p'], 'j2p'], ['*', 'h1p', ['*', 'h5p', 'h5p', 'h5p'], ['*', 'h6p', 'h6p'], ['*', 'j2p', 'j2p', 'j2p', 'j2p', 'j2p']], ['*', 4.0, 'h1p', ['*', 'h5p', 'h5p', 'h5p'], ['*', 'h6p', 'h6p'], ['*', 'j2p', 'j2p', 'j2p', 'j2p']], ['*', 6.0, 'h1p', ['*', 'h5p', 'h5p', 'h5p'], ['*', 'h6p', 'h6p'], ['*', 'j2p', 'j2p', 'j2p']], ['*', 4.0, 'h1p', ['*', 'h5p', 'h5p', 'h5p'], ['*', 'h6p', 'h6p'], ['*', 'j2p', 'j2p']], ['*', 'h1p', ['*', 'h5p', 'h5p', 'h5p'], ['*', 'h6p', 'h6p'], 'j2p'], ['*', 'h1p', ['*', 'h5p', 'h5p'], ['*', 'h6p', 'h6p', 'h6p'], ['*', 'j2p', 'j2p', 'j2p', 'j2p', 'j2p']], ['*', 4.0, 'h1p', ['*', 'h5p', 'h5p'], ['*', 'h6p', 'h6p', 'h6p'], ['*', 'j2p', 'j2p', 'j2p', 'j2p']], ['*', 6.0, 'h1p', ['*', 'h5p', 'h5p'], ['*', 'h6p', 'h6p', 'h6p'], ['*', 'j2p', 'j2p', 'j2p']], ['*', 4.0, 'h1p', ['*', 'h5p', 'h5p'], ['*', 'h6p', 'h6p', 'h6p'], ['*', 'j2p', 'j2p']], ['*', 'h1p', ['*', 'h5p', 'h5p'], ['*', 'h6p', 'h6p', 'h6p'], 'j2p'], ['*', 3.0, ['*', 'h2p', 'h2p', 'h2p'], ['*', 'h3p', 'h3p'], 'h5p', ['*', 'j2p', 'j2p']], ['*', 4.0, ['*', 'h2p', 'h2p', 'h2p'], ['*', 'h3p', 'h3p'], 'h5p', 'j2p'], ['*', ['*', 'h2p', 'h2p', 'h2p'], ['*', 'h3p', 'h3p'], 'h5p'], ['*', 4.0, ['*', 'h2p', 'h2p', 'h2p'], ['*', 'h3p', 'h3p'], 'h6p', ['*', 'j2p', 'j2p']], ['*', 4.0, ['*', 'h2p', 'h2p', 'h2p'], ['*', 'h3p', 'h3p'], 'h6p', 'j2p'], ['*', 3.0, ['*', 'h2p', 'h2p', 'h2p'], 'h3p', ['*', 'h5p', 'h5p'], ['*', 'j2p', 'j2p']], ['*', 4.0, ['*', 'h2p', 'h2p', 'h2p'], 'h3p', ['*', 'h5p', 'h5p'], 'j2p'], ['*', ['*', 'h2p', 'h2p', 'h2p'], 'h3p', ['*', 'h5p', 'h5p']], ['*', ['*', 'h2p', 'h2p', 'h2p'], 'h3p', 'h5p', 'h6p', ['*', 'j2p', 'j2p', 'j2p']], ['*', 6.0, ['*', 'h2p', 'h2p', 'h2p'], 'h3p', 'h5p', 'h6p', ['*', 'j2p', 'j2p']], ['*', 7.0, ['*', 'h2p', 'h2p', 'h2p'], 'h3p', 'h5p', 'h6p', 'j2p'], ['*', 2.0, ['*', 'h2p', 'h2p', 'h2p'], 'h3p', 'h5p', 'h6p'], ['*', 4.0, ['*', 'h2p', 'h2p', 'h2p'], 'h3p', ['*', 'h6p', 'h6p'], ['*', 'j2p', 'j2p']], ['*', 4.0, ['*', 'h2p', 'h2p', 'h2p'], 'h3p', ['*', 'h6p', 'h6p'], 'j2p'], ['*', ['*', 'h2p', 'h2p', 'h2p'], ['*', 'h5p', 'h5p'], 'h6p', ['*', 'j2p', 'j2p', 'j2p']], ['*', 3.0, ['*', 'h2p', 'h2p', 'h2p'], ['*', 'h5p', 'h5p'], 'h6p', ['*', 'j2p', 'j2p']], ['*', 3.0, ['*', 'h2p', 'h2p', 'h2p'], ['*', 'h5p', 'h5p'], 'h6p', 'j2p'], ['*', ['*', 'h2p', 'h2p', 'h2p'], ['*', 'h5p', 'h5p'], 'h6p'], ['*', ['*', 'h2p', 'h2p', 'h2p'], 'h5p', ['*', 'h6p', 'h6p'], ['*', 'j2p', 'j2p', 'j2p']], ['*', 3.0, ['*', 'h2p', 'h2p', 'h2p'], 'h5p', ['*', 'h6p', 'h6p'], ['*', 'j2p', 'j2p']], ['*', 3.0, ['*', 'h2p', 'h2p', 'h2p'], 'h5p', ['*', 'h6p', 'h6p'], 'j2p'], ['*', ['*', 'h2p', 'h2p', 'h2p'], 'h5p', ['*', 'h6p', 'h6p']], ['*', 3.0, ['*', 'h2p', 'h2p'], ['*', 'h3p', 'h3p', 'h3p'], 'h5p', ['*', 'j2p', 'j2p', 'j2p']], ['*', 7.0, ['*', 'h2p', 'h2p'], ['*', 'h3p', 'h3p', 'h3p'], 'h5p', ['*', 'j2p', 'j2p']], ['*', 5.0, ['*', 'h2p', 'h2p'], ['*', 'h3p', 'h3p', 'h3p'], 'h5p', 'j2p'], ['*', ['*', 'h2p', 'h2p'], ['*', 'h3p', 'h3p', 'h3p'], 'h5p'], ['*', 4.0, ['*', 'h2p', 'h2p'], ['*', 'h3p', 'h3p', 'h3p'], 'h6p', ['*', 'j2p', 'j2p', 'j2p']], ['*', 8.0, ['*', 'h2p', 'h2p'], ['*', 'h3p', 'h3p', 'h3p'], 'h6p', ['*', 'j2p', 'j2p']], ['*', 4.0, ['*', 'h2p', 'h2p'], ['*', 'h3p', 'h3p', 'h3p'], 'h6p', 'j2p'], ['*', 6.0, ['*', 'h2p', 'h2p'], ['*', 'h3p', 'h3p'], ['*', 'h5p', 'h5p'], ['*', 'j2p', 'j2p', 'j2p']], ['*', 11.0, ['*', 'h2p', 'h2p'], ['*', 'h3p', 'h3p'], ['*', 'h5p', 'h5p'], ['*', 'j2p', 'j2p']], ['*', 9.0, ['*', 'h2p', 'h2p'], ['*', 'h3p', 'h3p'], ['*', 'h5p', 'h5p'], 'j2p'], ['*', 2.0, ['*', 'h2p', 'h2p'], ['*', 'h3p', 'h3p'], ['*', 'h5p', 'h5p']], ['*', ['*', 'h2p', 'h2p'], ['*', 'h3p', 'h3p'], 'h5p', 'h6p', ['*', 'j2p', 'j2p', 'j2p', 'j2p']], ['*', 13.0, ['*', 'h2p', 'h2p'], ['*', 'h3p', 'h3p'], 'h5p', 'h6p', ['*', 'j2p', 'j2p', 'j2p']], ['*', 22.0, ['*', 'h2p', 'h2p'], ['*', 'h3p', 'h3p'], 'h5p', 'h6p', ['*', 'j2p', 'j2p']], ['*', 17.0, ['*', 'h2p', 'h2p'], ['*', 'h3p', 'h3p'], 'h5p', 'h6p', 'j2p'], ['*', 3.0, ['*', 'h2p', 'h2p'], ['*', 'h3p', 'h3p'], 'h5p', 'h6p'], ['*', 8.0, ['*', 'h2p', 'h2p'], ['*', 'h3p', 'h3p'], ['*', 'h6p', 'h6p'], ['*', 'j2p', 'j2p', 'j2p']], ['*', 16.0, ['*', 'h2p', 'h2p'], ['*', 'h3p', 'h3p'], ['*', 'h6p', 'h6p'], ['*', 'j2p', 'j2p']], ['*', 8.0, ['*', 'h2p', 'h2p'], ['*', 'h3p', 'h3p'], ['*', 'h6p', 'h6p'], 'j2p'], ['*', 3.0, ['*', 'h2p', 'h2p'], 'h3p', ['*', 'h5p', 'h5p', 'h5p'], ['*', 'j2p', 'j2p', 'j2p']], ['*', 7.0, ['*', 'h2p', 'h2p'], 'h3p', ['*', 'h5p', 'h5p', 'h5p'], ['*', 'j2p', 'j2p']], ['*', 5.0, ['*', 'h2p', 'h2p'], 'h3p', ['*', 'h5p', 'h5p', 'h5p'], 'j2p'], ['*', ['*', 'h2p', 'h2p'], 'h3p', ['*', 'h5p', 'h5p', 'h5p']], ['*', 2.0, ['*', 'h2p', 'h2p'], 'h3p', ['*', 'h5p', 'h5p'], 'h6p', ['*', 'j2p', 'j2p', 'j2p', 'j2p']], ['*', 12.0, ['*', 'h2p', 'h2p'], 'h3p', ['*', 'h5p', 'h5p'], 'h6p', ['*', 'j2p', 'j2p', 'j2p']], ['*', 22.0, ['*', 'h2p', 'h2p'], 'h3p', ['*', 'h5p', 'h5p'], 'h6p', ['*', 'j2p', 'j2p']], ['*', 16.0, ['*', 'h2p', 'h2p'], 'h3p', ['*', 'h5p', 'h5p'], 'h6p', 'j2p'], ['*', 4.0, ['*', 'h2p', 'h2p'], 'h3p', ['*', 'h5p', 'h5p'], 'h6p'], ['*', 2.0, ['*', 'h2p', 'h2p'], 'h3p', 'h5p', ['*', 'h6p', 'h6p'], ['*', 'j2p', 'j2p', 'j2p', 'j2p']], ['*', 14.0, ['*', 'h2p', 'h2p'], 'h3p', 'h5p', ['*', 'h6p', 'h6p'], ['*', 'j2p', 'j2p', 'j2p']], ['*', 25.0, ['*', 'h2p', 'h2p'], 'h3p', 'h5p', ['*', 'h6p', 'h6p'], ['*', 'j2p', 'j2p']], ['*', 16.0, ['*', 'h2p', 'h2p'], 'h3p', 'h5p', ['*', 'h6p', 'h6p'], 'j2p'], ['*', 3.0, ['*', 'h2p', 'h2p'], 'h3p', 'h5p', ['*', 'h6p', 'h6p']], ['*', 4.0, ['*', 'h2p', 'h2p'], 'h3p', ['*', 'h6p', 'h6p', 'h6p'], ['*', 'j2p', 'j2p', 'j2p']], ['*', 8.0, ['*', 'h2p', 'h2p'], 'h3p', ['*', 'h6p', 'h6p', 'h6p'], ['*', 'j2p', 'j2p']], ['*', 4.0, ['*', 'h2p', 'h2p'], 'h3p', ['*', 'h6p', 'h6p', 'h6p'], 'j2p'], ['*', ['*', 'h2p', 'h2p'], ['*', 'h5p', 'h5p', 'h5p'], 'h6p', ['*', 'j2p', 'j2p', 'j2p', 'j2p']], ['*', 4.0, ['*', 'h2p', 'h2p'], ['*', 'h5p', 'h5p', 'h5p'], 'h6p', ['*', 'j2p', 'j2p', 'j2p']], ['*', 6.0, ['*', 'h2p', 'h2p'], ['*', 'h5p', 'h5p', 'h5p'], 'h6p', ['*', 'j2p', 'j2p']], ['*', 4.0, ['*', 'h2p', 'h2p'], ['*', 'h5p', 'h5p', 'h5p'], 'h6p', 'j2p'], ['*', ['*', 'h2p', 'h2p'], ['*', 'h5p', 'h5p', 'h5p'], 'h6p'], ['*', 2.0, ['*', 'h2p', 'h2p'], ['*', 'h5p', 'h5p'], ['*', 'h6p', 'h6p'], ['*', 'j2p', 'j2p', 'j2p', 'j2p']], ['*', 8.0, ['*', 'h2p', 'h2p'], ['*', 'h5p', 'h5p'], ['*', 'h6p', 'h6p'], ['*', 'j2p', 'j2p', 'j2p']], ['*', 12.0, ['*', 'h2p', 'h2p'], ['*', 'h5p', 'h5p'], ['*', 'h6p', 'h6p'], ['*', 'j2p', 'j2p']], ['*', 8.0, ['*', 'h2p', 'h2p'], ['*', 'h5p', 'h5p'], ['*', 'h6p', 'h6p'], 'j2p'], ['*', 2.0, ['*', 'h2p', 'h2p'], ['*', 'h5p', 'h5p'], ['*', 'h6p', 'h6p']], ['*', ['*', 'h2p', 'h2p'], 'h5p', ['*', 'h6p', 'h6p', 'h6p'], ['*', 'j2p', 'j2p', 'j2p', 'j2p']], ['*', 4.0, ['*', 'h2p', 'h2p'], 'h5p', ['*', 'h6p', 'h6p', 'h6p'], ['*', 'j2p', 'j2p', 'j2p']], ['*', 6.0, ['*', 'h2p', 'h2p'], 'h5p', ['*', 'h6p', 'h6p', 'h6p'], ['*', 'j2p', 'j2p']], ['*', 4.0, ['*', 'h2p', 'h2p'], 'h5p', ['*', 'h6p', 'h6p', 'h6p'], 'j2p'], ['*', ['*', 'h2p', 'h2p'], 'h5p', ['*', 'h6p', 'h6p', 'h6p']], ['*', 6.0, 'h2p', ['*', 'h3p', 'h3p', 'h3p'], ['*', 'h5p', 'h5p'], ['*', 'j2p', 'j2p', 'j2p']], ['*', 11.0, 'h2p', ['*', 'h3p', 'h3p', 'h3p'], ['*', 'h5p', 'h5p'], ['*', 'j2p', 'j2p']], ['*', 6.0, 'h2p', ['*', 'h3p', 'h3p', 'h3p'], ['*', 'h5p', 'h5p'], 'j2p'], ['*', 'h2p', ['*', 'h3p', 'h3p', 'h3p'], ['*', 'h5p', 'h5p']], ['*', -1.0, ['*', 'h2p', ['*', 'h3p', 'h3p', 'h3p'], 'h5p', 'h6p', ['*', 'j2p', 'j2p', 'j2p', 'j2p']]], ['*', 17.0, 'h2p', ['*', 'h3p', 'h3p', 'h3p'], 'h5p', 'h6p', ['*', 'j2p', 'j2p', 'j2p']], ['*', 25.0, 'h2p', ['*', 'h3p', 'h3p', 'h3p'], 'h5p', 'h6p', ['*', 'j2p', 'j2p']], ['*', 7.0, 'h2p', ['*', 'h3p', 'h3p', 'h3p'], 'h5p', 'h6p', 'j2p'], ['*', 16.0, 'h2p', ['*', 'h3p', 'h3p', 'h3p'], ['*', 'h6p', 'h6p'], ['*', 'j2p', 'j2p', 'j2p']], ['*', 16.0, 'h2p', ['*', 'h3p', 'h3p', 'h3p'], ['*', 'h6p', 'h6p'], ['*', 'j2p', 'j2p']], ['*', 6.0, 'h2p', ['*', 'h3p', 'h3p'], ['*', 'h5p', 'h5p', 'h5p'], ['*', 'j2p', 'j2p', 'j2p']], ['*', 11.0, 'h2p', ['*', 'h3p', 'h3p'], ['*', 'h5p', 'h5p', 'h5p'], ['*', 'j2p', 'j2p']], ['*', 6.0, 'h2p', ['*', 'h3p', 'h3p'], ['*', 'h5p', 'h5p', 'h5p'], 'j2p'], ['*', 'h2p', ['*', 'h3p', 'h3p'], ['*', 'h5p', 'h5p', 'h5p']], ['*', 3.0, 'h2p', ['*', 'h3p', 'h3p'], ['*', 'h5p', 'h5p'], 'h6p', ['*', 'j2p', 'j2p', 'j2p', 'j2p']], ['*', 25.0, 'h2p', ['*', 'h3p', 'h3p'], ['*', 'h5p', 'h5p'], 'h6p', ['*', 'j2p', 'j2p', 'j2p']], ['*', 39.0, 'h2p', ['*', 'h3p', 'h3p'], ['*', 'h5p', 'h5p'], 'h6p', ['*', 'j2p', 'j2p']], ['*', 20.0, 'h2p', ['*', 'h3p', 'h3p'], ['*', 'h5p', 'h5p'], 'h6p', 'j2p'], ['*', 3.0, 'h2p', ['*', 'h3p', 'h3p'], ['*', 'h5p', 'h5p'], 'h6p'], ['*', 6.0, 'h2p', ['*', 'h3p', 'h3p'], 'h5p', ['*', 'h6p', 'h6p'], ['*', 'j2p', 'j2p', 'j2p', 'j2p']], ['*', 42.0, 'h2p', ['*', 'h3p', 'h3p'], 'h5p', ['*', 'h6p', 'h6p'], ['*', 'j2p', 'j2p', 'j2p']], ['*', 50.0, 'h2p', ['*', 'h3p', 'h3p'], 'h5p', ['*', 'h6p', 'h6p'], ['*', 'j2p', 'j2p']], ['*', 14.0, 'h2p', ['*', 'h3p', 'h3p'], 'h5p', ['*', 'h6p', 'h6p'], 'j2p'], ['*', 16.0, 'h2p', ['*', 'h3p', 'h3p'], ['*', 'h6p', 'h6p', 'h6p'], ['*', 'j2p', 'j2p', 'j2p']], ['*', 16.0, 'h2p', ['*', 'h3p', 'h3p'], ['*', 'h6p', 'h6p', 'h6p'], ['*', 'j2p', 'j2p']], ['*', 4.0, 'h2p', 'h3p', ['*', 'h5p', 'h5p', 'h5p'], 'h6p', ['*', 'j2p', 'j2p', 'j2p', 'j2p']], ['*', 14.0, 'h2p', 'h3p', ['*', 'h5p', 'h5p', 'h5p'], 'h6p', ['*', 'j2p', 'j2p', 'j2p']], ['*', 18.0, 'h2p', 'h3p', ['*', 'h5p', 'h5p', 'h5p'], 'h6p', ['*', 'j2p', 'j2p']], ['*', 10.0, 'h2p', 'h3p', ['*', 'h5p', 'h5p', 'h5p'], 'h6p', 'j2p'], ['*', 2.0, 'h2p', 'h3p', ['*', 'h5p', 'h5p', 'h5p'], 'h6p'], ['*', 'h2p', 'h3p', ['*', 'h5p', 'h5p'], ['*', 'h6p', 'h6p'], ['*', 'j2p', 'j2p', 'j2p', 'j2p', 'j2p']], ['*', 13.0, 'h2p', 'h3p', ['*', 'h5p', 'h5p'], ['*', 'h6p', 'h6p'], ['*', 'j2p', 'j2p', 'j2p', 'j2p']], ['*', 36.0, 'h2p', 'h3p', ['*', 'h5p', 'h5p'], ['*', 'h6p', 'h6p'], ['*', 'j2p', 'j2p', 'j2p']], ['*', 40.0, 'h2p', 'h3p', ['*', 'h5p', 'h5p'], ['*', 'h6p', 'h6p'], ['*', 'j2p', 'j2p']], ['*', 19.0, 'h2p', 'h3p', ['*', 'h5p', 'h5p'], ['*', 'h6p', 'h6p'], 'j2p'], ['*', 3.0, 'h2p', 'h3p', ['*', 'h5p', 'h5p'], ['*', 'h6p', 'h6p']], ['*', 7.0, 'h2p', 'h3p', 'h5p', ['*', 'h6p', 'h6p', 'h6p'], ['*', 'j2p', 'j2p', 'j2p', 'j2p']], ['*', 21.0, 'h2p', 'h3p', 'h5p', ['*', 'h6p', 'h6p', 'h6p'], ['*', 'j2p', 'j2p', 'j2p']], ['*', 21.0, 'h2p', 'h3p', 'h5p', ['*', 'h6p', 'h6p', 'h6p'], ['*', 'j2p', 'j2p']], ['*', 7.0, 'h2p', 'h3p', 'h5p', ['*', 'h6p', 'h6p', 'h6p'], 'j2p'], ['*', 'h2p', ['*', 'h5p', 'h5p', 'h5p'], ['*', 'h6p', 'h6p'], ['*', 'j2p', 'j2p', 'j2p', 'j2p', 'j2p']], ['*', 5.0, 'h2p', ['*', 'h5p', 'h5p', 'h5p'], ['*', 'h6p', 'h6p'], ['*', 'j2p', 'j2p', 'j2p', 'j2p']], ['*', 10.0, 'h2p', ['*', 'h5p', 'h5p', 'h5p'], ['*', 'h6p', 'h6p'], ['*', 'j2p', 'j2p', 'j2p']], ['*', 10.0, 'h2p', ['*', 'h5p', 'h5p', 'h5p'], ['*', 'h6p', 'h6p'], ['*', 'j2p', 'j2p']], ['*', 5.0, 'h2p', ['*', 'h5p', 'h5p', 'h5p'], ['*', 'h6p', 'h6p'], 'j2p'], ['*', 'h2p', ['*', 'h5p', 'h5p', 'h5p'], ['*', 'h6p', 'h6p']], ['*', 'h2p', ['*', 'h5p', 'h5p'], ['*', 'h6p', 'h6p', 'h6p'], ['*', 'j2p', 'j2p', 'j2p', 'j2p', 'j2p']], ['*', 5.0, 'h2p', ['*', 'h5p', 'h5p'], ['*', 'h6p', 'h6p', 'h6p'], ['*', 'j2p', 'j2p', 'j2p', 'j2p']], ['*', 10.0, 'h2p', ['*', 'h5p', 'h5p'], ['*', 'h6p', 'h6p', 'h6p'], ['*', 'j2p', 'j2p', 'j2p']], ['*', 10.0, 'h2p', ['*', 'h5p', 'h5p'], ['*', 'h6p', 'h6p', 'h6p'], ['*', 'j2p', 'j2p']], ['*', 5.0, 'h2p', ['*', 'h5p', 'h5p'], ['*', 'h6p', 'h6p', 'h6p'], 'j2p'], ['*', 'h2p', ['*', 'h5p', 'h5p'], ['*', 'h6p', 'h6p', 'h6p']], ['*', 6.0, ['*', 'h3p', 'h3p', 'h3p'], ['*', 'h5p', 'h5p'], 'h6p', ['*', 'j2p', 'j2p', 'j2p', 'j2p']], ['*', 15.0, ['*', 'h3p', 'h3p', 'h3p'], ['*', 'h5p', 'h5p'], 'h6p', ['*', 'j2p', 'j2p', 'j2p']], ['*', 12.0, ['*', 'h3p', 'h3p', 'h3p'], ['*', 'h5p', 'h5p'], 'h6p', ['*', 'j2p', 'j2p']], ['*', 3.0, ['*', 'h3p', 'h3p', 'h3p'], ['*', 'h5p', 'h5p'], 'h6p', 'j2p'], ['*', 12.0, ['*', 'h3p', 'h3p', 'h3p'], 'h5p', ['*', 'h6p', 'h6p'], ['*', 'j2p', 'j2p', 'j2p', 'j2p']], ['*', 24.0, ['*', 'h3p', 'h3p', 'h3p'], 'h5p', ['*', 'h6p', 'h6p'], ['*', 'j2p', 'j2p', 'j2p']], ['*', 12.0, ['*', 'h3p', 'h3p', 'h3p'], 'h5p', ['*', 'h6p', 'h6p'], ['*', 'j2p', 'j2p']], ['*', 6.0, ['*', 'h3p', 'h3p'], ['*', 'h5p', 'h5p', 'h5p'], 'h6p', ['*', 'j2p', 'j2p', 'j2p', 'j2p']], ['*', 15.0, ['*', 'h3p', 'h3p'], ['*', 'h5p', 'h5p', 'h5p'], 'h6p', ['*', 'j2p', 'j2p', 'j2p']], ['*', 12.0, ['*', 'h3p', 'h3p'], ['*', 'h5p', 'h5p', 'h5p'], 'h6p', ['*', 'j2p', 'j2p']], ['*', 3.0, ['*', 'h3p', 'h3p'], ['*', 'h5p', 'h5p', 'h5p'], 'h6p', 'j2p'], ['*', 3.0, ['*', 'h3p', 'h3p'], ['*', 'h5p', 'h5p'], ['*', 'h6p', 'h6p'], ['*', 'j2p', 'j2p', 'j2p', 'j2p', 'j2p']], ['*', 21.0, ['*', 'h3p', 'h3p'], ['*', 'h5p', 'h5p'], ['*', 'h6p', 'h6p'], ['*', 'j2p', 'j2p', 'j2p', 'j2p']], ['*', 39.0, ['*', 'h3p', 'h3p'], ['*', 'h5p', 'h5p'], ['*', 'h6p', 'h6p'], ['*', 'j2p', 'j2p', 'j2p']], ['*', 27.0, ['*', 'h3p', 'h3p'], ['*', 'h5p', 'h5p'], ['*', 'h6p', 'h6p'], ['*', 'j2p', 'j2p']], ['*', 6.0, ['*', 'h3p', 'h3p'], ['*', 'h5p', 'h5p'], ['*', 'h6p', 'h6p'], 'j2p'], ['*', 12.0, ['*', 'h3p', 'h3p'], 'h5p', ['*', 'h6p', 'h6p', 'h6p'], ['*', 'j2p', 'j2p', 'j2p', 'j2p']], ['*', 24.0, ['*', 'h3p', 'h3p'], 'h5p', ['*', 'h6p', 'h6p', 'h6p'], ['*', 'j2p', 'j2p', 'j2p']], ['*', 12.0, ['*', 'h3p', 'h3p'], 'h5p', ['*', 'h6p', 'h6p', 'h6p'], ['*', 'j2p', 'j2p']], ['*', 3.0, 'h3p', ['*', 'h5p', 'h5p', 'h5p'], ['*', 'h6p', 'h6p'], ['*', 'j2p', 'j2p', 'j2p', 'j2p', 'j2p']], ['*', 12.0, 'h3p', ['*', 'h5p', 'h5p', 'h5p'], ['*', 'h6p', 'h6p'], ['*', 'j2p', 'j2p', 'j2p', 'j2p']], ['*', 18.0, 'h3p', ['*', 'h5p', 'h5p', 'h5p'], ['*', 'h6p', 'h6p'], ['*', 'j2p', 'j2p', 'j2p']], ['*', 12.0, 'h3p', ['*', 'h5p', 'h5p', 'h5p'], ['*', 'h6p', 'h6p'], ['*', 'j2p', 'j2p']], ['*', 3.0, 'h3p', ['*', 'h5p', 'h5p', 'h5p'], ['*', 'h6p', 'h6p'], 'j2p'], ['*', 3.0, 'h3p', ['*', 'h5p', 'h5p'], ['*', 'h6p', 'h6p', 'h6p'], ['*', 'j2p', 'j2p', 'j2p', 'j2p', 'j2p']], ['*', 12.0, 'h3p', ['*', 'h5p', 'h5p'], ['*', 'h6p', 'h6p', 'h6p'], ['*', 'j2p', 'j2p', 'j2p', 'j2p']], ['*', 18.0, 'h3p', ['*', 'h5p', 'h5p'], ['*', 'h6p', 'h6p', 'h6p'], ['*', 'j2p', 'j2p', 'j2p']], ['*', 12.0, 'h3p', ['*', 'h5p', 'h5p'], ['*', 'h6p', 'h6p', 'h6p'], ['*', 'j2p', 'j2p']], ['*', 3.0, 'h3p', ['*', 'h5p', 'h5p'], ['*', 'h6p', 'h6p', 'h6p'], 'j2p']]],
['+', ['*', ['*', 'h1n', 'h1n', 'h1n'], ['*', 'h3n', 'h3n'], 'h5n', ['*', 'j2n', 'j2n']], ['*', ['*', 'h1n', 'h1n', 'h1n'], ['*', 'h3n', 'h3n'], 'h5n', 'j2n'], ['*', 4.0, ['*', 'h1n', 'h1n', 'h1n'], ['*', 'h3n', 'h3n'], 'h6n', ['*', 'j2n', 'j2n']], ['*', 4.0, ['*', 'h1n', 'h1n', 'h1n'], ['*', 'h3n', 'h3n'], 'h6n', 'j2n'], ['*', ['*', 'h1n', 'h1n', 'h1n'], 'h3n', ['*', 'h5n', 'h5n'], ['*', 'j2n', 'j2n']], ['*', ['*', 'h1n', 'h1n', 'h1n'], 'h3n', 'h5n', 'h6n', ['*', 'j2n', 'j2n', 'j2n']], ['*', 5.0, ['*', 'h1n', 'h1n', 'h1n'], 'h3n', 'h5n', 'h6n', ['*', 'j2n', 'j2n']], ['*', 2.0, ['*', 'h1n', 'h1n', 'h1n'], 'h3n', 'h5n', 'h6n', 'j2n'], ['*', 4.0, ['*', 'h1n', 'h1n', 'h1n'], 'h3n', ['*', 'h6n', 'h6n'], ['*', 'j2n', 'j2n']], ['*', 4.0, ['*', 'h1n', 'h1n', 'h1n'], 'h3n', ['*', 'h6n', 'h6n'], 'j2n'], ['*', ['*', 'h1n', 'h1n', 'h1n'], ['*', 'h5n', 'h5n'], 'h6n', ['*', 'j2n', 'j2n', 'j2n']], ['*', ['*', 'h1n', 'h1n', 'h1n'], ['*', 'h5n', 'h5n'], 'h6n', ['*', 'j2n', 'j2n']], ['*', ['*', 'h1n', 'h1n', 'h1n'], 'h5n', ['*', 'h6n', 'h6n'], ['*', 'j2n', 'j2n', 'j2n']], ['*', 2.0, ['*', 'h1n', 'h1n', 'h1n'], 'h5n', ['*', 'h6n', 'h6n'], ['*', 'j2n', 'j2n']], ['*', ['*', 'h1n', 'h1n', 'h1n'], 'h5n', ['*', 'h6n', 'h6n'], 'j2n'], ['*', 5.0, ['*', 'h1n', 'h1n'], 'h2n', ['*', 'h3n', 'h3n'], 'h5n', ['*', 'j2n', 'j2n']], ['*', 6.0, ['*', 'h1n', 'h1n'], 'h2n', ['*', 'h3n', 'h3n'], 'h5n', 'j2n'], ['*', ['*', 'h1n', 'h1n'], 'h2n', ['*', 'h3n', 'h3n'], 'h5n'], ['*', 12.0, ['*', 'h1n', 'h1n'], 'h2n', ['*', 'h3n', 'h3n'], 'h6n', ['*', 'j2n', 'j2n']], ['*', 12.0, ['*', 'h1n', 'h1n'], 'h2n', ['*', 'h3n', 'h3n'], 'h6n', 'j2n'], ['*', 5.0, ['*', 'h1n', 'h1n'], 'h2n', 'h3n', ['*', 'h5n', 'h5n'], ['*', 'j2n', 'j2n']], ['*', 2.0, ['*', 'h1n', 'h1n'], 'h2n', 'h3n', ['*', 'h5n', 'h5n'], 'j2n'], ['*', 3.0, ['*', 'h1n', 'h1n'], 'h2n', 'h3n', 'h5n', 'h6n', ['*', 'j2n', 'j2n', 'j2n']], ['*', 16.0, ['*', 'h1n', 'h1n'], 'h2n', 'h3n', 'h5n', 'h6n', ['*', 'j2n', 'j2n']], ['*', 11.0, ['*', 'h1n', 'h1n'], 'h2n', 'h3n', 'h5n', 'h6n', 'j2n'], ['*', 2.0, ['*', 'h1n', 'h1n'], 'h2n', 'h3n', 'h5n', 'h6n'], ['*', 12.0, ['*', 'h1n', 'h1n'], 'h2n', 'h3n', ['*', 'h6n', 'h6n'], ['*', 'j2n', 'j2n']], ['*', 12.0, ['*', 'h1n', 'h1n'], 'h2n', 'h3n', ['*', 'h6n', 'h6n'], 'j2n'], ['*', 3.0, ['*', 'h1n', 'h1n'], 'h2n', ['*', 'h5n', 'h5n'], 'h6n', ['*', 'j2n', 'j2n', 'j2n']], ['*', 5.0, ['*', 'h1n', 'h1n'], 'h2n', ['*', 'h5n', 'h5n'], 'h6n', ['*', 'j2n', 'j2n']], ['*', 2.0, ['*', 'h1n', 'h1n'], 'h2n', ['*', 'h5n', 'h5n'], 'h6n', 'j2n'], ['*', 3.0, ['*', 'h1n', 'h1n'], 'h2n', 'h5n', ['*', 'h6n', 'h6n'], ['*', 'j2n', 'j2n', 'j2n']], ['*', 7.0, ['*', 'h1n', 'h1n'], 'h2n', 'h5n', ['*', 'h6n', 'h6n'], ['*', 'j2n', 'j2n']], ['*', 5.0, ['*', 'h1n', 'h1n'], 'h2n', 'h5n', ['*', 'h6n', 'h6n'], 'j2n'], ['*', ['*', 'h1n', 'h1n'], 'h2n', 'h5n', ['*', 'h6n', 'h6n']], ['*', ['*', 'h1n', 'h1n'], ['*', 'h3n', 'h3n', 'h3n'], 'h5n', ['*', 'j2n', 'j2n', 'j2n']], ['*', 2.0, ['*', 'h1n', 'h1n'], ['*', 'h3n', 'h3n', 'h3n'], 'h5n', ['*', 'j2n', 'j2n']], ['*', ['*', 'h1n', 'h1n'], ['*', 'h3n', 'h3n', 'h3n'], 'h5n', 'j2n'], ['*', 4.0, ['*', 'h1n', 'h1n'], ['*', 'h3n', 'h3n', 'h3n'], 'h6n', ['*', 'j2n', 'j2n', 'j2n']], ['*', 8.0, ['*', 'h1n', 'h1n'], ['*', 'h3n', 'h3n', 'h3n'], 'h6n', ['*', 'j2n', 'j2n']], ['*', 4.0, ['*', 'h1n', 'h1n'], ['*', 'h3n', 'h3n', 'h3n'], 'h6n', 'j2n'], ['*', 2.0, ['*', 'h1n', 'h1n'], ['*', 'h3n', 'h3n'], ['*', 'h5n', 'h5n'], ['*', 'j2n', 'j2n', 'j2n']], ['*', 4.0, ['*', 'h1n', 'h1n'], ['*', 'h3n', 'h3n'], ['*', 'h5n', 'h5n'], ['*', 'j2n', 'j2n']], ['*', 2.0, ['*', 'h1n', 'h1n'], ['*', 'h3n', 'h3n'], ['*', 'h5n', 'h5n'], 'j2n'], ['*', ['*', 'h1n', 'h1n'], ['*', 'h3n', 'h3n'], 'h5n', 'h6n', ['*', 'j2n', 'j2n', 'j2n', 'j2n']], ['*', 12.0, ['*', 'h1n', 'h1n'], ['*', 'h3n', 'h3n'], 'h5n', 'h6n', ['*', 'j2n', 'j2n', 'j2n']], ['*', 25.0, ['*', 'h1n', 'h1n'], ['*', 'h3n', 'h3n'], 'h5n', 'h6n', ['*', 'j2n', 'j2n']], ['*', 14.0, ['*', 'h1n', 'h1n'], ['*', 'h3n', 'h3n'], 'h5n', 'h6n', 'j2n'], ['*', 8.0, ['*', 'h1n', 'h1n'], ['*', 'h3n', 'h3n'], ['*', 'h6n', 'h6n'], ['*', 'j2n', 'j2n', 'j2n']], ['*', 16.0, ['*', 'h1n', 'h1n'], ['*', 'h3n', 'h3n'], ['*', 'h6n', 'h6n'], ['*', 'j2n', 'j2n']], ['*', 8.0, ['*', 'h1n', 'h1n'], ['*', 'h3n', 'h3n'], ['*', 'h6n', 'h6n'], 'j2n'], ['*', ['*', 'h1n', 'h1n'], 'h3n', ['*', 'h5n', 'h5n', 'h5n'], ['*', 'j2n', 'j2n', 'j2n']], ['*', ['*', 'h1n', 'h1n'], 'h3n', ['*', 'h5n', 'h5n', 'h5n'], ['*', 'j2n', 'j2n']], ['*', 2.0, ['*', 'h1n', 'h1n'], 'h3n', ['*', 'h5n', 'h5n'], 'h6n', ['*', 'j2n', 'j2n', 'j2n', 'j2n']], ['*', 11.0, ['*', 'h1n', 'h1n'], 'h3n', ['*', 'h5n', 'h5n'], 'h6n', ['*', 'j2n', 'j2n', 'j2n']], ['*', 13.0, ['*', 'h1n', 'h1n'], 'h3n', ['*', 'h5n', 'h5n'], 'h6n', ['*', 'j2n', 'j2n']], ['*', 4.0, ['*', 'h1n', 'h1n'], 'h3n', ['*', 'h5n', 'h5n'], 'h6n', 'j2n'], ['*', 2.0, ['*', 'h1n', 'h1n'], 'h3n', 'h5n', ['*', 'h6n', 'h6n'], ['*', 'j2n', 'j2n', 'j2n', 'j2n']], ['*', 14.0, ['*', 'h1n', 'h1n'], 'h3n', 'h5n', ['*', 'h6n', 'h6n'], ['*', 'j2n', 'j2n', 'j2n']], ['*', 26.0, ['*', 'h1n', 'h1n'], 'h3n', 'h5n', ['*', 'h6n', 'h6n'], ['*', 'j2n', 'j2n']], ['*', 14.0, ['*', 'h1n', 'h1n'], 'h3n', 'h5n', ['*', 'h6n', 'h6n'], 'j2n'], ['*', 4.0, ['*', 'h1n', 'h1n'], 'h3n', ['*', 'h6n', 'h6n', 'h6n'], ['*', 'j2n', 'j2n', 'j2n']], ['*', 8.0, ['*', 'h1n', 'h1n'], 'h3n', ['*', 'h6n', 'h6n', 'h6n'], ['*', 'j2n', 'j2n']], ['*', 4.0, ['*', 'h1n', 'h1n'], 'h3n', ['*', 'h6n', 'h6n', 'h6n'], 'j2n'], ['*', ['*', 'h1n', 'h1n'], ['*', 'h5n', 'h5n', 'h5n'], 'h6n', ['*', 'j2n', 'j2n', 'j2n', 'j2n']], ['*', 2.0, ['*', 'h1n', 'h1n'], ['*', 'h5n', 'h5n', 'h5n'], 'h6n', ['*', 'j2n', 'j2n', 'j2n']], ['*', ['*', 'h1n', 'h1n'], ['*', 'h5n', 'h5n', 'h5n'], 'h6n', ['*', 'j2n', 'j2n']], ['*', 2.0, ['*', 'h1n', 'h1n'], ['*', 'h5n', 'h5n'], ['*', 'h6n', 'h6n'], ['*', 'j2n', 'j2n', 'j2n', 'j2n']], ['*', 6.0, ['*', 'h1n', 'h1n'], ['*', 'h5n', 'h5n'], ['*', 'h6n', 'h6n'], ['*', 'j2n', 'j2n', 'j2n']], ['*', 6.0, ['*', 'h1n', 'h1n'], ['*', 'h5n', 'h5n'], ['*', 'h6n', 'h6n'], ['*', 'j2n', 'j2n']], ['*', 2.0, ['*', 'h1n', 'h1n'], ['*', 'h5n', 'h5n'], ['*', 'h6n', 'h6n'], 'j2n'], ['*', ['*', 'h1n', 'h1n'], 'h5n', ['*', 'h6n', 'h6n', 'h6n'], ['*', 'j2n', 'j2n', 'j2n', 'j2n']], ['*', 3.0, ['*', 'h1n', 'h1n'], 'h5n', ['*', 'h6n', 'h6n', 'h6n'], ['*', 'j2n', 'j2n', 'j2n']], ['*', 3.0, ['*', 'h1n', 'h1n'], 'h5n', ['*', 'h6n', 'h6n', 'h6n'], ['*', 'j2n', 'j2n']], ['*', ['*', 'h1n', 'h1n'], 'h5n', ['*', 'h6n', 'h6n', 'h6n'], 'j2n'], ['*', 7.0, 'h1n', ['*', 'h2n', 'h2n'], ['*', 'h3n', 'h3n'], 'h5n', ['*', 'j2n', 'j2n']], ['*', 9.0, 'h1n', ['*', 'h2n', 'h2n'], ['*', 'h3n', 'h3n'], 'h5n', 'j2n'], ['*', 2.0, 'h1n', ['*', 'h2n', 'h2n'], ['*', 'h3n', 'h3n'], 'h5n'], ['*', 12.0, 'h1n', ['*', 'h2n', 'h2n'], ['*', 'h3n', 'h3n'], 'h6n', ['*', 'j2n', 'j2n']], ['*', 12.0, 'h1n', ['*', 'h2n', 'h2n'], ['*', 'h3n', 'h3n'], 'h6n', 'j2n'], ['*', 7.0, 'h1n', ['*', 'h2n', 'h2n'], 'h3n', ['*', 'h5n', 'h5n'], ['*', 'j2n', 'j2n']], ['*', 6.0, 'h1n', ['*', 'h2n', 'h2n'], 'h3n', ['*', 'h5n', 'h5n'], 'j2n'], ['*', 'h1n', ['*', 'h2n', 'h2n'], 'h3n', ['*', 'h5n', 'h5n']], ['*', 3.0, 'h1n', ['*', 'h2n', 'h2n'], 'h3n', 'h5n', 'h6n', ['*', 'j2n', 'j2n', 'j2n']], ['*', 17.0, 'h1n', ['*', 'h2n', 'h2n'], 'h3n', 'h5n', 'h6n', ['*', 'j2n', 'j2n']], ['*', 16.0, 'h1n', ['*', 'h2n', 'h2n'], 'h3n', 'h5n', 'h6n', 'j2n'], ['*', 4.0, 'h1n', ['*', 'h2n', 'h2n'], 'h3n', 'h5n', 'h6n'], ['*', 12.0, 'h1n', ['*', 'h2n', 'h2n'], 'h3n', ['*', 'h6n', 'h6n'], ['*', 'j2n', 'j2n']], ['*', 12.0, 'h1n', ['*', 'h2n', 'h2n'], 'h3n', ['*', 'h6n', 'h6n'], 'j2n'], ['*', 3.0, 'h1n', ['*', 'h2n', 'h2n'], ['*', 'h5n', 'h5n'], 'h6n', ['*', 'j2n', 'j2n', 'j2n']], ['*', 7.0, 'h1n', ['*', 'h2n', 'h2n'], ['*', 'h5n', 'h5n'], 'h6n', ['*', 'j2n', 'j2n']], ['*', 5.0, 'h1n', ['*', 'h2n', 'h2n'], ['*', 'h5n', 'h5n'], 'h6n', 'j2n'], ['*', 'h1n', ['*', 'h2n', 'h2n'], ['*', 'h5n', 'h5n'], 'h6n'], ['*', 3.0, 'h1n', ['*', 'h2n', 'h2n'], 'h5n', ['*', 'h6n', 'h6n'], ['*', 'j2n', 'j2n', 'j2n']], ['*', 8.0, 'h1n', ['*', 'h2n', 'h2n'], 'h5n', ['*', 'h6n', 'h6n'], ['*', 'j2n', 'j2n']], ['*', 7.0, 'h1n', ['*', 'h2n', 'h2n'], 'h5n', ['*', 'h6n', 'h6n'], 'j2n'], ['*', 2.0, 'h1n', ['*', 'h2n', 'h2n'], 'h5n', ['*', 'h6n', 'h6n']], ['*', 4.0, 'h1n', 'h2n', ['*', 'h3n', 'h3n', 'h3n'], 'h5n', ['*', 'j2n', 'j2n', 'j2n']], ['*', 9.0, 'h1n', 'h2n', ['*', 'h3n', 'h3n', 'h3n'], 'h5n', ['*', 'j2n', 'j2n']], ['*', 6.0, 'h1n', 'h2n', ['*', 'h3n', 'h3n', 'h3n'], 'h5n', 'j2n'], ['*', 'h1n', 'h2n', ['*', 'h3n', 'h3n', 'h3n'], 'h5n'], ['*', 8.0, 'h1n', 'h2n', ['*', 'h3n', 'h3n', 'h3n'], 'h6n', ['*', 'j2n', 'j2n', 'j2n']], ['*', 16.0, 'h1n', 'h2n', ['*', 'h3n', 'h3n', 'h3n'], 'h6n', ['*', 'j2n', 'j2n']], ['*', 8.0, 'h1n', 'h2n', ['*', 'h3n', 'h3n', 'h3n'], 'h6n', 'j2n'], ['*', 8.0, 'h1n', 'h2n', ['*', 'h3n', 'h3n'], ['*', 'h5n', 'h5n'], ['*', 'j2n', 'j2n', 'j2n']], ['*', 17.0, 'h1n', 'h2n', ['*', 'h3n', 'h3n'], ['*', 'h5n', 'h5n'], ['*', 'j2n', 'j2n']], ['*', 12.0, 'h1n', 'h2n', ['*', 'h3n', 'h3n'], ['*', 'h5n', 'h5n'], 'j2n'], ['*', 2.0, 'h1n', 'h2n', ['*', 'h3n', 'h3n'], ['*', 'h5n', 'h5n']], ['*', 2.0, 'h1n', 'h2n', ['*', 'h3n', 'h3n'], 'h5n', 'h6n', ['*', 'j2n', 'j2n', 'j2n', 'j2n']], ['*', 25.0, 'h1n', 'h2n', ['*', 'h3n', 'h3n'], 'h5n', 'h6n', ['*', 'j2n', 'j2n', 'j2n']], ['*', 47.0, 'h1n', 'h2n', ['*', 'h3n', 'h3n'], 'h5n', 'h6n', ['*', 'j2n', 'j2n']], ['*', 31.0, 'h1n', 'h2n', ['*', 'h3n', 'h3n'], 'h5n', 'h6n', 'j2n'], ['*', 3.0, 'h1n', 'h2n', ['*', 'h3n', 'h3n'], 'h5n', 'h6n'], ['*', 16.0, 'h1n', 'h2n', ['*', 'h3n', 'h3n'], ['*', 'h6n', 'h6n'], ['*', 'j2n', 'j2n', 'j2n']], ['*', 32.0, 'h1n', 'h2n', ['*', 'h3n', 'h3n'], ['*', 'h6n', 'h6n'], ['*', 'j2n', 'j2n']], ['*', 16.0, 'h1n', 'h2n', ['*', 'h3n', 'h3n'], ['*', 'h6n', 'h6n'], 'j2n'], ['*', 4.0, 'h1n', 'h2n', 'h3n', ['*', 'h5n', 'h5n', 'h5n'], ['*', 'j2n', 'j2n', 'j2n']], ['*', 6.0, 'h1n', 'h2n', 'h3n', ['*', 'h5n', 'h5n', 'h5n'], ['*', 'j2n', 'j2n']], ['*', 2.0, 'h1n', 'h2n', 'h3n', ['*', 'h5n', 'h5n', 'h5n'], 'j2n'], ['*', 4.0, 'h1n', 'h2n', 'h3n', ['*', 'h5n', 'h5n'], 'h6n', ['*', 'j2n', 'j2n', 'j2n', 'j2n']], ['*', 23.0, 'h1n', 'h2n', 'h3n', ['*', 'h5n', 'h5n'], 'h6n', ['*', 'j2n', 'j2n', 'j2n']], ['*', 36.0, 'h1n', 'h2n', 'h3n', ['*', 'h5n', 'h5n'], 'h6n', ['*', 'j2n', 'j2n']], ['*', 21.0, 'h1n', 'h2n', 'h3n', ['*', 'h5n', 'h5n'], 'h6n', 'j2n'], ['*', 4.0, 'h1n', 'h2n', 'h3n', ['*', 'h5n', 'h5n'], 'h6n'], ['*', 4.0, 'h1n', 'h2n', 'h3n', 'h5n', ['*', 'h6n', 'h6n'], ['*', 'j2n', 'j2n', 'j2n', 'j2n']], ['*', 28.0, 'h1n', 'h2n', 'h3n', 'h5n', ['*', 'h6n', 'h6n'], ['*', 'j2n', 'j2n', 'j2n']], ['*', 51.0, 'h1n', 'h2n', 'h3n', 'h5n', ['*', 'h6n', 'h6n'], ['*', 'j2n', 'j2n']], ['*', 30.0, 'h1n', 'h2n', 'h3n', 'h5n', ['*', 'h6n', 'h6n'], 'j2n'], ['*', 3.0, 'h1n', 'h2n', 'h3n', 'h5n', ['*', 'h6n', 'h6n']], ['*', 8.0, 'h1n', 'h2n', 'h3n', ['*', 'h6n', 'h6n', 'h6n'], ['*', 'j2n', 'j2n', 'j2n']], ['*', 16.0, 'h1n', 'h2n', 'h3n', ['*', 'h6n', 'h6n', 'h6n'], ['*', 'j2n', 'j2n']], ['*', 8.0, 'h1n', 'h2n', 'h3n', ['*', 'h6n', 'h6n', 'h6n'], 'j2n'], ['*', 2.0, 'h1n', 'h2n', ['*', 'h5n', 'h5n', 'h5n'], 'h6n', ['*', 'j2n', 'j2n', 'j2n', 'j2n']], ['*', 6.0, 'h1n', 'h2n', ['*', 'h5n', 'h5n', 'h5n'], 'h6n', ['*', 'j2n', 'j2n', 'j2n']], ['*', 6.0, 'h1n', 'h2n', ['*', 'h5n', 'h5n', 'h5n'], 'h6n', ['*', 'j2n', 'j2n']], ['*', 2.0, 'h1n', 'h2n', ['*', 'h5n', 'h5n', 'h5n'], 'h6n', 'j2n'], ['*', 4.0, 'h1n', 'h2n', ['*', 'h5n', 'h5n'], ['*', 'h6n', 'h6n'], ['*', 'j2n', 'j2n', 'j2n', 'j2n']], ['*', 14.0, 'h1n', 'h2n', ['*', 'h5n', 'h5n'], ['*', 'h6n', 'h6n'], ['*', 'j2n', 'j2n', 'j2n']], ['*', 18.0, 'h1n', 'h2n', ['*', 'h5n', 'h5n'], ['*', 'h6n', 'h6n'], ['*', 'j2n', 'j2n']], ['*', 10.0, 'h1n', 'h2n', ['*', 'h5n', 'h5n'], ['*', 'h6n', 'h6n'], 'j2n'], ['*', 2.0, 'h1n', 'h2n', ['*', 'h5n', 'h5n'], ['*', 'h6n', 'h6n']], ['*', 2.0, 'h1n', 'h2n', 'h5n', ['*', 'h6n', 'h6n', 'h6n'], ['*', 'j2n', 'j2n', 'j2n', 'j2n']], ['*', 7.0, 'h1n', 'h2n', 'h5n', ['*', 'h6n', 'h6n', 'h6n'], ['*', 'j2n', 'j2n', 'j2n']], ['*', 9.0, 'h1n', 'h2n', 'h5n', ['*', 'h6n', 'h6n', 'h6n'], ['*', 'j2n', 'j2n']], ['*', 5.0, 'h1n', 'h2n', 'h5n', ['*', 'h6n', 'h6n', 'h6n'], 'j2n'], ['*', 'h1n', 'h2n', 'h5n', ['*', 'h6n', 'h6n', 'h6n']], ['*', 2.0, 'h1n', ['*', 'h3n', 'h3n', 'h3n'], ['*', 'h5n', 'h5n'], ['*', 'j2n', 'j2n', 'j2n']], ['*', 3.0, 'h1n', ['*', 'h3n', 'h3n', 'h3n'], ['*', 'h5n', 'h5n'], ['*', 'j2n', 'j2n']], ['*', 'h1n', ['*', 'h3n', 'h3n', 'h3n'], ['*', 'h5n', 'h5n'], 'j2n'], ['*', 'h1n', ['*', 'h3n', 'h3n', 'h3n'], 'h5n', 'h6n', ['*', 'j2n', 'j2n', 'j2n', 'j2n']], ['*', 17.0, 'h1n', ['*', 'h3n', 'h3n', 'h3n'], 'h5n', 'h6n', ['*', 'j2n', 'j2n', 'j2n']], ['*', 23.0, 'h1n', ['*', 'h3n', 'h3n', 'h3n'], 'h5n', 'h6n', ['*', 'j2n', 'j2n']], ['*', 7.0, 'h1n', ['*', 'h3n', 'h3n', 'h3n'], 'h5n', 'h6n', 'j2n'], ['*', 16.0, 'h1n', ['*', 'h3n', 'h3n', 'h3n'], ['*', 'h6n', 'h6n'], ['*', 'j2n', 'j2n', 'j2n']], ['*', 16.0, 'h1n', ['*', 'h3n', 'h3n', 'h3n'], ['*', 'h6n', 'h6n'], ['*', 'j2n', 'j2n']], ['*', 2.0, 'h1n', ['*', 'h3n', 'h3n'], ['*', 'h5n', 'h5n', 'h5n'], ['*', 'j2n', 'j2n', 'j2n']], ['*', 3.0, 'h1n', ['*', 'h3n', 'h3n'], ['*', 'h5n', 'h5n', 'h5n'], ['*', 'j2n', 'j2n']], ['*', 'h1n', ['*', 'h3n', 'h3n'], ['*', 'h5n', 'h5n', 'h5n'], 'j2n'], ['*', 5.0, 'h1n', ['*', 'h3n', 'h3n'], ['*', 'h5n', 'h5n'], 'h6n', ['*', 'j2n', 'j2n', 'j2n', 'j2n']], ['*', 29.0, 'h1n', ['*', 'h3n', 'h3n'], ['*', 'h5n', 'h5n'], 'h6n', ['*', 'j2n', 'j2n', 'j2n']], ['*', 37.0, 'h1n', ['*', 'h3n', 'h3n'], ['*', 'h5n', 'h5n'], 'h6n', ['*', 'j2n', 'j2n']], ['*', 13.0, 'h1n', ['*', 'h3n', 'h3n'], ['*', 'h5n', 'h5n'], 'h6n', 'j2n'], ['*', 10.0, 'h1n', ['*', 'h3n', 'h3n'], 'h5n', ['*', 'h6n', 'h6n'], ['*', 'j2n', 'j2n', 'j2n', 'j2n']], ['*', 46.0, 'h1n', ['*', 'h3n', 'h3n'], 'h5n', ['*', 'h6n', 'h6n'], ['*', 'j2n', 'j2n', 'j2n']], ['*', 50.0, 'h1n', ['*', 'h3n', 'h3n'], 'h5n', ['*', 'h6n', 'h6n'], ['*', 'j2n', 'j2n']], ['*', 14.0, 'h1n', ['*', 'h3n', 'h3n'], 'h5n', ['*', 'h6n', 'h6n'], 'j2n'], ['*', 16.0, 'h1n', ['*', 'h3n', 'h3n'], ['*', 'h6n', 'h6n', 'h6n'], ['*', 'j2n', 'j2n', 'j2n']], ['*', 16.0, 'h1n', ['*', 'h3n', 'h3n'], ['*', 'h6n', 'h6n', 'h6n'], ['*', 'j2n', 'j2n']], ['*', 4.0, 'h1n', 'h3n', ['*', 'h5n', 'h5n', 'h5n'], 'h6n', ['*', 'j2n', 'j2n', 'j2n', 'j2n']], ['*', 10.0, 'h1n', 'h3n', ['*', 'h5n', 'h5n', 'h5n'], 'h6n', ['*', 'j2n', 'j2n', 'j2n']], ['*', 8.0, 'h1n', 'h3n', ['*', 'h5n', 'h5n', 'h5n'], 'h6n', ['*', 'j2n', 'j2n']], ['*', 2.0, 'h1n', 'h3n', ['*', 'h5n', 'h5n', 'h5n'], 'h6n', 'j2n'], ['*', 'h1n', 'h3n', ['*', 'h5n', 'h5n'], ['*', 'h6n', 'h6n'], ['*', 'j2n', 'j2n', 'j2n', 'j2n', 'j2n']], ['*', 14.0, 'h1n', 'h3n', ['*', 'h5n', 'h5n'], ['*', 'h6n', 'h6n'], ['*', 'j2n', 'j2n', 'j2n', 'j2n']], ['*', 38.0, 'h1n', 'h3n', ['*', 'h5n', 'h5n'], ['*', 'h6n', 'h6n'], ['*', 'j2n', 'j2n', 'j2n']], ['*', 38.0, 'h1n', 'h3n', ['*', 'h5n', 'h5n'], ['*', 'h6n', 'h6n'], ['*', 'j2n', 'j2n']], ['*', 13.0, 'h1n', 'h3n', ['*', 'h5n', 'h5n'], ['*', 'h6n', 'h6n'], 'j2n'], ['*', 9.0, 'h1n', 'h3n', 'h5n', ['*', 'h6n', 'h6n', 'h6n'], ['*', 'j2n', 'j2n', 'j2n', 'j2n']], ['*', 25.0, 'h1n', 'h3n', 'h5n', ['*', 'h6n', 'h6n', 'h6n'], ['*', 'j2n', 'j2n', 'j2n']], ['*', 23.0, 'h1n', 'h3n', 'h5n', ['*', 'h6n', 'h6n', 'h6n'], ['*', 'j2n', 'j2n']], ['*', 7.0, 'h1n', 'h3n', 'h5n', ['*', 'h6n', 'h6n', 'h6n'], 'j2n'], ['*', 'h1n', ['*', 'h5n', 'h5n', 'h5n'], ['*', 'h6n', 'h6n'], ['*', 'j2n', 'j2n', 'j2n', 'j2n', 'j2n']], ['*', 4.0, 'h1n', ['*', 'h5n', 'h5n', 'h5n'], ['*', 'h6n', 'h6n'], ['*', 'j2n', 'j2n', 'j2n', 'j2n']], ['*', 6.0, 'h1n', ['*', 'h5n', 'h5n', 'h5n'], ['*', 'h6n', 'h6n'], ['*', 'j2n', 'j2n', 'j2n']], ['*', 4.0, 'h1n', ['*', 'h5n', 'h5n', 'h5n'], ['*', 'h6n', 'h6n'], ['*', 'j2n', 'j2n']], ['*', 'h1n', ['*', 'h5n', 'h5n', 'h5n'], ['*', 'h6n', 'h6n'], 'j2n'], ['*', 'h1n', ['*', 'h5n', 'h5n'], ['*', 'h6n', 'h6n', 'h6n'], ['*', 'j2n', 'j2n', 'j2n', 'j2n', 'j2n']], ['*', 4.0, 'h1n', ['*', 'h5n', 'h5n'], ['*', 'h6n', 'h6n', 'h6n'], ['*', 'j2n', 'j2n', 'j2n', 'j2n']], ['*', 6.0, 'h1n', ['*', 'h5n', 'h5n'], ['*', 'h6n', 'h6n', 'h6n'], ['*', 'j2n', 'j2n', 'j2n']], ['*', 4.0, 'h1n', ['*', 'h5n', 'h5n'], ['*', 'h6n', 'h6n', 'h6n'], ['*', 'j2n', 'j2n']], ['*', 'h1n', ['*', 'h5n', 'h5n'], ['*', 'h6n', 'h6n', 'h6n'], 'j2n'], ['*', 3.0, ['*', 'h2n', 'h2n', 'h2n'], ['*', 'h3n', 'h3n'], 'h5n', ['*', 'j2n', 'j2n']], ['*', 4.0, ['*', 'h2n', 'h2n', 'h2n'], ['*', 'h3n', 'h3n'], 'h5n', 'j2n'], ['*', ['*', 'h2n', 'h2n', 'h2n'], ['*', 'h3n', 'h3n'], 'h5n'], ['*', 4.0, ['*', 'h2n', 'h2n', 'h2n'], ['*', 'h3n', 'h3n'], 'h6n', ['*', 'j2n', 'j2n']], ['*', 4.0, ['*', 'h2n', 'h2n', 'h2n'], ['*', 'h3n', 'h3n'], 'h6n', 'j2n'], ['*', 3.0, ['*', 'h2n', 'h2n', 'h2n'], 'h3n', ['*', 'h5n', 'h5n'], ['*', 'j2n', 'j2n']], ['*', 4.0, ['*', 'h2n', 'h2n', 'h2n'], 'h3n', ['*', 'h5n', 'h5n'], 'j2n'], ['*', ['*', 'h2n', 'h2n', 'h2n'], 'h3n', ['*', 'h5n', 'h5n']], ['*', ['*', 'h2n', 'h2n', 'h2n'], 'h3n', 'h5n', 'h6n', ['*', 'j2n', 'j2n', 'j2n']], ['*', 6.0, ['*', 'h2n', 'h2n', 'h2n'], 'h3n', 'h5n', 'h6n', ['*', 'j2n', 'j2n']], ['*', 7.0, ['*', 'h2n', 'h2n', 'h2n'], 'h3n', 'h5n', 'h6n', 'j2n'], ['*', 2.0, ['*', 'h2n', 'h2n', 'h2n'], 'h3n', 'h5n', 'h6n'], ['*', 4.0, ['*', 'h2n', 'h2n', 'h2n'], 'h3n', ['*', 'h6n', 'h6n'], ['*', 'j2n', 'j2n']], ['*', 4.0, ['*', 'h2n', 'h2n', 'h2n'], 'h3n', ['*', 'h6n', 'h6n'], 'j2n'], ['*', ['*', 'h2n', 'h2n', 'h2n'], ['*', 'h5n', 'h5n'], 'h6n', ['*', 'j2n', 'j2n', 'j2n']], ['*', 3.0, ['*', 'h2n', 'h2n', 'h2n'], ['*', 'h5n', 'h5n'], 'h6n', ['*', 'j2n', 'j2n']], ['*', 3.0, ['*', 'h2n', 'h2n', 'h2n'], ['*', 'h5n', 'h5n'], 'h6n', 'j2n'], ['*', ['*', 'h2n', 'h2n', 'h2n'], ['*', 'h5n', 'h5n'], 'h6n'], ['*', ['*', 'h2n', 'h2n', 'h2n'], 'h5n', ['*', 'h6n', 'h6n'], ['*', 'j2n', 'j2n', 'j2n']], ['*', 3.0, ['*', 'h2n', 'h2n', 'h2n'], 'h5n', ['*', 'h6n', 'h6n'], ['*', 'j2n', 'j2n']], ['*', 3.0, ['*', 'h2n', 'h2n', 'h2n'], 'h5n', ['*', 'h6n', 'h6n'], 'j2n'], ['*', ['*', 'h2n', 'h2n', 'h2n'], 'h5n', ['*', 'h6n', 'h6n']], ['*', 3.0, ['*', 'h2n', 'h2n'], ['*', 'h3n', 'h3n', 'h3n'], 'h5n', ['*', 'j2n', 'j2n', 'j2n']], ['*', 7.0, ['*', 'h2n', 'h2n'], ['*', 'h3n', 'h3n', 'h3n'], 'h5n', ['*', 'j2n', 'j2n']], ['*', 5.0, ['*', 'h2n', 'h2n'], ['*', 'h3n', 'h3n', 'h3n'], 'h5n', 'j2n'], ['*', ['*', 'h2n', 'h2n'], ['*', 'h3n', 'h3n', 'h3n'], 'h5n'], ['*', 4.0, ['*', 'h2n', 'h2n'], ['*', 'h3n', 'h3n', 'h3n'], 'h6n', ['*', 'j2n', 'j2n', 'j2n']], ['*', 8.0, ['*', 'h2n', 'h2n'], ['*', 'h3n', 'h3n', 'h3n'], 'h6n', ['*', 'j2n', 'j2n']], ['*', 4.0, ['*', 'h2n', 'h2n'], ['*', 'h3n', 'h3n', 'h3n'], 'h6n', 'j2n'], ['*', 6.0, ['*', 'h2n', 'h2n'], ['*', 'h3n', 'h3n'], ['*', 'h5n', 'h5n'], ['*', 'j2n', 'j2n', 'j2n']], ['*', 11.0, ['*', 'h2n', 'h2n'], ['*', 'h3n', 'h3n'], ['*', 'h5n', 'h5n'], ['*', 'j2n', 'j2n']], ['*', 9.0, ['*', 'h2n', 'h2n'], ['*', 'h3n', 'h3n'], ['*', 'h5n', 'h5n'], 'j2n'], ['*', 2.0, ['*', 'h2n', 'h2n'], ['*', 'h3n', 'h3n'], ['*', 'h5n', 'h5n']], ['*', ['*', 'h2n', 'h2n'], ['*', 'h3n', 'h3n'], 'h5n', 'h6n', ['*', 'j2n', 'j2n', 'j2n', 'j2n']], ['*', 13.0, ['*', 'h2n', 'h2n'], ['*', 'h3n', 'h3n'], 'h5n', 'h6n', ['*', 'j2n', 'j2n', 'j2n']], ['*', 22.0, ['*', 'h2n', 'h2n'], ['*', 'h3n', 'h3n'], 'h5n', 'h6n', ['*', 'j2n', 'j2n']], ['*', 17.0, ['*', 'h2n', 'h2n'], ['*', 'h3n', 'h3n'], 'h5n', 'h6n', 'j2n'], ['*', 3.0, ['*', 'h2n', 'h2n'], ['*', 'h3n', 'h3n'], 'h5n', 'h6n'], ['*', 8.0, ['*', 'h2n', 'h2n'], ['*', 'h3n', 'h3n'], ['*', 'h6n', 'h6n'], ['*', 'j2n', 'j2n', 'j2n']], ['*', 16.0, ['*', 'h2n', 'h2n'], ['*', 'h3n', 'h3n'], ['*', 'h6n', 'h6n'], ['*', 'j2n', 'j2n']], ['*', 8.0, ['*', 'h2n', 'h2n'], ['*', 'h3n', 'h3n'], ['*', 'h6n', 'h6n'], 'j2n'], ['*', 3.0, ['*', 'h2n', 'h2n'], 'h3n', ['*', 'h5n', 'h5n', 'h5n'], ['*', 'j2n', 'j2n', 'j2n']], ['*', 7.0, ['*', 'h2n', 'h2n'], 'h3n', ['*', 'h5n', 'h5n', 'h5n'], ['*', 'j2n', 'j2n']], ['*', 5.0, ['*', 'h2n', 'h2n'], 'h3n', ['*', 'h5n', 'h5n', 'h5n'], 'j2n'], ['*', ['*', 'h2n', 'h2n'], 'h3n', ['*', 'h5n', 'h5n', 'h5n']], ['*', 2.0, ['*', 'h2n', 'h2n'], 'h3n', ['*', 'h5n', 'h5n'], 'h6n', ['*', 'j2n', 'j2n', 'j2n', 'j2n']], ['*', 12.0, ['*', 'h2n', 'h2n'], 'h3n', ['*', 'h5n', 'h5n'], 'h6n', ['*', 'j2n', 'j2n', 'j2n']], ['*', 22.0, ['*', 'h2n', 'h2n'], 'h3n', ['*', 'h5n', 'h5n'], 'h6n', ['*', 'j2n', 'j2n']], ['*', 16.0, ['*', 'h2n', 'h2n'], 'h3n', ['*', 'h5n', 'h5n'], 'h6n', 'j2n'], ['*', 4.0, ['*', 'h2n', 'h2n'], 'h3n', ['*', 'h5n', 'h5n'], 'h6n'], ['*', 2.0, ['*', 'h2n', 'h2n'], 'h3n', 'h5n', ['*', 'h6n', 'h6n'], ['*', 'j2n', 'j2n', 'j2n', 'j2n']], ['*', 14.0, ['*', 'h2n', 'h2n'], 'h3n', 'h5n', ['*', 'h6n', 'h6n'], ['*', 'j2n', 'j2n', 'j2n']], ['*', 25.0, ['*', 'h2n', 'h2n'], 'h3n', 'h5n', ['*', 'h6n', 'h6n'], ['*', 'j2n', 'j2n']], ['*', 16.0, ['*', 'h2n', 'h2n'], 'h3n', 'h5n', ['*', 'h6n', 'h6n'], 'j2n'], ['*', 3.0, ['*', 'h2n', 'h2n'], 'h3n', 'h5n', ['*', 'h6n', 'h6n']], ['*', 4.0, ['*', 'h2n', 'h2n'], 'h3n', ['*', 'h6n', 'h6n', 'h6n'], ['*', 'j2n', 'j2n', 'j2n']], ['*', 8.0, ['*', 'h2n', 'h2n'], 'h3n', ['*', 'h6n', 'h6n', 'h6n'], ['*', 'j2n', 'j2n']], ['*', 4.0, ['*', 'h2n', 'h2n'], 'h3n', ['*', 'h6n', 'h6n', 'h6n'], 'j2n'], ['*', ['*', 'h2n', 'h2n'], ['*', 'h5n', 'h5n', 'h5n'], 'h6n', ['*', 'j2n', 'j2n', 'j2n', 'j2n']], ['*', 4.0, ['*', 'h2n', 'h2n'], ['*', 'h5n', 'h5n', 'h5n'], 'h6n', ['*', 'j2n', 'j2n', 'j2n']], ['*', 6.0, ['*', 'h2n', 'h2n'], ['*', 'h5n', 'h5n', 'h5n'], 'h6n', ['*', 'j2n', 'j2n']], ['*', 4.0, ['*', 'h2n', 'h2n'], ['*', 'h5n', 'h5n', 'h5n'], 'h6n', 'j2n'], ['*', ['*', 'h2n', 'h2n'], ['*', 'h5n', 'h5n', 'h5n'], 'h6n'], ['*', 2.0, ['*', 'h2n', 'h2n'], ['*', 'h5n', 'h5n'], ['*', 'h6n', 'h6n'], ['*', 'j2n', 'j2n', 'j2n', 'j2n']], ['*', 8.0, ['*', 'h2n', 'h2n'], ['*', 'h5n', 'h5n'], ['*', 'h6n', 'h6n'], ['*', 'j2n', 'j2n', 'j2n']], ['*', 12.0, ['*', 'h2n', 'h2n'], ['*', 'h5n', 'h5n'], ['*', 'h6n', 'h6n'], ['*', 'j2n', 'j2n']], ['*', 8.0, ['*', 'h2n', 'h2n'], ['*', 'h5n', 'h5n'], ['*', 'h6n', 'h6n'], 'j2n'], ['*', 2.0, ['*', 'h2n', 'h2n'], ['*', 'h5n', 'h5n'], ['*', 'h6n', 'h6n']], ['*', ['*', 'h2n', 'h2n'], 'h5n', ['*', 'h6n', 'h6n', 'h6n'], ['*', 'j2n', 'j2n', 'j2n', 'j2n']], ['*', 4.0, ['*', 'h2n', 'h2n'], 'h5n', ['*', 'h6n', 'h6n', 'h6n'], ['*', 'j2n', 'j2n', 'j2n']], ['*', 6.0, ['*', 'h2n', 'h2n'], 'h5n', ['*', 'h6n', 'h6n', 'h6n'], ['*', 'j2n', 'j2n']], ['*', 4.0, ['*', 'h2n', 'h2n'], 'h5n', ['*', 'h6n', 'h6n', 'h6n'], 'j2n'], ['*', ['*', 'h2n', 'h2n'], 'h5n', ['*', 'h6n', 'h6n', 'h6n']], ['*', 6.0, 'h2n', ['*', 'h3n', 'h3n', 'h3n'], ['*', 'h5n', 'h5n'], ['*', 'j2n', 'j2n', 'j2n']], ['*', 11.0, 'h2n', ['*', 'h3n', 'h3n', 'h3n'], ['*', 'h5n', 'h5n'], ['*', 'j2n', 'j2n']], ['*', 6.0, 'h2n', ['*', 'h3n', 'h3n', 'h3n'], ['*', 'h5n', 'h5n'], 'j2n'], ['*', 'h2n', ['*', 'h3n', 'h3n', 'h3n'], ['*', 'h5n', 'h5n']], ['*', -1.0, ['*', 'h2n', ['*', 'h3n', 'h3n', 'h3n'], 'h5n', 'h6n', ['*', 'j2n', 'j2n', 'j2n', 'j2n']]], ['*', 17.0, 'h2n', ['*', 'h3n', 'h3n', 'h3n'], 'h5n', 'h6n', ['*', 'j2n', 'j2n', 'j2n']], ['*', 25.0, 'h2n', ['*', 'h3n', 'h3n', 'h3n'], 'h5n', 'h6n', ['*', 'j2n', 'j2n']], ['*', 7.0, 'h2n', ['*', 'h3n', 'h3n', 'h3n'], 'h5n', 'h6n', 'j2n'], ['*', 16.0, 'h2n', ['*', 'h3n', 'h3n', 'h3n'], ['*', 'h6n', 'h6n'], ['*', 'j2n', 'j2n', 'j2n']], ['*', 16.0, 'h2n', ['*', 'h3n', 'h3n', 'h3n'], ['*', 'h6n', 'h6n'], ['*', 'j2n', 'j2n']], ['*', 6.0, 'h2n', ['*', 'h3n', 'h3n'], ['*', 'h5n', 'h5n', 'h5n'], ['*', 'j2n', 'j2n', 'j2n']], ['*', 11.0, 'h2n', ['*', 'h3n', 'h3n'], ['*', 'h5n', 'h5n', 'h5n'], ['*', 'j2n', 'j2n']], ['*', 6.0, 'h2n', ['*', 'h3n', 'h3n'], ['*', 'h5n', 'h5n', 'h5n'], 'j2n'], ['*', 'h2n', ['*', 'h3n', 'h3n'], ['*', 'h5n', 'h5n', 'h5n']], ['*', 3.0, 'h2n', ['*', 'h3n', 'h3n'], ['*', 'h5n', 'h5n'], 'h6n', ['*', 'j2n', 'j2n', 'j2n', 'j2n']], ['*', 25.0, 'h2n', ['*', 'h3n', 'h3n'], ['*', 'h5n', 'h5n'], 'h6n', ['*', 'j2n', 'j2n', 'j2n']], ['*', 39.0, 'h2n', ['*', 'h3n', 'h3n'], ['*', 'h5n', 'h5n'], 'h6n', ['*', 'j2n', 'j2n']], ['*', 20.0, 'h2n', ['*', 'h3n', 'h3n'], ['*', 'h5n', 'h5n'], 'h6n', 'j2n'], ['*', 3.0, 'h2n', ['*', 'h3n', 'h3n'], ['*', 'h5n', 'h5n'], 'h6n'], ['*', 6.0, 'h2n', ['*', 'h3n', 'h3n'], 'h5n', ['*', 'h6n', 'h6n'], ['*', 'j2n', 'j2n', 'j2n', 'j2n']], ['*', 42.0, 'h2n', ['*', 'h3n', 'h3n'], 'h5n', ['*', 'h6n', 'h6n'], ['*', 'j2n', 'j2n', 'j2n']], ['*', 50.0, 'h2n', ['*', 'h3n', 'h3n'], 'h5n', ['*', 'h6n', 'h6n'], ['*', 'j2n', 'j2n']], ['*', 14.0, 'h2n', ['*', 'h3n', 'h3n'], 'h5n', ['*', 'h6n', 'h6n'], 'j2n'], ['*', 16.0, 'h2n', ['*', 'h3n', 'h3n'], ['*', 'h6n', 'h6n', 'h6n'], ['*', 'j2n', 'j2n', 'j2n']], ['*', 16.0, 'h2n', ['*', 'h3n', 'h3n'], ['*', 'h6n', 'h6n', 'h6n'], ['*', 'j2n', 'j2n']], ['*', 4.0, 'h2n', 'h3n', ['*', 'h5n', 'h5n', 'h5n'], 'h6n', ['*', 'j2n', 'j2n', 'j2n', 'j2n']], ['*', 14.0, 'h2n', 'h3n', ['*', 'h5n', 'h5n', 'h5n'], 'h6n', ['*', 'j2n', 'j2n', 'j2n']], ['*', 18.0, 'h2n', 'h3n', ['*', 'h5n', 'h5n', 'h5n'], 'h6n', ['*', 'j2n', 'j2n']], ['*', 10.0, 'h2n', 'h3n', ['*', 'h5n', 'h5n', 'h5n'], 'h6n', 'j2n'], ['*', 2.0, 'h2n', 'h3n', ['*', 'h5n', 'h5n', 'h5n'], 'h6n'], ['*', 'h2n', 'h3n', ['*', 'h5n', 'h5n'], ['*', 'h6n', 'h6n'], ['*', 'j2n', 'j2n', 'j2n', 'j2n', 'j2n']], ['*', 13.0, 'h2n', 'h3n', ['*', 'h5n', 'h5n'], ['*', 'h6n', 'h6n'], ['*', 'j2n', 'j2n', 'j2n', 'j2n']], ['*', 36.0, 'h2n', 'h3n', ['*', 'h5n', 'h5n'], ['*', 'h6n', 'h6n'], ['*', 'j2n', 'j2n', 'j2n']], ['*', 40.0, 'h2n', 'h3n', ['*', 'h5n', 'h5n'], ['*', 'h6n', 'h6n'], ['*', 'j2n', 'j2n']], ['*', 19.0, 'h2n', 'h3n', ['*', 'h5n', 'h5n'], ['*', 'h6n', 'h6n'], 'j2n'], ['*', 3.0, 'h2n', 'h3n', ['*', 'h5n', 'h5n'], ['*', 'h6n', 'h6n']], ['*', 7.0, 'h2n', 'h3n', 'h5n', ['*', 'h6n', 'h6n', 'h6n'], ['*', 'j2n', 'j2n', 'j2n', 'j2n']], ['*', 21.0, 'h2n', 'h3n', 'h5n', ['*', 'h6n', 'h6n', 'h6n'], ['*', 'j2n', 'j2n', 'j2n']], ['*', 21.0, 'h2n', 'h3n', 'h5n', ['*', 'h6n', 'h6n', 'h6n'], ['*', 'j2n', 'j2n']], ['*', 7.0, 'h2n', 'h3n', 'h5n', ['*', 'h6n', 'h6n', 'h6n'], 'j2n'], ['*', 'h2n', ['*', 'h5n', 'h5n', 'h5n'], ['*', 'h6n', 'h6n'], ['*', 'j2n', 'j2n', 'j2n', 'j2n', 'j2n']], ['*', 5.0, 'h2n', ['*', 'h5n', 'h5n', 'h5n'], ['*', 'h6n', 'h6n'], ['*', 'j2n', 'j2n', 'j2n', 'j2n']], ['*', 10.0, 'h2n', ['*', 'h5n', 'h5n', 'h5n'], ['*', 'h6n', 'h6n'], ['*', 'j2n', 'j2n', 'j2n']], ['*', 10.0, 'h2n', ['*', 'h5n', 'h5n', 'h5n'], ['*', 'h6n', 'h6n'], ['*', 'j2n', 'j2n']], ['*', 5.0, 'h2n', ['*', 'h5n', 'h5n', 'h5n'], ['*', 'h6n', 'h6n'], 'j2n'], ['*', 'h2n', ['*', 'h5n', 'h5n', 'h5n'], ['*', 'h6n', 'h6n']], ['*', 'h2n', ['*', 'h5n', 'h5n'], ['*', 'h6n', 'h6n', 'h6n'], ['*', 'j2n', 'j2n', 'j2n', 'j2n', 'j2n']], ['*', 5.0, 'h2n', ['*', 'h5n', 'h5n'], ['*', 'h6n', 'h6n', 'h6n'], ['*', 'j2n', 'j2n', 'j2n', 'j2n']], ['*', 10.0, 'h2n', ['*', 'h5n', 'h5n'], ['*', 'h6n', 'h6n', 'h6n'], ['*', 'j2n', 'j2n', 'j2n']], ['*', 10.0, 'h2n', ['*', 'h5n', 'h5n'], ['*', 'h6n', 'h6n', 'h6n'], ['*', 'j2n', 'j2n']], ['*', 5.0, 'h2n', ['*', 'h5n', 'h5n'], ['*', 'h6n', 'h6n', 'h6n'], 'j2n'], ['*', 'h2n', ['*', 'h5n', 'h5n'], ['*', 'h6n', 'h6n', 'h6n']], ['*', 6.0, ['*', 'h3n', 'h3n', 'h3n'], ['*', 'h5n', 'h5n'], 'h6n', ['*', 'j2n', 'j2n', 'j2n', 'j2n']], ['*', 15.0, ['*', 'h3n', 'h3n', 'h3n'], ['*', 'h5n', 'h5n'], 'h6n', ['*', 'j2n', 'j2n', 'j2n']], ['*', 12.0, ['*', 'h3n', 'h3n', 'h3n'], ['*', 'h5n', 'h5n'], 'h6n', ['*', 'j2n', 'j2n']], ['*', 3.0, ['*', 'h3n', 'h3n', 'h3n'], ['*', 'h5n', 'h5n'], 'h6n', 'j2n'], ['*', 12.0, ['*', 'h3n', 'h3n', 'h3n'], 'h5n', ['*', 'h6n', 'h6n'], ['*', 'j2n', 'j2n', 'j2n', 'j2n']], ['*', 24.0, ['*', 'h3n', 'h3n', 'h3n'], 'h5n', ['*', 'h6n', 'h6n'], ['*', 'j2n', 'j2n', 'j2n']], ['*', 12.0, ['*', 'h3n', 'h3n', 'h3n'], 'h5n', ['*', 'h6n', 'h6n'], ['*', 'j2n', 'j2n']], ['*', 6.0, ['*', 'h3n', 'h3n'], ['*', 'h5n', 'h5n', 'h5n'], 'h6n', ['*', 'j2n', 'j2n', 'j2n', 'j2n']], ['*', 15.0, ['*', 'h3n', 'h3n'], ['*', 'h5n', 'h5n', 'h5n'], 'h6n', ['*', 'j2n', 'j2n', 'j2n']], ['*', 12.0, ['*', 'h3n', 'h3n'], ['*', 'h5n', 'h5n', 'h5n'], 'h6n', ['*', 'j2n', 'j2n']], ['*', 3.0, ['*', 'h3n', 'h3n'], ['*', 'h5n', 'h5n', 'h5n'], 'h6n', 'j2n'], ['*', 3.0, ['*', 'h3n', 'h3n'], ['*', 'h5n', 'h5n'], ['*', 'h6n', 'h6n'], ['*', 'j2n', 'j2n', 'j2n', 'j2n', 'j2n']], ['*', 21.0, ['*', 'h3n', 'h3n'], ['*', 'h5n', 'h5n'], ['*', 'h6n', 'h6n'], ['*', 'j2n', 'j2n', 'j2n', 'j2n']], ['*', 39.0, ['*', 'h3n', 'h3n'], ['*', 'h5n', 'h5n'], ['*', 'h6n', 'h6n'], ['*', 'j2n', 'j2n', 'j2n']], ['*', 27.0, ['*', 'h3n', 'h3n'], ['*', 'h5n', 'h5n'], ['*', 'h6n', 'h6n'], ['*', 'j2n', 'j2n']], ['*', 6.0, ['*', 'h3n', 'h3n'], ['*', 'h5n', 'h5n'], ['*', 'h6n', 'h6n'], 'j2n'], ['*', 12.0, ['*', 'h3n', 'h3n'], 'h5n', ['*', 'h6n', 'h6n', 'h6n'], ['*', 'j2n', 'j2n', 'j2n', 'j2n']], ['*', 24.0, ['*', 'h3n', 'h3n'], 'h5n', ['*', 'h6n', 'h6n', 'h6n'], ['*', 'j2n', 'j2n', 'j2n']], ['*', 12.0, ['*', 'h3n', 'h3n'], 'h5n', ['*', 'h6n', 'h6n', 'h6n'], ['*', 'j2n', 'j2n']], ['*', 3.0, 'h3n', ['*', 'h5n', 'h5n', 'h5n'], ['*', 'h6n', 'h6n'], ['*', 'j2n', 'j2n', 'j2n', 'j2n', 'j2n']], ['*', 12.0, 'h3n', ['*', 'h5n', 'h5n', 'h5n'], ['*', 'h6n', 'h6n'], ['*', 'j2n', 'j2n', 'j2n', 'j2n']], ['*', 18.0, 'h3n', ['*', 'h5n', 'h5n', 'h5n'], ['*', 'h6n', 'h6n'], ['*', 'j2n', 'j2n', 'j2n']], ['*', 12.0, 'h3n', ['*', 'h5n', 'h5n', 'h5n'], ['*', 'h6n', 'h6n'], ['*', 'j2n', 'j2n']], ['*', 3.0, 'h3n', ['*', 'h5n', 'h5n', 'h5n'], ['*', 'h6n', 'h6n'], 'j2n'], ['*', 3.0, 'h3n', ['*', 'h5n', 'h5n'], ['*', 'h6n', 'h6n', 'h6n'], ['*', 'j2n', 'j2n', 'j2n', 'j2n', 'j2n']], ['*', 12.0, 'h3n', ['*', 'h5n', 'h5n'], ['*', 'h6n', 'h6n', 'h6n'], ['*', 'j2n', 'j2n', 'j2n', 'j2n']], ['*', 18.0, 'h3n', ['*', 'h5n', 'h5n'], ['*', 'h6n', 'h6n', 'h6n'], ['*', 'j2n', 'j2n', 'j2n']], ['*', 12.0, 'h3n', ['*', 'h5n', 'h5n'], ['*', 'h6n', 'h6n', 'h6n'], ['*', 'j2n', 'j2n']], ['*', 3.0, 'h3n', ['*', 'h5n', 'h5n'], ['*', 'h6n', 'h6n', 'h6n'], 'j2n']]
]

Perform the following steps: First, analyze the example and detect the underlying patterns. Finally, implement the logic and provide the python code. Your code must be able to deal with negative variables, too. So do not use logarithms. Make sure to place all tensors on the same device that the x tensor is on.

Write a function `compute_result(formulas, x, inputs)` that receives the formulas, the variables x and the mapping to variable names as an input, and returns a tensor with the corresponding results.
\end{spverbatim}

\paragraph{LLM Response}
\begin{spverbatim}
To solve this problem, we'll parse the formulas to extract individual terms, compute the necessary variable powers and products efficiently, and sum the terms to obtain the final result for each formula. Here's the code:

```python
import torch
from collections import defaultdict, Counter

def compute_result(formulas, x, inputs):
    device = x.device
    batch_size = x.size(0)
    num_vars = len(inputs)

    # Map variable names to indices in x
    varname_to_index = {var_name: idx for idx, var_name in enumerate(inputs)}

    terms_coefficients = []
    terms_exponents = []
    terms_formula_idx = []

    # Function to parse terms recursively
    def parse_term(node):
        if isinstance(node, float):
            return node, Counter()
        elif isinstance(node, str):
            return 1.0, Counter({node: 1})
        elif isinstance(node, list):
            op = node[0]
            args = node[1:]
            if op == '*':
                coeff = 1.0
                exponents = Counter()
                for arg in args:
                    c, e = parse_term(arg)
                    coeff *= c
                    exponents.update(e)
                return coeff, exponents
            else:
                raise ValueError(f"Unsupported operation '{op}' in term")
        else:
            raise ValueError(f"Unsupported term node: {node}")

    # Function to parse formulas and collect terms
    def parse_formula(node, formula_idx, formula_coeff):
        if isinstance(node, list):
            op = node[0]
            args = node[1:]
            if op == '+' or op == '-':
                op_coeff = 1.0 if op == '+' else -1.0
                for arg in args:
                    parse_formula(arg, formula_idx, formula_coeff * op_coeff)
            elif op == '*':
                coeff, exponents = parse_term(node)
                terms_coefficients.append(formula_coeff * coeff)
                terms_exponents.append(exponents)
                terms_formula_idx.append(formula_idx)
            else:
                raise ValueError(f"Unsupported operation '{op}' in formula")
        else:
            coeff, exponents = parse_term(node)
            terms_coefficients.append(formula_coeff * coeff)
            terms_exponents.append(exponents)
            terms_formula_idx.append(formula_idx)

    # Parse each formula
    for formula_idx, formula in enumerate(formulas):
        if formula[0] == '-':
            formula_coeff = -1.0
            parse_formula(formula[2], formula_idx, formula_coeff)
        elif formula[0] == '+':
            formula_coeff = 1.0
            parse_formula(formula[1:], formula_idx, formula_coeff)
        else:
            raise ValueError(f"Unsupported formula structure: {formula}")

    num_terms = len(terms_coefficients)

    # Collect all variable names
    all_var_names = set()
    for exponents in terms_exponents:
        all_var_names.update(exponents.keys())

    # Prepare exponents tensor
    exponents_tensor = torch.zeros((num_terms, num_vars), dtype=torch.float32, device=device)
    for term_idx, exponents in enumerate(terms_exponents):
        for var_name, exponent in exponents.items():
            var_idx = varname_to_index[var_name]
            exponents_tensor[term_idx, var_idx] = exponent

    # Prepare coefficients tensor
    coefficients_tensor = torch.tensor(terms_coefficients, dtype=torch.float32, device=device)

    # Prepare term formula indices tensor
    terms_formula_idx_tensor = torch.tensor(terms_formula_idx, dtype=torch.long, device=device)

    # Compute x_powers
    x_expanded = x.unsqueeze(1)  # Shape: [batch_size, 1, num_vars]
    exponents_expanded = exponents_tensor.unsqueeze(0)  # Shape: [1, num_terms, num_vars]
    x_powers = torch.pow(x_expanded, exponents_expanded)  # Shape: [batch_size, num_terms, num_vars]

    # Compute term values
    term_products = torch.prod(x_powers, dim=2)  # Shape: [batch_size, num_terms]
    term_values = coefficients_tensor.unsqueeze(0) * term_products  # Shape: [batch_size, num_terms]

    # Compute outputs
    num_formulas = len(formulas)
    outputs = torch.zeros((batch_size, num_formulas), dtype=torch.float32, device=device)
    outputs.scatter_add_(1, terms_formula_idx_tensor.unsqueeze(0).expand(batch_size, num_terms), term_values)

    return outputs
```

This function first parses the formulas to extract terms, coefficients, and variable exponents. It then computes the necessary powers of variables and calculates the term values efficiently using tensor operations. Finally, it sums the term values for each formula to produce the output tensor.

**Usage Example:**
```python
# Example inputs
formulas = [
    ['-', 0.0, ['+', ['*', 2.0, ['*', 'x1', 'x1'], 'x2'], ['*', -1.0, 'x1', ['*', 'x2', 'x2']]]],
    ['+', ['*', 3.0, 'x1', 'x2']]
]
x = torch.tensor([[1.0, 2.0], [3.0, 4.0]])  # Shape: [batch_size, num_vars]
inputs = ['x1', 'x2']

# Compute the result
outputs = compute_result(formulas, x, inputs)
print(outputs)
```

**Note:**
- This code assumes that all variables in the formulas are present in the `inputs` list and that there are no unsupported operations in the formulas.
- It efficiently handles the computation by vectorizing over the batch and terms, avoiding any for-loops over tensors.
- All tensors are placed on the same device as the input tensor `x`.
\end{spverbatim}

\paragraph{Feedback to the LLM}
\begin{spverbatim}
ValueError: Unsupported operation '['*', -1.0, ['*', 2.0, ['*', 'h1n', 'h1n', 'h1n'], ['*', 'h2n', 'h2n'], 'h3n']]' in formula
\end{spverbatim}

\paragraph{LLM Response}
\begin{spverbatim}
You're right; my previous code didn't correctly handle the nested structures and the '-' operator with multiple operands in the formulas. I'll update the code to properly parse and handle nested formulas, including '+' and '-' operators with multiple operands.

Here's the corrected and optimized code:

```python
import torch

def compute_result(formulas, x, inputs):
    device = x.device
    batch_size = x.size(0)
    num_vars = len(inputs)

    # Map variable names to indices in x
    varname_to_index = {var_name: idx for idx, var_name in enumerate(inputs)}

    terms_coefficients = []
    terms_exponents = []
    terms_formula_idx = []

    # Function to parse nodes recursively and collect terms
    def parse_node(node):
        if isinstance(node, float):
            return [(node, {})]
        elif isinstance(node, str):
            return [(1.0, {node: 1})]
        elif isinstance(node, list):
            op = node[0]
            operands = node[1:]
            if op == '+':
                terms = []
                for operand in operands:
                    terms.extend(parse_node(operand))
                return terms
            elif op == '-':
                if len(operands) == 1:
                    # Unary minus
                    terms = parse_node(operands[0])
                    return [(-coeff, exponents) for coeff, exponents in terms]
                elif len(operands) == 2:
                    # Binary subtraction
                    terms1 = parse_node(operands[0])
                    terms2 = parse_node(operands[1])
                    negated_terms2 = [(-coeff, exponents) for coeff, exponents in terms2]
                    return terms1 + negated_terms2
                else:
                    raise ValueError(f"Unsupported '-' operation with operands: {operands}")
            elif op == '*':
                # Compute product of terms
                terms = [(1.0, {})]
                for operand in operands:
                    operand_terms = parse_node(operand)
                    new_terms = []
                    for coeff1, exponents1 in terms:
                        for coeff2, exponents2 in operand_terms:
                            coeff = coeff1 * coeff2
                            exponents = exponents1.copy()
                            for var, exp in exponents2.items():
                                exponents[var] = exponents.get(var, 0) + exp
                            new_terms.append((coeff, exponents))
                    terms = new_terms
                return terms
            else:
                raise ValueError(f"Unsupported operation '{op}'")
        else:
            raise ValueError(f"Unsupported node type: {node}")

    # Parse each formula
    for formula_idx, formula in enumerate(formulas):
        terms = parse_node(formula)
        for coeff, exponents in terms:
            terms_coefficients.append(coeff)
            terms_exponents.append(exponents)
            terms_formula_idx.append(formula_idx)

    num_terms = len(terms_coefficients)

    # Prepare exponents tensor
    exponents_tensor = torch.zeros((num_terms, num_vars), dtype=torch.float32, device=device)
    for term_idx, exponents in enumerate(terms_exponents):
        for var_name, exponent in exponents.items():
            var_idx = varname_to_index[var_name]
            exponents_tensor[term_idx, var_idx] = exponent

    # Prepare coefficients tensor
    coefficients_tensor = torch.tensor(terms_coefficients, dtype=torch.float32, device=device)

    # Prepare term formula indices tensor
    terms_formula_idx_tensor = torch.tensor(terms_formula_idx, dtype=torch.long, device=device)

    # Compute x to the power of exponents
    x_expanded = x.unsqueeze(1)  # Shape: [batch_size, 1, num_vars]
    exponents_expanded = exponents_tensor.unsqueeze(0)  # Shape: [1, num_terms, num_vars]
    x_powers = torch.pow(x_expanded, exponents_expanded)  # Shape: [batch_size, num_terms, num_vars]

    # Compute term values
    term_products = torch.prod(x_powers, dim=2)  # Shape: [batch_size, num_terms]
    term_values = coefficients_tensor.unsqueeze(0) * term_products  # Shape: [batch_size, num_terms]

    # Compute outputs
    num_formulas = len(formulas)
    outputs = torch.zeros((batch_size, num_formulas), dtype=torch.float32, device=device)
    outputs.scatter_add_(1, terms_formula_idx_tensor.unsqueeze(0).expand(batch_size, num_terms), term_values)

    return outputs
```

**Explanation:**

- The `parse_node` function recursively parses the formula tree and handles the nested operations correctly, including unary and binary '-' operators.
- It collects all terms with their coefficients and exponents.
- After collecting all terms, it constructs tensors for coefficients and exponents and computes the powers of variables efficiently using tensor operations.
- It avoids any for-loops over tensors to ensure high efficiency and leverages PyTorch's broadcast and vectorized operations.
- All computations are performed on the same device as the input tensor `x`.

**Usage Example:**

Let's test this function with an example similar to your provided formulas:

```python
import torch

# Example formulas
formulas = [
    ['-', 0.0, ['+', ['*', -1.0, ['*', 2.0, ['*', 'h1p', 'h1p', 'h1p'], ['*', 'h2p', 'h2p'], 'h3p']]]],
    ['+', ['*', 3.0, 'h1n', 'h2n']]
]

# Input tensor x (random values for illustration)
inputs = ['h1p', 'h2p', 'h3p', 'h1n', 'h2n']
batch_size = 4
x = torch.randn(batch_size, len(inputs))

# Compute the result
outputs = compute_result(formulas, x, inputs)
print(outputs)
```

This code will compute the results for the provided formulas without any errors, handling negative variables and nested operations properly.

**Note:**

- The function assumes that variable names in the formulas match exactly with those in the `inputs` list.
- The code efficiently computes the powers and products using tensor operations, ensuring high performance, especially on a GPU.
- Since the formulas can be complex and generate many terms, the function is designed to handle them efficiently without recomputing shared sub-terms.
\end{spverbatim}
\end{document}